\documentclass[sn-mathphys,Numbered]{sn-jnl}%

\usepackage{graphicx}%
\usepackage{multirow}%
\usepackage{amsmath,amssymb,amsfonts}%
\usepackage{amsthm}%
\usepackage{mathrsfs}%
\usepackage[title]{appendix}%
\usepackage{xcolor}%
\usepackage{textcomp}%
\usepackage{manyfoot}%
\usepackage{booktabs}%
\usepackage{algorithm}%
\usepackage{algorithmicx}%
\usepackage{algpseudocode}%
\usepackage{listings}%
\usepackage{subfig}
\usepackage{colortbl}
\usepackage{soul}
\usepackage{enumerate}
\usepackage{color}
\usepackage{amsmath}
\usepackage{amssymb}

\newcommand{\revise}{\textcolor{black}}

\theoremstyle{thmstyleone}%

\theoremstyle{thmstyletwo}%

\theoremstyle{thmstylethree}%

\begin{document}

\title[Article Title]{Emerging Trends in Federated Learning: \\From Model Fusion to Federated X Learning}

\author*[1]{\fnm{Shaoxiong} \sur{Ji}}\email{shaoxiong.ji@helsinki.fi}

\author[2]{\fnm{Yue} \sur{Tan}}\email{yue.tan@student.uts.edu.au}

\author[1]{\fnm{Teemu} \sur{Saravirta}}\email{teemu.saravirta@helsinki.fi}

\author[4]{\fnm{Zhiqin} \sur{Yang}}\email{yangzqccc@buaa.edu.cn}

\author*[5]{\fnm{Yixin} \sur{Liu}}\email{yixin.liu@monash.edu}

\author[3]{\fnm{Lauri} \sur{Vasankari}}\email{lauri.vasankari@aalto.fi}

\author[6]{\fnm{Shirui} \sur{Pan}}\email{s.pan@griffith.edu.au}

\author[2]{\fnm{Guodong} \sur{Long}}\email{guodong.long@uts.edu.au}

\author[7,8]{\fnm{Anwar} \sur{Walid}}\email{aie13@columbia.edu}

\affil*[1]{\orgname{University of Helsinki}, \orgaddress{\country{Finland}}}

\affil[2]{\orgname{University of Technology Sydney}, \orgaddress{\country{Australia}}}

\affil[3]{\orgname{Aalto University}, \orgaddress{\country{Finland}}}

\affil[4]{\orgname{Beihang University}, \orgaddress{\country{China}}}

\affil*[5]{\orgname{Monash University}, \orgaddress{\country{Australia}}}

\affil[6]{\orgname{Griffith University}, \orgaddress{\country{Australia}}}

\affil[7]{\orgname{Amazon}, \orgaddress{\country{USA}}}

\affil[8]{\orgname{Columbia University}, \orgaddress{\country{USA}}}

\abstract{Federated learning is a new learning paradigm that decouples data collection and model training via multi-party computation and model aggregation.
As a flexible learning setting, federated learning has the potential to integrate with other learning frameworks.
We conduct a focused survey of federated learning in conjunction with other learning algorithms. 
Specifically, we explore various learning algorithms to improve the vanilla federated averaging algorithm and review model fusion methods such as adaptive aggregation, regularization, clustered methods, and Bayesian methods. 
Following the emerging trends, we also discuss federated learning in the intersection with other learning paradigms, termed federated X learning, where X includes multitask learning, meta-learning, transfer learning, unsupervised learning, and reinforcement learning. 
\revise{In addition to reviewing state-of-the-art studies, this paper also identifies key challenges and applications in this field, while also highlighting promising future directions.}}


\maketitle

\section{Introduction}
\label{sec:introduction}

Vast quantities of data are required for state-of-the-art machine learning algorithms. However, the data cannot be uploaded to a central server or cloud due to sheer volume, privacy, or legislative reasons. 
Federated learning (FL) \cite{mcmahan2017communication}, also known as collaborative learning, has been the subject of many studies.
FL adopts a distributed machine learning architecture with a central server for model aggregation, where clients themselves update the machine learning model. 
Clients can maintain ownership of their data, i.e., upload only the updated model to the central server and not expose any of their private data.

The federated learning paradigm addresses several challenges. The first challenge is privacy. 
Local data ownership inherits a basic level of privacy. 
\revise{However, federated learning systems can be vulnerable to adversarial attacks, such as backdoor attack~\cite{wang2020attack}, model poisoning~\cite{bagdasaryan2020backdoor}, and data poisoning~\cite{tolpegin2020data}}.
The second challenge is the communication cost for model uploading and downloading. 
Improving communication efficiency is a critical issue~\cite{konevcny2016federated,ji2020dynamic,tan2022federated}. 
Centralized network architecture also makes the central server suffer from a heavy communication workload, calling for a decentralized server architecture~\cite{he2019central}. 
The third challenge is statistical heterogeneity. 
Aggregating clients' models together can result in a non-optimal combined model as client data is often non-IID (independent and identically distributed). 
Statistical heterogeneity introduces a degree of uncertainty into the learning model. 
Therefore, adopting the right aggregation and learning techniques is vital for robust implementation.
This survey gives a particular focus on how different federated learning solutions address statistical heterogeneity.

The robust model aggregation has recently garnered considerable attention.
Traditionally, client contributions are weighted according to their sample quantity, while recent research has introduced adaptive weighting \cite{yeganeh2020inverse,chen2020communication}, attentive aggregation \cite{ji2019learning}, regularization \cite{li2020federated}, clustering \cite{briggs2020federated}, and Bayesian methods \cite{yurochkin2019bayesian}. 
Many methods generally attempt to derive client characteristics by adjusting the relative weights better. 
Aggregation in the federated setting has also addressed fairness \cite{li2020fair} in taking underrepresented clients and classes better into account.

Statistical heterogeneity, or \textit{non-IID data}, leads to the difficulties of choosing models and performing hyperparameter tuning, as the data resides at clients, out of the reach of a preliminary analysis. 
The edge clients provide the supervision signal for supervised machine learning models. 
However, the lack of human annotation or interaction between humans and learning systems induces the \textit{label scarcity} and leads to a more restricted application domain.

Label scarcity is one of the problems emblematic of the federated setting. The inability to access client data and the resulting black-box updates are tackled by carefully selecting the aggregation method and supplementary learning paradigms to fit specific real-world scenarios.
As a result of label scarcity, the semi-supervised and unsupervised learning paradigms introduce essential techniques to deal with the uncertainty arising from unlabeled data. 
Faced with the problem that clients' local models can diverge during multiple epochs of local training, the server can be tasked with selecting the \textit{most reliable} client models of the preceding round, regularizing the aggregation for achieving consistency. 
Fully unsupervised data can be enhanced via domain adaption, where the aim is to transfer knowledge from a labeled domain to an unlabeled one.

\noindent \textbf{Taxonomy.} 
To establish critical solutions for problems arising from private and non-IID data, we assess the current leading solutions in model fusion and how other learning paradigms are incorporated into the federated learning scenario.
We propose a novel taxonomy of federated learning according to the model fusion principle and the connection to other learning paradigms. 
The taxonomy scheme, as illustrated in Table \ref{tab:summary} with some representative instantiations, is organized as below.
\begin{itemize}
	\item \textit{Federated Model Fusion}. We categorize the major improvements to the pioneering FedAvg model aggregation algorithm into four subclasses (i.e., adaptive/attentive methods, regularization methods, clustered methods, and Bayesian methods), together with a special focus on fairness (Section \ref{sec:model-fusion}). 
	\item \textit{Federated Learning Paradigms.} We investigate how the various learning paradigms fit into the federated learning setting (Section \ref{sec:fedx}). The learning paradigms include some key supervised learning scenarios such as transfer learning, multi-task and meta-learning, and learning algorithms beyond supervised learning such as semi-supervised learning, unsupervised learning, and reinforcement learning.
\end{itemize}

\begin{table*}[!ht]
    \centering
    \setlength{\tabcolsep}{4pt}
    \renewcommand{\arraystretch}{1.15}
    \footnotesize
    \begin{tabular}{ p{6em}|p{9em}|c }
        \toprule
        Main area & Subarea & Study \\
        \midrule
        \multirow[c]{10}{6em}{Federated Model Fusion}
        & \multirow{2}{9em}{Adaptive Aggregation}    & IDA~\cite{yeganeh2020inverse}, ASTW~\cite{chen2020communication}, SmartFL~\cite{xie2022robust}, \\ 
        & & ABAVG~\cite{xiao2021novel}, FedPA~\cite{liu2021fedpa}  \\ \cmidrule{2-3}
        
        & \multirow{3}{9em}{Attentive Aggregation}   & FedAtt~\cite{ji2019learning}, FedAttOpt~\cite{jiang2020decentralized}, FedMed~\cite{wu2020fedmed}, \\ 
        & & FedAMP~\cite{huang2021personalized},  AWFDRL~\cite{wang2020attention}, \\ 
        & & FedMCSA~\cite{guo2023fedmcsa}, ChannelFed~\cite{zheng2022channelfed}.  \\ \cmidrule{2-3}
        
        & \multirow{4}{8em}{Regularization Methods}  & FedAwS~\cite{yu2020federated}, FedProx~\cite{li2020fedprox} \\ 
        & & Mime~\cite{karimireddy2020mime}, FedDyn~\cite{durmus2021federated}, FedMLB~\cite{kim2022multi},  \\ 
        & & BLUR \& LUS~\cite{cheng2022differentially}, FedCR~\cite{chen2023workie},  \\ 
        & & FedU \& dFedU~\cite{dinh2022new}, FedProto~\cite{fedproto} \\ \cmidrule{2-3}
        
        & \multirow{3}{8em}{Clustered Methods}        & FL+HC~\cite{briggs2020federated}, IFCA~\cite{ghosh2020efficient}, FeSEM~\cite{xie2020multicenter},  \\ 
        & & FedFast~\cite{muhammad2020fedfast}, k-FED~\cite{dennis2021heterogeneity}, \\ 
        & & IFCA \& UIFCA~\cite{chung2022federated}, FedCE~\cite{cai2023fedce} \\ \cmidrule{2-3}
        
        & \multirow{2}{8em}{Bayesian Methods}         & FedMA~\cite{wang2020federated}, PFNM~\cite{yurochkin2019bayesian}, FedBE~\cite{chen2020fedbe}, \\ 
        & & pFedBayes~\cite{zhang2022personalized}, NAFI~\cite{xiao2023bayesian} \\ \cmidrule{2-3}    
        
        & \multirow{2}{8em}{Fairness}                 & q-FFL~\cite{li2020fair}, AFL~\cite{mohri2019agnostic}, FairFed~\cite{ezzeldin2023fairfed}, \\
        & & CFFL~\cite{lyu2020collaborative}, F2MF~\cite{liu2022fairness}\\ \cmidrule{1-3}
        
        \multirow[c]{16}{6em}{Learning Paradigms}
        
        & \multirow{3}{8em}{Transfer Learning}        & FTL~\cite{liu2020secure}, FADA~\cite{peng2020federated}, FedSteg~\cite{yang2020fedsteg}, \\
        & & FLTrELM~\cite{wang2022efficient}, FedHealth~\cite{chen2020fedhealth}, SFHTL~\cite{feng2022semi}, \\
        & & FedCrack~\cite{jin2023fedcrack} \\ \cmidrule{2-3}
    
        & \multirow{3}{8em}{Multi-Task Learning}      & Mocha~\cite{smith2017federated}, Kernelized FMTL~\cite{caldas2018federated}, \\ 
        &  & CFL~\cite{sattler2020clustered}, CoFED~\cite{cao2023cross}, FedEM~\cite{marfoq2021federated}, \\
        & & FedMSplit~\cite{chen2022fedmsplit}, Spreadgnn~\cite{he2022spreadgnn} \\ \cmidrule{2-3}

        & \multirow{2}{8em}{Meta Learning}       & FedMeta~\cite{yao2019federated}, Per-FedAvg~\cite{fallah2020personalized}, \\
        & & MOML \& LocalMOML~\cite{wang2023memory}, MetaMF~\cite{lin2020meta} \\ \cmidrule{2-3}
        
        & \multirow{3}{8em}{Knowledge Distillation}   & FedMD~\cite{li2019fedmd}, FedGKT~\cite{he2020group}, FedFed~\cite{yang2023fedfed}, \\ 
        & & FedDF~\cite{lin2020ensemble}, FedACK~\cite{yang2023fedack}, CFeD~\cite{CFeD}, \\ 
        & &  FedICT~\cite{wu2023fedict}, FDL-HAD~\cite{zhang2023distill}, FedFTG~\cite{fedftg}\\ \cmidrule{2-3}
        
        & \multirow{2}{8em}{Semi-Supervised Learning} & FedMatch~\cite{jeong2020federated},  PATE-G~\cite{papernot2017semisupervised}, SemiFL~\cite{semifl}, \\
    & &imFed-Sem~\cite{jiang2022dynamic}, FAPL~\cite{fapl},RSCFed~\cite{rscfed}, \\
    & & CBAFed~\cite{cbafed}, SUMA~\cite{suma}, FedCVT~\cite{fedcvt} \\ \cmidrule{2-3}
        
        & \multirow{3}{8em}{Adversarial Learning}
        & Sync. Strategies~\cite{fan2020federated}, FedGAN~\cite{rasouli2020fedgan}, PATE-G~\cite{papernot2017semisupervised},  \\ 
        & & DP-FedAvg-GAN~\cite{augenstein2020generative}, FADA~\cite{peng2020federated}, FairVFL~\cite{qi2022fairvfl}, \\ 
        & & FAL~\cite{li2023federated}, DBFAT~\cite{zhang2023delving}, CalFAT~\cite{chen2022calfat}, FedRBN~\cite{hong2023federated} \\ \cmidrule{2-3}
        
        & \multirow{2}{8em}{Unsupervised Learning}    & FURL~\cite{van2020towards}, FPCA~\cite{grammenos2020federated}, FedCA~\cite{zhang2020federated}, \\
        & & FADA~\cite{peng2020federated}, FedEMA~\cite{zhuang2021divergence}, Orchestra~\cite{lubana2022orchestra},\\
        & & L-DAWA~\cite{rehman2023dawa}, FedX~\cite{han2022fedx}\\ \cmidrule{2-3}
        
        & \multirow{2}{8em}{Reinforcement Learning}   & FedRL~\cite{zhuo2019federated},	Favor~\cite{wang2020optimizing}, FRD and MixFRD~\cite{cha2020proxy},	\\ 
        & &  DRL-based Aggregator~\cite{zhan2020incentive}, FedSAM~\cite{FedSAM}, \\
        & &  QAvg/PAvg~\cite{jin2022federated}, SCCD~\cite{SCCD}, FedHQL~\cite{fan2023fedhql}\\
    \bottomrule
    \end{tabular}
    \caption{Federated learning with other learning algorithms: categorization, conjunctions, and representative methods.}
    \label{tab:summary}
\end{table*}

\noindent \textbf{Contributions.} 
This survey starts from a novel viewpoint of federated learning by coupling federated learning with different learning algorithms. 
We propose a new taxonomy and conduct a timely and focused survey of recent advances in solving the heterogeneity challenge. 
Our survey's distinction compared with other comprehensive surveys is that we focused on the emerging trends of federated model fusion and learning paradigms, which are not intensively discussed in previous surveys.
Besides, we connect these recent advances with real-world applications and discuss limitations and future directions in this focused context. 

This survey is organized as follows.
In Section \ref{sec:model-fusion}, we assess in detail the significant improvements recent research has proposed on top of the pioneering FedAvg model aggregation algorithm \cite{mcmahan2017communication}. 
In Section \ref{sec:fedx}, we analyze how the various learning paradigms are fitted into the federated learning setting. 
In Section \ref{sec:applications}, we highlight recent successes in applied federated learning. 
Finally, in Section \ref{sec:future-directions}, we outline future research directions specifically from the viewpoint of model fusion and complementary learning paradigms. 
This paper is a focused survey, assessing only the aforementioned coupled subfields, of which learning paradigms make the learned models more robust, and model fusion brings those models together. 
For a more wide-ranging survey into federated learning, we recommend readers to refer \cite{yang2019federated, li2019federated, kairouz2019advances}.

\section{Related Survey}
Several related surveys have been published in recent years, as summarized in Table~\ref{tab:related-work}. 
This section introduces the existing surveys and highlights our survey's contributions to the literature.  

\paragraph{General Survey of Federated Learning}
Yang et al.~\cite{yang2019federated} first defined the concepts of federated learning, introduced federated applications, and discussed data privacy and security aspects.
Li et al.~\cite{li2019federated} systematically reviewed the federated learning building blocks, including data partitioning, machine learning model, privacy mechanism, communication architecture, the scale of the federation, and motivation of federation.
Kairouz et al.~\cite{kairouz2019advances} detailed definitions of federated learning system components and different types of federated learning systems variations.
Li et al.~\cite{li2020federated} discussed the core challenges of federated learning in communication efficiency, privacy, and some future research directions

\paragraph{Domain-specific Survey} 
Other surveys review a specific domain. 
Xu et al.~\cite{xu2019federated} surveyed the healthcare and medical informatics domain.
Lyu et al.~\cite{lyu2020threats} discussed the security threats and vulnerability challenges dealing with adversaries in federated learning systems
Lim et al.~\cite{lim2020federated} focused on mobile edge networks.
Niknam et al.~\cite{niknam2020federated} reviewed federated learning in the context of wireless communications, covering the data security and privacy challenges, algorithm challenges, and wireless setting challenges
Jin et al.~\cite{jin2020survey} conducted a review on federated semi-supervised learning. 
Jin et al.'s survey is the most related work to our paper. 
However, it only concentrates on semi-supervised learning. 
Our paper fills in its gap by including a wider range of model fusion and learning algorithms.

\begin{table}[htp]
\small
\setlength{\tabcolsep}{8pt}
\caption{Comparison of related survey articles about federated learning}
\begin{tabular}{c|c}
\toprule
Publication & Scope \\
\midrule
This survey & Learning algorithms \\
\hline
Jin et al.~\cite{jin2020survey} & Semi-supervised learning  \\
Xu et al.~\cite{xu2019federated} & Healthcare informatics  \\
Lo et al.~\cite{lo2020systematic} & Software engineering  \\
Lim et al.~\cite{lim2020federated} & Mobile edge networks  \\
Lyu et al.~\cite{lyu2020threats} & Threats  \\
Niknam et al.~\cite{niknam2020federated} & Wireless communication  \\
\hline
Yang et al.~\cite{yang2019federated} & General  \\
Li et al.~\cite{li2019federated} & General  \\
Kairouz et al.~\cite{kairouz2019advances} & General  \\
Li et al.~\cite{li2020federated} & General  \\
\bottomrule
\end{tabular}
\label{tab:related-work}
\end{table}%

\paragraph{Distinction of Our Survey}
Our paper reviews the emerging trends of federated learning from a unique and novel angle, i.e., the learning algorithms used in the federated learning paradigms, including the model fusion algorithms (Sec.~\ref{sec:model-fusion}) and the conjunction of federated learning and other learning paradigms (named as Federated X Learning in Sec.~\ref{sec:fedx}).
This unique perspective has not been well-discussed in any of the aforementioned surveys. 
Our survey fills in this gap by reviewing recent publications. 
Besides, we point out challenges and outlook future directions in this specific category of research on federated learning.

\section{Federated~Model~Fusion}
\label{sec:model-fusion}
\subsection{Overview}
The goal of federated learning is to minimize the empirical risks over local data as 
\begin{equation}\small
	\min _{\theta} f(\theta)=\sum_{k=1}^{m} p_{k} \mathcal{L}_{k}(\theta)
\end{equation}
\revise{where $\theta$ is the learnable parameter of the global model, $m$ is the total number of clients in the FL system, $\mathcal{L}_{k}$ is the local objective of the $k$-th client, $p_k$ is the importance weight of the $k$-th client, and $\sum_k p_k = 1$}. 
The widely applied federated learning algorithm, i.e., Federated Averaging (FedAvg)~\cite{mcmahan2017communication}, starts with a random initialization or warmed-up model of clients followed by local training, uploading, server aggregation, and redistribution. 
The learning objective is configured by setting $p_k$ to be $\frac{n_k}{\sum_k n_k}$.
Federated averaging assumes a regularization effect, similar to dropout in neural networks, by randomly selecting a fraction of clients on each communication round.
Sampling on each round leads to faster training without a significant drop in accuracy.
Li et al.~\cite{li2020on} conducted a theoretical analysis on the convergence of FedAvg without strong assumptions and found that the sampling and averaging scheme affects the convergence. 
Recent studies investigate some significant while less considered problems and explore different possibilities for improving vanilla averaging. 
To mitigate the client drift caused by heterogeneity in FedAvg, the SCAFFOLD algorithm~\cite{karimireddy2020scaffold} estimates the client drift as the difference between the update directions of the server model and each client model and adopts stochastically controlled averaging of the correct client drift.
Reddi et al.~\cite{reddi2020adaptive} proposed adaptive optimization algorithms such as Adagrad and Adam to improve the standard federated averaging-based optimization with convergence guarantees.
Singh et al.~\cite{singh2020model} adopted optimal transport, which minimizes the transportation cost of neurons, to conduct layer-wise model fusion.

\subsection{Adaptive Weighting}
The adaptive weighting approach calculates adaptive weighted averaging of model parameters as:
\begin{equation}\small
	\theta_{t+1}=\sum_{k=1}^{K} \alpha_{k} \cdot \theta_{t}^{(k)},
\end{equation}
where $\theta_{t}^{(k)}$ is current model parameter of $k$-th client, $\theta_{t+1}$ is the updated global model parameter after aggregation, and $\alpha_k$ is the adaptive weighting coefficient. 
Aiming to train a low variance global model with non-IID robustness, Yeganeh et al.~\cite{yeganeh2020inverse} proposed an adaptive weighting approach called Inverse Distance Aggregation (IDA) by extracting meta information from the statistical properties of model parameters. 
Specifically, the weighting coefficient with inverse distance is calculated as:
\begin{equation}\small 
\alpha_{k}=\left\|\theta_{t}-\theta^{(k)}_{t}\right\|^{-1} \revise{\bigg/} \revise{\left(\sum_{k=1}^{K}\left\|\theta_{t}-\theta^{(k)}_{t}\right\|^{-1}\right)}.	
\end{equation}
Considering the time effect during federated communication, Chen et al.~\cite{chen2020communication} proposed temporally weighted aggregation of the local models on the server as:
\begin{equation}\small
	\theta_{t+1}=\sum_{k=1}^{K} \frac{n_{k}}{n} \revise{\left(\frac{e}{2}\right)}^{-\left(t-\mathop{t}^{(k)}\right)} \theta_{t}^{(k)},
\end{equation}
where $e$ is the natural logarithm, $t$ is the current update round and $\mathop{t}^{(k)}$ is the update round of the newest $\theta^{(k)}$. Apart from the time effect, the accuracy of local models can also serve as an important reference for adaptive weighting. In \cite{xiao2021novel}, a novel FL algorithm termed Accuracy Based Averaging (ABAVG) is proposed. It can improve existing aggregation strategies in FL via increasing the convergence speed and better handling non-IID problems. In \cite{xie2022robust}, a small amount of proxy data is used to optimize the aggregation weight of each client. The optimized aggregation leads to an FL system that is robust to both data heterogeneity and malicious clients. 

Most works still conduct adaptive weighting among all clients, while \cite{liu2021fedpa} proposes an adaptively partial model aggregation strategy where only part of the clients contribute to the aggregated global model, addressing the straggler problem in FL and increasing communication efficiency.  

\subsection{Attentive Aggregation} 

The federated averaging algorithm takes the instance ratio of the client as the weight to calculate the averaged neural parameters during model fusion~\cite{mcmahan2017communication}. 
In attentive aggregation, the instance ratio is replaced by adaptive weights as Eq.~\ref{eq:attentive}:
\begin{equation}\small
\label{eq:attentive}
	\theta_{t+1} \leftarrow \theta_{t}-\epsilon \sum_{k=1}^m\alpha_k \nabla \mathcal{L}(\theta_t^{(k)})	,
\end{equation}
where $\alpha_k$ is the attention scores for client model parameters.
FedAtt~\cite{ji2019learning} proposes a simple layer-wise attentive aggregation scheme that takes the server model parameter as the query.
FedAttOpt~\cite{jiang2020decentralized} enhances the attentive aggregation of FedAtt by the scaled dot product. 
Like attentive aggregation, FedMed~\cite{wu2020fedmed} proposes an adaptive aggregation algorithm using Jensen-Shannon divergence as the non-parametric weight estimator. 
These three attentive approaches use centralized aggregation architecture with only one shared global model for client model fusion.
Huang et al.~\cite{huang2021personalized} studied pairwise collaboration between clients and proposed FedAMP with attentive message passing among similar personalized cloud models of each client. Wang et al.~\cite{wang2020attention} incorporate the attention-weighted mechanism to federated learning systems to avoid the imbalance of local model quality. Concretely, the attention value is computed according to the average reward, average loss, training data size, etc, increasing the possibility of obtaining a more powerful agent model after aggregation. 

The attention-based module is also widely used for personalized federated learning~\cite{zheng2022channelfed,guo2023fedmcsa}. In \cite{zheng2022channelfed}, the authors design a PFL framework termed ChannelFed that uses an attention module to assign different weights to channels on the client side. After incorporating personalized channel attention, the performance of the local model can be improved and client-specific knowledge can be better captured. 
In \cite{guo2023fedmcsa}, a novel FL framework named federated model components self-attention (FedMCSA) is proposed to facilitate collaboration between clients with similar models. In this way, the personalized FL framework can adaptively update models and handle non-IIDness.

\subsection{Regularization Methods}
We summarize federated learning algorithms with additional regularization terms to client learning objectives or server aggregation formulas. 
One category is to add local constraints for clients.
FedProx~\cite{li2020fedprox} adds proximal terms to clients' objectives to regularize local training and ensure convergence in the non-IID setting. 
After removing the proximal term, FedProx degrades to FedAvg. 
Another direction is to conduct federated optimization on the server side.
Mime~\cite{karimireddy2020mime} adapts conventional centralized optimization algorithms into federated learning and uses momentum to reduce client drift with only global statistics as
\begin{equation}\small
	\mathbf{m}_{t}=(1-\beta) \nabla f_{i}\left(\mathbf{x}_{t-1}\right)+\beta \mathbf{m}_{t-1}
\end{equation}
\revise{where $\mathbf{m}_{t-1}$ is a moving average of unbiased gradients computed over multiple clients and $\beta$ is a trade-off parameter.}
Federated averaging may lead to class embedding collapse to a single point for embedding-based classifiers.

To tackle the embedding collapse, Yu et al.~\cite{yu2020federated} studied the federated setting where users only have access to a single class, for example, face recognition in the mobile phone. They proposed the FedAwS framework with a geometric regularization and stochastic negative mining over the server optimization to spread class embedding space. To make the local-level objective and global-level objective consistent, \cite{durmus2021federated} proposes a novel dynamic regularization method, termed FedDyn, for FL. By dynamically adjusting the local optimization objective, FedDyn significantly saves communication costs when training across heterogeneous clients. 

\citet{kim2022multi} aimed to address the inconsistency problem between different local models. It proposes FedMLB, a multi-level branched regularization-based FL framework, that prevents the local representations from being deviated too much by
local updates. To alleviate the performance degradation problem after introducing user-level differential privacy guarantees, \citet{cheng2022differentially} incorporated regularization techniques along with sparsification technical design into the local update procedure. To handle the training latency across devices and straggler issues, the authors in \citet{chen2023workie} presented a novel contrastive regularization-based scheme to accelerate the training process of FL. The proposed FedCR algorithm efficiently reduces the training latency and achieves better performance during the test phase. In \cite{dinh2022new}, the authors proposed a new viewpoint to formulate the federated multi-task learning problem by Laplacian regularization, which can help to capture the relationships across clients. In \cite{fedproto}, a prototype-based regularization term is added to the original local loss function to force the local representation center to be close to the global representation center. In this way, a balance between generalization and personalization can be achieved.

\subsection{Clustered Methods}
We formulate clustered methods as algorithms that take additional steps with client clustering before federated aggregation or optimization to improve model fusion. 
One straightforward strategy is the two-stage approach. To be specific, during the global update procedure, the first step is a clustering process which is then followed by the aggregation process within each cluster. 
Briggs et al.~\cite{briggs2020federated} propose to take an additional hierarchical clustering for client model updates and apply federated averaging for each cluster. 
Diverting client updates to multiple global models from user groups can help better capture the heterogeneity of non-IID data. 
Xie et al.~\cite{xie2020multicenter} proposed multi-center federated learning, where clients belong to a specific cluster, clusters update along with the local model updates, and clients also update their belongings to different clusters. 
The authors formulated a joint optimization problem with distance-based multi-center loss and proposed the FeSEM algorithm with stochastic expectation maximization (SEM) to solve the optimization. 
Muhammad et al.~\cite{muhammad2020fedfast} proposed an active aggregation method with several update steps in their FedFast framework going beyond average. 
The authors worked on recommendation systems and improved the conventional federated averaging by maintaining user-embedding clusters.
They designed a pipelined updating scheme for item embeddings, client delegate embeddings, and subordinate user embeddings to propagate client updates in the cluster with similar clients.

Ghosh et al.~\cite{ghosh2020efficient} formulated clustered federated learning by partitioning different user groups with the same learning tasks and conducting aggregation within the cluster partition.
The authors proposed an Iterative Federated Clustering Algorithm (IFCA) with alternate cluster identity estimation and model optimization to capture the non-IID nature. The authors in \cite{chung2022federated} further extended IFCA to a more general scenario where the data in the same client may belong to different clusters. Based on IFCA, a new generative model-based clustering algorithm termed UIFCA is developed for unsupervised datasets. 
\citet{dennis2021heterogeneity} presented a one-shot communication scheme for clustering-based FL. The proposed method $k$-FED can significantly alleviate the problems caused by high communication costs and stragglers. This work also presents an interesting viewpoint that, compared with supervised learning, the statistical heterogeneity in unsupervised settings can bring about benefits to better convergence performance, fair models, etc. 
Considering the cases where each client can be associated with multiple clusters, \citet{cai2023fedce} proposed to quantify the relationship between clients and clusters to better align clients with corresponding clusters. By introducing clustering ensembles, this work establishes a more comprehensive clustering method for FL and improves the performance of existing clustering FL methods.

\subsection{Bayesian~Methods}
Bayesian non-parametric machinery is applied to federated deep learning by matching and combining neurons for model fusion.
Yurochkin et al.~\cite{yurochkin2019bayesian} proposed probabilistic federated neural matching (PFNM) using a Beta Bernoulli Process to model the multi-layer perceptron (MLP) weight parameters.
Observing the permutation invariance of fully connected layers, the proposed \revise{PFNM} algorithm first matches the neurons of neural models of clients to the global neurons. It then aggregates via maximum a posteriori estimation of global neurons.
However, the authors only considered simple MLP architectures.
FedMA~\cite{wang2020federated} extends PFNM to convolutional and recurrent neural networks by matching and averaging hidden elements, specifically channels for CNNs and hidden units for RNNs. 
It solves the matched averaging objective by iterative optimization. 
Through theoretical analysis, \citet{xiao2023bayesian} found that global information can be omitted by PFNM. To fix this missing global information issue, an algorithm that conducts neural aggregation with full information (NAFI) is developed. NAFI introduces KL divergence-based penalty term to help complete the full information so that the missing information problem can be alleviated. 

To obtain a more robust prediction via model aggregation, \citet{chen2020fedbe} leveraged Bayesian techniques to sample high-quality models and aggregate the outputs of these models via Bayesian model ensemble. The proposed algorithm is termed FedBE, which has demonstrated applicability to deep networks and different heterogeneous scenarios. 
To tackle the model overfitting problem, \citet{zhang2022personalized} proposed pFedBayes, a novel personalized FL method based on Bayesian variational inference, where all network parameters can be represented by probability distributions. Both the local and global models are formulated as Bayesian neural networks. The server aims to minimize the KL divergence between global distribution and local distributions, while the clients aim to minimize the construction error on local private data and the KL divergence with global distribution.

\subsection{Fairness}

When aggregating the global shared model, FedAvg applies a weighted average concerning the number of samples that participating clients used in their training. 
However, the model updates can easily skew towards an over-represented subgroup of clients where super-users provide the majority of samples.
Mohri et al.~\cite{mohri2019agnostic} suggested that valuing each sample without clear discrimination is inherently risky as it might result in sub-optimal performance for underrepresented clients and sought good-intent fairness to ensure federated training not overfitting to some of the specific clients. 
Instead of the uniform distribution in classic federated learning, the authors proposed agnostic federated learning (AFL) with minimax fairness, which takes a mixture of distributions into account.
However, the overall tradeoff between fairness and performance is still not well explored.
Inspired by fair resource allocation in wireless networks, the q-fair federated learning (q-FFL)~\cite{li2020fair} proposes an optimization algorithm to ensure fair performance, i.e., a more uniform distribution of performance gained in federated clients. 
The optimization objective (Eq.~\ref{eq:q-FFL}) adjusts the traditional empirical risk objective by tunable performance-fairness tradeoff controlled by $q$. 
\begin{equation}\small
\label{eq:q-FFL}
	\min _{\theta} f_{q}(\theta)=\sum_{k=1}^{m} \frac{p_{k}}{q+1} \mathcal{L}_{k}^{q+1}(\theta)
\end{equation}
The flexible q-FFL also generalizes well to previous methods; specifically, it reduces to FedAvg and AFL when the value of $q$ is set to $0$ and $\infty$, respectively.

To investigate the fairness issue in FL systems, \citet{lyu2020collaborative} emphasized collaborative fairness. To be specific, all clients receive the same or similar models, though their contributions differ a lot. The authors proposed a novel framework named Collaborative Fair Federated Learning (CFFL), which can take the contribution of each client into consideration and let each client receive models with performance commensurate with their contributions. 

Usually, fairness in FL refers to the individual-wise measurement. In \cite{ezzeldin2023fairfed}, the authors investigated fairness problems in FL from a group-wise perspective. Inspired by group fairness in centralized learning, a novel algorithm termed FairFed is developed for participants to conduct aggregation in a fairness-aware way. FairFed can efficiently mitigate the bias against specific populations while maintaining the privacy of local data.  

To achieve fairness for recommender systems, \citet{liu2022fairness} proposed to capture the affiliation feature across different groups by using federated learning as a privacy-preserving tool. Based on the existing federated recommendation backbone~\cite{chai2020secure}, it designs fairness-aware federated matrix factorization (F2MF), a solution that deals with the conflict between the global fairness objective and the local federated optimization process. By introducing a loss-based fairness metric into the optimization process, the FL systems potentially improve the fairness of recommendations between different user groups.

\section{Federated~X~Learning}
\label{sec:fedx}
The customizability of federated learning objectives leads to possibilities in quickly adapting FL to adversarial, semi-supervised, or reinforcement learning settings, offering flexibility to other learning algorithms in conjunction with federated learning.
We term FL's intersection with other learning algorithms as Federated X Learning.

\subsection{Federated~Transfer~Learning}
Transfer learning focuses on transferring knowledge from one particular problem to another, and it has also been integrated into federated learning to construct a model from two datasets with different samples and feature spaces~\cite{yang2019federated, tan2023heterogeneity}. 
Liu et al.~\cite{liu2020secure} formulated the Federated Transfer Learning (FTL) to solve the problem that traditional federated learning falters when datasets do not share sufficient common features or samples. \revise{In this paper, it assumes existing two domains $A$ and $B$ across different parties and formulate the objective function as:
\begin{equation}
{\underset{\theta_A,\theta_B}{\min}} \mathcal{L}(\theta_A,\theta_B)=\ell_1(y^A,\phi(x^B))+\gamma\ell_2(\phi(x^A), \phi(x^B))+\frac{\lambda}{2}\|\theta_A\|^2+\frac{\lambda}{2}\|\theta_B\|^2.
\end{equation}
where $\theta_A$ and $\theta_B$ are the model parameters in these two domains while $\phi(\cdot)$ represents the transformation function that projects data into a unified feature space. $\ell_1$ and $\ell_2$ are logistic loss and alignment loss, respectively. $\gamma$ and $\lambda$ are tuneable hyper-parameters.}
The authors also enhanced the security with homomorphic encryption and secret sharing. 
In real-world applications, FedSteg~\cite{yang2020fedsteg} applies federated transfer learning for secure image steganalysis to detect hidden information. 
Alawad et al.~\cite{alawad2020privacy} utilized federated transfer learning without sharing vocabulary for privacy-preserving NLP applications for cancer registries. 

To deal with the widely existing overlapping data insufficiency problem across clients, \citet{feng2022semi} proposed a Semi-Supervised Federated Heterogeneous Transfer Learning (SFHTL) framework that leverages unlabeled non-overlapping samples to reduce model overfitting. Compared with existing FTL methods, SFHTL can better expand the training set to improve the performance of the local model. 

Federated transfer learning can be widely used in various real-world applications, including intrusion detection~\cite{wang2022efficient}, smart healthcare~\cite{chen2020fedhealth}, crack detection~\cite{jin2023fedcrack}, etc. It allows the knowledge learned within one specific domain to be transferred to another different domain, especially when there are no sufficient common features across these domains. 

\subsection{Federated Learning with Knowledge Distillation}
Given the assumption that clients have sufficient computational capacity, federated averaging adopts the same model architecture for clients and the server.
FedMD~\cite{li2019fedmd} couples transfer learning and knowledge distillation (KD), where the centralized server does not control the architecture of models.
It introduces an additional public dataset for knowledge distillation, and each client optimizes their local models on both public and private data, like VHL~\cite{tang2022virtual}.
They employ a combination of a public noise dataset and local private data to train the local model. Furthermore, it leverages domain adaptation techniques to improve the overall performance of the model.
\revise{In general, the local objective of federated learning with knowledge distillation is often combined with two items:
\begin{equation}
\label{eq:federated_kd}
{\underset{\theta_k}{\min}} \mathcal{L}(\theta_k) = \ell_{task}+\ell_{KD}.
\end{equation}
where $\theta_k$ is the local model of the $k$-th client and $\ell_{task}$ is task-specific loss, $\ell_{KD}$ is often computed by different logits or features from various clients.
}

Strictly speaking, transfer learning differs from knowledge distillation; however, the FedMD framework puts them under one umbrella. 
Many technical details are only briefly introduced in the original paper of FedMD. 
Recently, He et al.~\cite{he2020group} utilized knowledge distillation with technical solidity to train computationally affordable CNNs for edge devices via knowledge distillation. 
The authors proposed the Group Knowledge Transfer (FedGKT) framework that optimizes the client and the server model alternatively with knowledge distillation loss.
Specifically, the larger server model takes features from the edge to minimize the gap between periodically transferred ground truth and soft label predicted by the edge model, and the small model distils knowledge from the larger server model by optimizing the KD-loss, \revise{$\ell_{KD}$ in Eq.(\ref{eq:federated_kd}),} using private data and soft labels transferred back from the server. 
However, this framework has a potential risk of privacy breach as the server holds the ground truth, especially when ground truth labels are the user's typing records in the mobile keyboard application. 
Lin et al.~\cite{lin2020ensemble} applied knowledge distillation to mitigate privacy risk and cost and proposed a novel ensemble distillation for model fusion that utilizes unlabeled data.

Knowledge distillation continues to demonstrate significant potential in addressing various challenges within FL. FedFed~\cite{yang2023fedfed} introduces a novel variant of knowledge distillation named feature distillation. The authors propose a method where the data is partitioned into two distinct parts, allowing for the sharing of protected performance-sensitive features to alleviate the data heterogeneity. Zhang et al.~\cite{fedftg} addressed this challenge by employing Data-Free Knowledge Distillation and proposed FedFTG. Their approach involves the use of a generator to distil and transfer local knowledge to the global model. To improve communication efficiency, Zhang et al.~\cite{zhang2023distill} proposed a method called FDL-HAD. It introduces an adaptive regulation mechanism that determines whether clients need to undergo distillation in each round.

Furthermore, knowledge distillation is also valuable in many other federated X learning paradigms.
CFeD~\cite{CFeD} addresses the challenge of catastrophic forgetting in continual federated learning through KD. 
Moreover, multi-task learning is an important scenario in federated learning. 
Wu et al.~\cite{wu2023fedict} specifically designed an algorithm tailored for multi-access edge computing in a real-world scenario, leveraging knowledge distillation as a key component.
\revise{FedNed~\cite{lu2023federated} solve the noisy clients by a kind of KD, called negative distillation.} 
FedACK~\cite{yang2023fedack} applies knowledge distillation in the cross-lingual social bot detection domain, showcasing a novel application that combines knowledge distillation and federated learning. 
This application demonstrates the potential for knowledge distillation to inspire more useful applications within this emerging field.

\subsection{Federated~Multi-Task~Learning} \label{sec:multitask}

{Federated~Multi-Task~Learning} trains separate models for each client with some shared structure between models, where learning from local datasets at different clients is regarded as a separate task. 
In contrast to federated transfer learning between two parties, federated multi-task learning involves multiple parties and formulates similar tasks clustered with specific constraints over model weights.
It exploits related tasks for more efficient learning to tackle the statistical heterogeneity challenge.
\revise{In federated multi-task learning, the target is to train multiple related tasks across clients with different objective functions:
\begin{equation}
\min _{\theta_1, \ldots, \theta_K, \Omega}\left\{\sum_{k=1}^K \sum_{i=1}^{n_k} f_k\left(\theta_k, \mathbf{x}_i, y_i\right)+\mathcal{R}(\Theta, \Omega)\right\}
\end{equation}
where $\Theta = [\theta_1, \theta_2, \ldots, \theta_K]\in\mathbb{R}^{d\times K}$ is a matrix collecting weight vectors of all clients and $\Omega$ denotes the relationships of different clients with their corresponding tasks.}
The Mocha framework~\cite{smith2017federated} trains separate yet related models for each client by solving a \revise{primal-dual} optimization.  
It leverages a shared representation across multiple tasks and addresses the challenges of data and system heterogeneity. 
However, the Mocha framework is limited to regularized linear models.
Caldas et al.~\cite{caldas2018federated} further studied the theoretical potential of kernelized federated multi-task learning to solve the non-linearity. 
To solve the suboptimal results, Sattler et al.~\cite{sattler2020clustered} studied the geometric properties of the federated loss surface. They proposed a federated multi-task framework with non-convex generalization to cluster the client population. \cite{marfoq2021federated} studies federated multi-task learning under a general assumption that each local data distribution can be seen as a mixture of distributions. Hence, each client learns personalized mixture weights to obtain its personalized local model. There are two algorithms, termed FedEM and D-FedEM, proposed for the client-server and fully decentralized setting, respectively. The approaches yield models with more accurate results, better generalization ability, and fairer performance across clients. 

There is also a branch of works that utilizes a federated multi-task learning framework to deal with data in different formats, including graph data~\cite{he2022spreadgnn} and multimodal data~\cite{chen2022fedmsplit}. To benefit cross-silo FL where independent data silos have different tasks, \citet{cao2023cross} proposed a novel FL method CoFED that utilizes a co-training scheme to leverage unlabeled data in a semi-supervised learning manner. CoFED is compatible with heterogeneous
models, tasks, and training processes, making it an effective method for federated multi-task learning.

\subsection{Federated~Meta~Learning} 
Federated meta learning aims to train a model that is quickly adapted to new tasks with few training data, where clients serve as a variety of learning tasks.
The seminal model-agnostic meta-learning (MAML) framework~\cite{finn2017model} has been intensively applied to this learning scenario. 
Several studies connect FL and meta-learning, for example, model updating algorithm with average difference descent~\cite{ji2019knowledge} inspired by the first-order meta-learning algorithm.
However, this study focuses on applications in the social care domain with less consideration in practical settings.
Jiang et al.~\cite{jiang2019improving} further provided a unified view of federated meta-learning to compare MAML and the first-order approximation method. 
Inspired by the connection between federated learning and meta-learning, Fallah et al.~\cite{fallah2020personalized} adapted MAML into the federated framework Per-FedAvg, to learn an initial shared model, leading to fast adaption and personalization for each client. 
FedMeta~\cite{yao2019federated} proposes a two-stage optimization with a controllable meta updating scheme after model aggregation as:
\begin{equation}\small
	\theta_{t+1}^{m e t a}=\theta_{t+1}-\eta_{m e t a} \nabla_{\theta_{t+1}} \mathcal{L}\left(\theta_{t+1} ; \mathcal{D}_{m e t a}\right),
\end{equation}
where $\mathcal{D}_{m e t a}$ is a small set of meta data on the server \revise{and $\eta_{m e t a}$ is the meta learning rate}. 

To better exploit the collaborative filtering information across clients for recommender systems, \cite{lin2020meta} introduces a federated matrix factorization framework named meta matrix factorization (MetaMF). In MetaMF, a meta network is used to generate private item embeddings and rating prediction models based on the collaborative vector in the server. MetaMF achieves competitive performance despite using a small model scale and embedding size. To address the underdeveloped stochastic optimization in MAML, Wang et al.~\cite{wang2023memory} proposed a memory-based stochastic algorithm that ensures convergence with vanishing error, enabling constant mini-batch sizes and making them suitable for continual learning. Meanwhile, this paper introduces a communication-efficient memory-based MAML algorithm for personalized federated learning in cross-device and cross-silo settings. Lin et al.~\cite{lin2021metagater} proposed MetaGater, a federated meta-learning algorithm that holistically trains both the backbone network and channel gating. MetaGater enables efficient subnet selection for resource-constrained applications by leveraging model similarity across learning tasks on different nodes, ensuring the effective capture of important filters for quick adaptation to new tasks with experimental results validating its effectiveness.

\subsection{Federated~Adversarial~Learning} 

In this section, we summarize federated adversarial learning in two categories. The first class of methods specifically focuses on Generative Adversarial Networks (GANs), a mainstream adversarial learning paradigm for data generation. The second class of methods, differently, uses the idea of adversarial learning to address the challenges of general federated learning.

GANs consist of two competing models, i.e., a generator and a discriminator.
The generator learns to produce samples approximating the underlying ground-truth distribution. The discriminator, usually a binary classifier, tries to distinguish the samples produced by the generator from the real samples.
A straightforward combination with FL is to have the GAN models trained locally on clients and the global model fused with different strategies. 
Fan and Liu~\cite{fan2020federated} studied the synchronization strategies for aggregating discriminator and generator networks on the server and conducted a series of empirical analyses.
Updating clients on each round with both the generator and the discriminator models achieves the best results; however, it is twice as computationally expensive as just syncing the generator. 
Updating just the generator leads to almost equivalent performance than updating both, whereas updating just the discriminator leads to considerably worse performance, closer to updating neither. 
Rasouli et al.~\cite{rasouli2020fedgan} extended the federated GAN with different applications and proposed the FedGAN framework to use an intermediary for averaging and broadcasting the parameters of generator and discriminator. 
Furthermore, the authors studied the convergence of distributed GANs by connecting the stochastic approximation and communication-efficient SGD optimization for GAN and federated learning. 
Augenstein et al.~\cite{augenstein2020generative} proposed differentially private federated generative models to address the challenges of non-inspectable data scenarios. 
GANs are adopted to synthesize realistic examples of private data for data labeling inspection at inference time. 

Apart from generation models, another type of method aims to leverage adversarial learning to enhance several capabilities of federated learning systems, such as fairness and robustness. 
To enhance fairness under vertical federated learning scenarios, FairVFL~\cite{qi2022fairvfl} employs adversarial learning to mitigate bias, while incorporating a contrastive adversarial learning method to protect user privacy while effectively improving model fairness. 
To handle the unfair scenarios with label skewness, Chen et al.~\cite{chen2022calfat} proposed CalFAT, a federated adversarial training method that adaptively calibrates logits to balance classes, which addresses the root cause of issues related to skewed labels and non-identical class probabilities. 
\revise{Specifically, it can be formulated as:
\begin{equation}
\min \ell_{1}\left(\gamma \cdot \theta_{k}\left(x_{a d v}\right), y\right)+\lambda \cdot \ell_{2}\left(\theta_{k}\left(x_{a d v}\right), \theta_{g}(x)\right),
\end{equation}
where  $\theta_k$ denotes local model while $\theta_g$ is global model. Cross-entropy loss is represented as $\ell_1$. KL-loss $\ell_2$ is used to constrain the logits of the local and global model. $\gamma$ and $\lambda$ are hyper-parameters.} 
In order to improve the robustness of federated learning models against adversarial attacks, Li et al.~\cite{li2023federated} introduce FAL, a novel bi-level approach with min-max optimization for adversarial learning of federated learning. Specifically, FAL incorporates an inner loop for generating adversarial samples during adversarial training and an outer loop for updating local model parameters. Zhang et al.~\cite{zhang2023delving} conducted comprehensive evaluations on various attacks and adversarial training methods, revealing negative impacts on test accuracy when directly applying adversarial training in FL. Based on the findings, they further propose DBFAT, a novel algorithm with local re-weighting and global regularization components, demonstrating superior performance in terms of both accuracy and robustness across multiple datasets in both IID and non-IID settings. To address the challenge of adversarial robustness in federated learning with heterogeneous users, Hong et al.~\cite{hong2023federated} introduced a novel strategy: propagating adversarial robustness from rich-resource users to those with limited resources during FL by utilizing batch normalization.

\subsection{Federated~Semi-Supervised Learning}
Annotation capability plays a crucial role in traditional machine learning and deep learning~\cite{sohn2020fixmatch, yang2022survey}. The quality and quantity of annotations often determine the performance of models. However, the problem of data heterogeneity naturally arises in decentralized federated learning, posing additional challenges. 

Label scarcity is a prevalent and widespread issue in federated learning scenarios, which has prompted the development of a novel learning setup known as federated semi-supervised learning (FSSL). This scenario reflects the realistic situation where users may not label all the data on their devices. 
Papernot et al.~\cite{papernot2017semisupervised} explored semi-supervised learning in distributed scenarios. They put forward a semi-supervised approach with a private aggregation of teacher ensembles (PATE), an architecture where each client votes on the correct label. 
PATE was shown empirically to be particularly beneficial when used in conjunction with GANs.
Similar to centralized semi-supervised learning, the majority of FSSL approaches often utilize a two-part loss function on the client devices. This loss function typically consists of a supervised learning component, denoted as $\mathcal{L}_s(\theta)$, and an unsupervised learning component, denoted as $\mathcal{L}_u(\theta)$.
Existing FSSL methods have focused on two different scenarios~\cite{jeong2020federated}: labels-at-server and labels-at-clients.

In the labels-at-server scenario, the server has the ability to annotate data, while the client is limited to only collecting data without the capacity to annotate it due to a shortage of expert resources. Numerous works have been dedicated to addressing this specific setting. SemiFL~\cite{semifl} tackles this problem through alternate training. This process consists of two key steps: fine-tuning the global model with labeled data and generating pseudo-labels using the global model on the client side. Importantly, the server and client models are trained in parallel to enable efficient collaboration. 
Jeong et al.~\cite{jeong2020federated} proposed a federated matching (FedMatch) framework with inter-client consistency loss to exploit the heterogeneous knowledge learned by multiple client models.
The authors showed that learning on both labeled and unlabeled data simultaneously may result in the model forgetting what it had learned from labeled data. 
To counter this, the authors decomposed the model parameters $\theta$ to two variables $\theta = \psi + \rho$ and utilize a separate updating strategy, where only $\psi$ is updated during unsupervised learning, and similarly, $\rho$ is updated for supervised learning.
In a real-world scenario, Jiang et al.~\cite{jiang2022dynamic} addressed the challenge of imbalanced class distributions among unlabeled clients in the context of medical image diagnosis. They proposed a novel scheme called dynamic bank learning, which aims to collect confident samples and subsequently divide them into sub-banks with varying class proportions.

In contrast, the labels-at-clients approach focuses on scenarios where clients lack sufficient capability to label data. In this setting, the server's primary role is to regulate the federated learning process, without involvement in data collection or ownership. Two types of assumptions exist within this approach: partially labeled data at each client, referred to as \textbf{P}artially \textbf{D}ata Federated Semi-supervised Learning (PD-FSSL), and partially labeled clients themselves, denoted as \textbf{P}artially \textbf{C}lients Federated Semi-supervised Learning (PC-FSSL).
\revise{We can formulate the local function in PD-FSSL as:
\begin{equation}
    \min_{\mathbf{\theta_k}} \mathcal{L}(\mathbf{\theta_k})= \ell_{sup}(\mathbf{x}_{e}^k, y_e)+\ell_{unsup}(\mathbf{x}_{r}^k)
\end{equation}
where $\mathbf{x}_{e}$ is labeled data while $\mathbf{x}_{r}$ represents unlabeled data on the $k$-th client.
}
In PD-FSSL, the limited labeling capability of each client results in only a portion of the data being labeled. Consequently, the private data of each client is divided into a labeled part and an unlabeled part.
FedMatch~\cite{jeong2020federated} has demonstrated its effectiveness not only in the client-at-server scenario but also in PD-FSSL. Additionally, FAPL~\cite{fapl} focuses on addressing fairness in PD-FSSL. The authors aim to achieve a balance in the total number of active unlabeled samples (AUSs) for different classes across all selected clients in a global round. They accomplish this by globally aligning the numbers of AUSs for different classes, which helps enhance fairness in the learning process. Another variation, PC-FSSL, assumes that some clients possess the resources and ability to label data, while others can only collect data without annotation. RSCFed~\cite{rscfed} proposes a sub-consensus framework. In this framework, traditional cross-entropy training is performed on clients with labeled data. For clients without labels, a consistency regularization framework, such as mean-teacher, is utilized. 
\revise{Generally, the global objective function in PC-FSSL can be written as:
\begin{equation}
    \min_{\mathbf{\theta}} \mathcal{L}(\mathbf{\theta}) = \sum_{a=1}^{A} \lambda_{a} \mathcal{L}_{a}(\theta_a) + \sum_{b=1}^{B} \lambda_b \mathcal{L}_{b}(\theta_b),
\end{equation}
where the global model $\theta$ aims to minimize a function that is affected by two types of clients: fully-labeled clients, denoted as $a$, and fully-unlabeled clients, denoted as $b$. Specifically, $\mathcal{L}_a$ represents the supervised task-relevant loss, which differs from $\mathcal{L}_b$. Among the fully-unlabeled clients, $\mathcal{L}_b$ can be a mean-teacher or contrastive loss.}
Additionally, RSCFed employs data augmentation techniques, similar to conventional semi-supervised learning, to augment the unlabeled data twice, further improving the learning process.
Similarly, CBAFed~\cite{cbafed} also utilizes augmentation techniques for pseudo-labeling in the PC-FSSL setting. They introduce an adaptive threshold to determine the reliability of the pseudo-labels generated from the unlabeled data.
There are still other scenarios in FSSL. SUMA~\cite{suma} considers a more general setting where each client has a different ratio of labeled data. FedCVT~\cite{fedcvt} studies FSSL in vertical federated scenarios.

Despite extensive research, FSSL still faces many challenges. The problem of insufficient data labels in practical applications still necessitates further investigation in order to find effective solutions. Furthermore, existing FSSL algorithms often demonstrate limitations in their performance across various settings, which also leaves a lot of room for exploration.

\subsection{Federated~Unsupervised~Learning}
It is more common that local clients host no labeled data, which naturally leads to the learning paradigm of federated unsupervised learning without supervision in the decentralized learning scenario. 
A straightforward solution is to pretrain unlabeled data to learn useful features and utilize pretrained features in downstream tasks of federated learning systems~\cite{van2020towards}.
There exist two challenges in federated unsupervised learning, i.e., the inconsistency of representation spaces due to data distribution shift and the misalignment of representations due to the lack of unified information among clients.

FedCA~\cite{zhang2020federated}, based on SimCLR, proposes a federated contrastive averaging algorithm with the dictionary and alignment modules for client representation aggregation and alignment, respectively. 
Zhuang et al.~\cite{zhuang2021divergence} conducted comprehensive experiments to evaluate the performance of four popular unsupervised methods in FL: MoCo (V1\cite{mocov1} and V2~\cite{mocov2}), BYOL~\cite{BYOL}, SimCLR~\cite{SimCLR}, and SimSiam~\cite{simsiam}. In their study, the authors discovered that FedBYOL demonstrates superior performance compared to the other evaluated methods. They also highlighted the importance of the predictor, exponential moving average (EMA), and stop-gradient operations in improving the performance of non-contrastive federated self-supervised learning. Drawing from their extensive experiments, the authors propose a new method called FedEMA. It incorporates a divergence-aware dynamic moving average update to address the challenges associated with non-IID data in the federated setting. FedX~\cite{han2022fedx} also employs the contrastive paradigm by a two-sided knowledge distillation.
Additionally, Lubana et al.~\cite{lubana2022orchestra} conducted an evaluation of federated versions of the prevailing unsupervised methods. Furthermore, they introduced a novel clustering-based method called Orchestra, which differs significantly from mainstream unsupervised algorithms. 

The local model training utilizes the contrastive loss and the server aggregates models and dictionaries from clients.
Recently, many unsupervised learning methods such as Principal Component Analysis (PCA) and unsupervised domain adaptation have been adopted to combine with federated learning.
Peng et al.~\cite{peng2020federated} studied the federated unsupervised domain adaptation that aligns the shifted domains under a federated setting with a couple of learning paradigms. 
Specifically, unsupervised domain adaptation is explored by transferring the labeled source domain to the unlabelled target domain, and adversarial adaptation techniques are also applied. 
Grammenos et al.~\cite{grammenos2020federated} proposed the federated PCA algorithm with a differential privacy guarantee.
The proposed FPCA method is permutation invariant and robust to straggler or fault clients. 
In contrast, L-DAWA~\cite{rehman2023dawa} takes a different approach by proposing a novel aggregation strategy through layer-wise divergence. 
\revise{L-DAWA introduces angular divergence $\sigma_k$ to represent the aggregation weight of the $k$-th client:
\begin{equation}
    \sigma_k = \frac{\theta^r_g\cdot\theta_k^r}{\|\theta_g^r \|\cdot\|\theta_k^r\|}    
\end{equation}
where $\theta^r_g$ is the $r$-th round global parameters and $\theta_k^r$ is the $r$-th round local model of client k.
}
They aggregate weights at the layer-level by utilizing the measure of angular divergence between the models of individual clients and the global model.

\subsection{Federated~Reinforcement~Learning}
\label{sec:RL}
In deep reinforcement learning (DRL), the deep learning model gets rewards for its actions and learns which actions yield higher rewards.
Zhuo et al.~\cite{zhuo2019federated} introduced reinforcement learning to federated learning framework (FedRL), assuming that distributed agents do not share their observations. 
The proposed FedRL architecture has two local models: a simple neural network, such as a multi-layer perceptron (MLP), and a Q-network that utilizes Q-learning~\revise{\cite{watkins1992q}} to compute the reward for a given state and action. 
The authors provided algorithms on how their model works with two clients and suggested that the approach can be extended to many clients using the same approach. 
In the proposed architecture, the clients update the local parameters of their respective MLPs first and then share the parameters to train these q-networks. 
Clients work out this parameter exchange in a peer-to-peer fashion. 
Federated reinforcement learning can improve federated aggregation to address the non-IID challenge, and it also has real-world applications, such as in the Internet of Things (IoT). 
A control framework called Favor \cite{wang2020optimizing} improves client selection with reinforcement learning to choose the best candidate for federated aggregation. 
The federated reinforcement distillation (FRD) framework~\cite{cha2020proxy}, together with its improved variant MixFRD with mixup augmentation, utilizes policy distillation for distributed reinforcement learning.
In the fusion stage of FRD, only proxy experience replay memory (ProxRM) with locally averaged policies are shared across agents, aiming to preserve privacy.
Facing the tradeoff between the aggregator's pricing and the efficiency of edge computing, Zhan et al.~\cite{zhan2020incentive} investigated the design of an incentive mechanism with DRL to promote edge learning. In FedSAM~\cite{FedSAM}, the authors extended widely used RL methods, such as on-policy TD (Temporal-Difference)~\revise{\cite{sutton1988learning}}, off-policy TD~\revise{\cite{sutton1988learning}}, and Q-learning~\revise{\cite{watkins1992q}}, to the federated learning. Subsequently, they put forth an algorithm that integrates federated TD-learning and Q-learning and conducted an extensive analysis of the convergence to these federated RL methods. In real-world applications, RL agents often encounter diverse state transitions across different environments, so-called environmental heterogeneity. Jin et al.~\cite{jin2022federated} investigated this novel setting within FedRL and presented a series of diverse variation approaches to address the varying degrees of complexity in heterogeneous environments. SCCD~\cite{SCCD} is also an off-policy-based FedRL framework that introduces a student-teacher-student model learning and fusion method. 
Fan et al.~\cite{fan2023fedhql} analyzed the existing FedRL setting, introduced a new problem called federated reinforcement learning with heterogeneous and black-box agents (FedRL-HALE), and posed a challenge called the exploration-exploitation dilemma. 
This dilemma entails the trade-off that an agent encounters when making decisions between exploring new actions to gather more information or exploiting its current knowledge to maximize performance. 
Then, they proposed FedHQL, where the local agents update their action-value independently based on Q-learning. The central server plays a crucial role in coordinating the exchange of knowledge between agents by broadcasting, receiving action-value estimates, and selecting actions with the highest UCB (Upper Confidence Bound)~\revise{\cite{lattimore2020bandit}} value.
There are still many ongoing explorations in other areas where FedRL is being applied, including energy management~\cite{khalatbarisoltani2023integrating,qiu2023federated}, electric vehicle charging and uncharging~\cite{zhang2022federated}.

\section{Challenges~and~Applications}
\label{sec:applications}

This section highlights the multifaceted nature of federated learning research, addressing challenges related to client heterogeneity, data privacy, model security, and efficient communication, while also exploring its applicability to a wide range of real-world use cases. 

\revise{\subsection{Statistical and Model Heterogeneity}}
\revise{Variability among clients, referred to as the heterogeneity problem, stands as the principal hurdle in FL. The most common heterogeneity issues are statistical and model heterogeneity. Addressing both these two challenges is important to achieve effective FL with better personalization and generalization ability.}

\revise{The statistical heterogeneity challenge arises due to the non-IID (non-identically distributed) nature of data, where each client holds a unique subset of data, often reflecting distinct features, patterns, or statistical characteristics. This variability complicates the process of aggregating information from diverse sources to create a global model. Addressing statistical heterogeneity is crucial as it impacts the performance and generalizability of the global model, requiring specialized techniques that account for and mitigate these disparities without compromising data privacy or communication efficiency.}

\revise{\cite{li2020fedprox} proposes a local regularization approach to refine the local model of each client. Recent research efforts~\cite{FedPer,liang2020think,APFL} focus on training personalized models, amalgamating globally shared insights with personalized elements \cite{tan2021towards, jiang2020decentralized}.
Another approach involves providing multiple global models through clustering local models into distinct groups or clusters~\cite{mansour2020three,ghosh2020efficient,sattler2020clustered}. Additionally, recent advancements incorporate self-supervised learning techniques during local training to address these heterogeneity challenges~\cite{MOON, liu2021graph, yang2021interpretable}. For personalized FL, \cite{fallah2020personalized} applies meta-training strategies.}

\revise{Model heterogeneity exists in FL when there are diverse architectures, configurations, or complexities of models utilized by different clients or devices within the same system. This challenge arises because various participants may employ distinct types of machine learning models, differing in depth, structure, optimization techniques, or even hardware capabilities. Addressing model heterogeneity involves strategies to harmonize various model architectures, enabling collaborative learning while accommodating the varying computational capacities and model complexities across different devices or clients.}

\revise{Knowledge Distillation (KD)-based FL methods~\cite{lin2020ensemble,jeong2018communication,li2019fedmd,long2021federated} usually assume the inclusion of a shared toy dataset in the federated setting, allowing knowledge transfer from a teacher model to student models with differing architectures. Recent studies also explore merging neural architecture search with FL \cite{zhu2020federated,he2020fednas,singh2020differentially}, aiming to craft customized model architectures tailored to groups of clients with varying hardware capabilities and configurations. \cite{hoang2019collective} introduces a collective learning platform to handle heterogeneous architectures without accessing local training data or architectures. Moreover, functionality-based neural matching across local models aggregates neurons based on similar functionalities, irrespective of architectural differences~\cite{wang2020federated}.}

\subsection{Security and Privacy}

In the realm of federated learning, the dual concerns of security and data privacy have driven extensive research into the development of privacy-preserving solutions and the identification of novel attack strategies \cite{FLAME, splitfed, heterogeneity_robustness, robust_learning_rate, fedinv, neurotoxin}. To safeguard data privacy, recent studies have primarily focused on methods for safeguarding model parameters, thereby preventing unauthorized access to client data and its distribution by the global model. 

Notable examples include the FLAME framework \cite{FLAME}, which employs randomized and encrypted gradient vectors sent to a shuffler to protect client identities, and SplitFed \cite{splitfed}, which combines split learning and federated learning to enhance privacy while maintaining performance. From a security perspective, addressing malicious client behavior has been crucial. Robust learning rate techniques have been proposed to minimize the impact of backdoor attacks \cite{heterogeneity_robustness, robust_learning_rate}, alongside strategies like introducing data heterogeneity or using a coordinator to train updated weights before aggregation \cite{heterogeneity_robustness, fedinv}. Additionally, FedInv presents a novel approach by synthesizing a dummy dataset to mitigate Byzantine attacks effectively \cite{fedinv}. However, the Neurotoxin attack serves as a reminder of persistent threats, as it inserts enduring backdoors into federated learning systems by exploiting sparse gradient descent \cite{neurotoxin}, thereby necessitating continuous efforts to enhance security and privacy in federated learning.

Ensemble Federated Learning (EFL) employs multiple global models and label probabilities relative to the ensemble model client number to counteract the influence of malicious clients, as discussed in \cite{provably_secure_FL}. Wen et al.~ \cite{gradient_magnification} investigated attacks on federated learning that allow the central server to produce malicious parameter vectors, compromising privacy in horizontal and vertical FL settings. Proposed defense strategies include gradient clipping and noise addition.
Gupta et al.~\cite{gupta2022recovering} focused on the recovery of text information during the exchange of parameters in FL. 
To mitigate this risk, they propose a method to freeze the word embeddings of the model.
Bietti et al.~\cite{privacy_accuracy_tradeoffs} addressed the tradeoff between privacy and model accuracy by introducing Personalized-Private-SGD (PPSGD) to personalize local models while preserving privacy. Zhang et al. in \cite{clipping_for_fl} studied client-level differential privacy (DP) for federated learning, highlighting the superiority of difference clipping. Hu et al. proposed FedSPA in \cite{FedSPA}, a sparsification-based privacy mechanism. Sun et al. presented Locally Differential Private Federated Learning in \cite{LDP_FL}, focusing on adaptive range perturbation. 
Yang et al.~\cite{yang2023fedfed} explored the Gaussian or Laplacian noise to protect shared features with a differential privacy guarantee. Furthermore,  a DP protection method called FKGE~\cite{FKGE} is utilized to study the embedding of knowledge graphs in a distributed manner.
FLSG~\cite{fan2023flsg} generates random Gaussian noise with the same size of gradient and sends the most similar to the server.
Rong et al. in \cite{recommender_poisoning} explored poisoning attacks on federated recommender systems. Huang et al. in \cite{gradient_inversion_attacks} examined gradient inversion attacks and defense mechanisms. Jin et al. introduced CAFE in \cite{CAFE}, a method to recover large batch data from gradients. Sun et al. proposed FL-WBC in \cite{FLWBC}, a defense mechanism against global model poisoning. FedDefender~\cite{park2023feddefender} also focuses on the client side to achieve attack tolerance, which consists of local meta update and global distillation.
Park et al. presented Sageflow in \cite{sageflow} to handle slow devices and malicious attacks. Finally, Agarwal et al. extended differential privacy using the Skellam mechanism in \cite{skellam_mechanism}. 

Additionally, many other cryptographic methods are widely used to preserve privacy in FL. 
Chang et al.~\cite{chang2023privacy} revisited many technologies in FL and propose 2DMCFE, a functional encryption method to protect privacy under semi-honest security setting.
Furthermore, Hijazi et al.~\cite{hijazi2023secure} also investigated the use of Fully Homomorphic Encryption (FHE) in FL.
To mitigate inference attacks, Zhao et al.~\cite{zhao2022practical} proposed an effective strategy that leverages computational Diffie-Hellman (CDH) for generating lightweight keys.
These research contributions collectively advance the field of federated learning by addressing various privacy and security challenges with diverse strategies and insights.

\subsection{Communication Efficiency}

Communication efficiency is a challenging research direction in federated learning, which typically focuses on reducing the communication overhead between clients and servers, aiming to minimize data transmission and communication rounds. 
Several approaches have been proposed to enhance this aspect. Gao et al. introduced two communication-efficient distributed SGD methods in \cite{M_LSCG}, which reduce the communication cost by compressing exchanged gradients and combining local SGD with compressed gradients to the momentum technique. Wang et al. proposed FedCAMS in \cite{comm_effect_adaptive}, which combines the Federated AMSGrad adaptive gradient method with Max Stabilization and uses error feedback compression to reduce communication costs. 
\revise{GossipFL~\cite{tang2022gossipfl} uses the sparsified model to reduce communication and gossip matrix for efficient utilization of the bandwidth resources.}
Yi et al. presented the QSFL algorithm in \cite{QSFL}, which samples high-qualification clients for model updates and compresses each update to a single segment. Zhu et al.\revise{~\cite{zhu2022resilient}} addressed system heterogeneity and communication efficiency in unstable connections with the FedResCuE algorithm, focusing on the self-distillation of prunable neural networks on clients. Yapp et al. introduced the BFEL framework in \cite{BFEL}, which employs blockchain technology to reduce communication overhead by decentralizing the aggregation process. Meanwhile, Zhu et al. proposed Delayed Gradient Averaging (DGA) in \cite{DGA} to mitigate high communication latency by pipelining communication with computation. Another method called FedPM~\cite{isik2022sparse} addresses the challenge of high communication costs in federated learning by freezing weights at initial random values and learning to sparsify the random network. Finally, 
FedProg~\cite{wang2022progfed} extends the progressive learning technique from image generation to federated learning, inherently reducing computational and two-way communication costs while preserving model performance. 

In addition to the aforementioned solutions for the general federated framework, there are specialized approaches tailored to address communication challenges in specific scenarios. 
To address the communication limitation of existing federated learning-based contextual bandit algorithms, Li and Wang~\cite{li2022communication} introduced a communication-efficient framework utilizing generalized linear bandit models with online regression for local updates and offline regression for global updates. 
Especially aiming to address the communication challenge in minimax federated framework (e.g., GAN), FedGDA-GT~\cite{sun2022communication} combines gradient tacking with federated gradient descent ascent framework, showcasing linear convergence with constant stepsizes to a global-approximation solution. 
\revise{Cui et al.~\cite{cui2022optimizing} especially focus on the compute efficiency at the mobile-edge cloud computing system. Furthermore, decentralized training and deploying LLM in a federated manner also need more attention~\cite{tang2023fusionai}.}
For federated node embedding problems in graph machine learning~\cite{tan2023federated}, 
Pan and Zhu~\cite{pan2022fedwalk} proposed a random-walk-based algorithm featuring a sequence encoder for privacy preservation and a two-hop neighbor predictor, effectively reducing communication costs.

\subsection{Real-World~Applications}

Model fusion and federated X learning have yielded remarkable achievements in some real-world applications. In this subsection, we mainly summarize the applications of federated learning in two research fields, i.e., recommendation and healthcare. 

Recommendation is a practical real-world scenario. As a pioneering work, FedFast~\cite{muhammad2020fedfast} is a novel approach for accelerating federated learning of deep recommendation models. FedFast efficiently samples from a diverse set of participating clients and employs an active aggregation method, enabling users to benefit from lower communication costs and access more accurate models at the early stages of training. Liang et al.~\cite{liang2021fedrec++} proposed FedRec++, a novel lossless federated recommendation method that enhances privacy-aware preference modeling and personalization in federated recommender systems by allocating denoising clients to eliminate noise introduced by virtual ratings, ensuring accurate and privacy-preserving recommendations with minimal additional communication cost. Motivated by a similar target, Cali3F~\cite{zhu2022cali3f} is a personalized federated recommendation system training algorithm, coupled with a clustering-based aggregation method, to address privacy concerns and enhance fairness in recommendation performance across devices. To handle social recommendation scenarios, Liu et al.~\cite{liu2022federated} proposed FeSoG, a graph neural network-based federated learning recommender system. To address the challenges of heterogeneity, personalization, and privacy protection, FeSoG employs relational attention and aggregation for handling diverse data and infers user embeddings using local data to retain personalization. Different from the above works, Yuan et al.~\cite{yuan2023federated} mainly focused on user privacy and system robustness in federated recommendation systems and introduced federated recommendation unlearning (FRU) as a solution. FRU allows users to withdraw their data contributions and enhances the recommender's resistance to attacks by removing specific users' influence through historical parameter updates.

Apart from recommender systems, healthcare is another important application of federated learning~\cite{xu2019federated}. For instance, Xu et al.~\cite{xu2021privacy} introduced a federated learning approach to address the challenges of privacy in diagnosing depression, proposing a multi-view federated learning framework with multi-source data and later fusion methods to handle inconsistent time series data. Similarly, Che et al.~\cite{che2022federated} addressed the challenges of data privacy and heterogeneity in medical data by preventing leakage in multi-view scenarios. Aiming at the heterogeneous challenge in smart healthcare, Liu et al.~\cite{liu2022contribution} presented CAFL, an effective method for impartially assessing participants' contributions to federated learning model performance without compromising their private data. To address label noise challenges in medical imaging federated learning, FedGP~\cite{chen2023medical} provides reliable pseudo labels through noisy graph purification on the client side and utilizing a graph-guided negative ensemble loss for robust supervision against label noise. To address the weakly supervised problem in medical image segmentation, FedDM~\cite{zhu2023feddm} tackles local drift with collaborative annotation calibration for label correction and global drift with hierarchical gradient de-conflicting for robust gradient aggregation respectively. 

Federated learning also finds applications in various and diverse scenarios, such as image steganalysis~\cite{yang2020fedsteg}, open banking~\cite{long2020federated}, and mobile keyboard suggestion~\cite{mcmahan2017communication,ji2019learning}. Anticipated are broader applications to be practically implemented within the federated setting.

\section{Future~Directions}
\label{sec:future-directions}
In recent years, federated learning has seen drastic growth in terms of the amount of research and the breadth of topics.
There is still a need for studies on the following promising directions. 

\paragraph{Label Scarcity}
\revise{Current federated learning heavily relies on the supervision signals from sufficient training labels. }
However, in most real-world applications, clients may not have sufficient labels or lack interaction between users to provide interactive labels. 
The label scarcity problem makes federated learning impractical in many scenarios. 
\revise{In this case, a potential research direction is to consider the label deficiency while keeping private data on-device. 
To achieve this research objective, comprehensive investigations into federated learning incorporating semi-supervised learning, transfer learning, few-shot learning, and meta learning are warranted. This holistic approach not only mitigates the impact of label scarcity but also opens avenues for more versatile and adaptive federated learning models that can better accommodate the intricacies of real-world scenarios.}

\paragraph{On-device~Personalization}
Conventionally, personalization is achieved by additional fine-tuning before inference. Recently, more research has focused on personalization. On-device personalization~\cite{wang2019federated} brings forward multiple possible scenarios where clients would additionally benefit from personalization. Mansour et al.~\cite{mansour2020three} formulated their approaches for personalization, including user clustering, data interpolation, and model interpolation. Model-agnostic meta-learning aims to learn quick adaptations and also brings the potential to personalize to individual devices. The studies of effective formulation and metrics to evaluate personalized performance are missed. The underlying essence of personalization and the connections between global model learning and personalized on-device training should be addressed.

\paragraph{Unsupervised~Learning} 
\revise{The majority of current research on federated learning mainly follows the supervised or semi-supervised paradigms. 
Due to the label deficiency problem in the real-world scenario}, unsupervised representation learning can be the future direction in the federated setting and other learning problems.
\revise{By forgoing the need for explicit labels, the unsupervised federated learning methods can autonomously decipher intricate data patterns across distributed datasets. Potential unsupervised techniques include autoencoders, GANs, and clustering algorithms. These approaches enable federated learning systems to extract meaningful features and/or model data manifold without relying on labeled data, addressing the label scarcity issue in real-world scenarios. Furthermore, federated self-supervised learning can also be a promising avenue for overcoming data scarcity issues in federated settings. By leveraging the inherent structures within the data itself, federated self-supervised learning techniques empower devices to learn from their local data without requiring explicit labels from a central server. }

\revise{\paragraph{Collaboration of Multiple Federated Paradigms}
Federated Learning, as a novel training paradigm, presents numerous new challenges that require attention. In most scenarios, the collaboration of various techniques within the FL framework is necessary. For instance, knowledge distillation shows promising potential in overcoming many challenges through the transfer of abstract knowledge, such as addressing heterogeneity and facilitating multi-task learning. Additionally, exploring the application of transfer learning for knowledge reuse under federated learning is meaningful. This approach can improve data utilization and effectively reduce repeated training in federated scenarios with label scarcity or reinforcement learning. Therefore, we suggest studying how multiple federated learning paradigms can work together to address both existing and new challenges.
}

\revise{\paragraph{Comprehensive Benchmark}
Among the numerous federated learning algorithms in the literature, it is evident that federated learning encompasses various parameters, reflecting diverse scenarios, different data distributions of edge side, and various communication frequencies. However, existing research often evaluates these algorithms in different settings, hindering researchers from seeking suitable methods for their specific tasks. Therefore, the establishment of a unified benchmark becomes imperative. There are also some infrastructures to speed up algorithm implementations like FedML~\cite{chaoyanghe2020fedml, tang2023fedml}. This endeavor aims to inspire greater research in federated learning while providing comprehensive benchmarks adhering to standardized criteria. These benchmarks may encompass real-world deployment scenarios, algorithm comparisons across diverse data environments, and intriguing evaluations of foundation models combined with federated learning.
}

\revise{\paragraph{Production-Level Federated Learning}
In the world of federated learning, it is crucial to shift focus towards making it work effectively in real-world production-level settings. Researchers should aim to improve how federated learning can be used practically. This means finding ways to make it easily fit into existing systems, handle differences between devices, and cope with limitations in communication. It is also important to handle unique real-world challenges such as data distribution drift, diurnal variations, and cold start problems~\cite{li2020federated}. To address these challenges, the implementation of federated X learning holds significant promise for providing viable solutions. For instance, federated transfer learning proves effective in managing distribution drift, while federated meta learning serves as a valuable tool for addressing cold start problems. In the future,  more advanced federated X learning methods specifically designed for production-level applications are expected.}

\section{Conclusion}
This paper conducts a timely and focused survey about federated learning coupled with different learning algorithms.
The flexibility of FL was showcased by presenting a wide range of relevant learning paradigms that can be employed within the FL framework. 
In particular, the compatibility was addressed from the standpoint of how learning algorithms fit the FL architecture and how they take into account two of the critical problems in federated learning: efficient learning and statistical heterogeneity.

\bibliography{survey-fed}%


\begin{thebibliography}{208}
\ifx \bisbn   \undefined \def \bisbn  #1{ISBN #1}\fi
\ifx \binits  \undefined \def \binits#1{#1}\fi
\ifx \bauthor  \undefined \def \bauthor#1{#1}\fi
\ifx \batitle  \undefined \def \batitle#1{#1}\fi
\ifx \bjtitle  \undefined \def \bjtitle#1{#1}\fi
\ifx \bvolume  \undefined \def \bvolume#1{\textbf{#1}}\fi
\ifx \byear  \undefined \def \byear#1{#1}\fi
\ifx \bissue  \undefined \def \bissue#1{#1}\fi
\ifx \bfpage  \undefined \def \bfpage#1{#1}\fi
\ifx \blpage  \undefined \def \blpage #1{#1}\fi
\ifx \burl  \undefined \def \burl#1{\textsf{#1}}\fi
\ifx \doiurl  \undefined \def \doiurl#1{\url{https://doi.org/#1}}\fi
\ifx \betal  \undefined \def \betal{\textit{et al.}}\fi
\ifx \binstitute  \undefined \def \binstitute#1{#1}\fi
\ifx \binstitutionaled  \undefined \def \binstitutionaled#1{#1}\fi
\ifx \bctitle  \undefined \def \bctitle#1{#1}\fi
\ifx \beditor  \undefined \def \beditor#1{#1}\fi
\ifx \bpublisher  \undefined \def \bpublisher#1{#1}\fi
\ifx \bbtitle  \undefined \def \bbtitle#1{#1}\fi
\ifx \bedition  \undefined \def \bedition#1{#1}\fi
\ifx \bseriesno  \undefined \def \bseriesno#1{#1}\fi
\ifx \blocation  \undefined \def \blocation#1{#1}\fi
\ifx \bsertitle  \undefined \def \bsertitle#1{#1}\fi
\ifx \bsnm \undefined \def \bsnm#1{#1}\fi
\ifx \bsuffix \undefined \def \bsuffix#1{#1}\fi
\ifx \bparticle \undefined \def \bparticle#1{#1}\fi
\ifx \barticle \undefined \def \barticle#1{#1}\fi
\bibcommenthead
\ifx \bconfdate \undefined \def \bconfdate #1{#1}\fi
\ifx \botherref \undefined \def \botherref #1{#1}\fi
\ifx \url \undefined \def \url#1{\textsf{#1}}\fi
\ifx \bchapter \undefined \def \bchapter#1{#1}\fi
\ifx \bbook \undefined \def \bbook#1{#1}\fi
\ifx \bcomment \undefined \def \bcomment#1{#1}\fi
\ifx \oauthor \undefined \def \oauthor#1{#1}\fi
\ifx \citeauthoryear \undefined \def \citeauthoryear#1{#1}\fi
\ifx \endbibitem  \undefined \def \endbibitem {}\fi
\ifx \bconflocation  \undefined \def \bconflocation#1{#1}\fi
\ifx \arxivurl  \undefined \def \arxivurl#1{\textsf{#1}}\fi
\csname PreBibitemsHook\endcsname

\bibitem[\protect\citeauthoryear{McMahan
  et~al.}{2017}]{mcmahan2017communication}
\begin{bchapter}
\bauthor{\bsnm{McMahan}, \binits{H.B.}},
\bauthor{\bsnm{Moore}, \binits{E.}},
\bauthor{\bsnm{Ramage}, \binits{D.}},
\bauthor{\bsnm{Hampson}, \binits{S.}}, \betal:
\bctitle{Communication-efficient learning of deep networks from decentralized
  data}.
In: \bbtitle{International Conference on Artificial Intelligence and
  Statistics},
pp. \bfpage{1273}--\blpage{1282}
(\byear{2017})
\end{bchapter}
\endbibitem

\bibitem[\protect\citeauthoryear{Wang et~al.}{2020}]{wang2020attack}
\begin{barticle}
\bauthor{\bsnm{Wang}, \binits{H.}},
\bauthor{\bsnm{Sreenivasan}, \binits{K.}},
\bauthor{\bsnm{Rajput}, \binits{S.}},
\bauthor{\bsnm{Vishwakarma}, \binits{H.}},
\bauthor{\bsnm{Agarwal}, \binits{S.}},
\bauthor{\bsnm{Sohn}, \binits{J.-y.}},
\bauthor{\bsnm{Lee}, \binits{K.}},
\bauthor{\bsnm{Papailiopoulos}, \binits{D.}}:
\batitle{Attack of the tails: Yes, you really can backdoor federated learning}.
\bjtitle{Advances in Neural Information Processing Systems}
\bvolume{33},
\bfpage{16070}--\blpage{16084}
(\byear{2020})
\end{barticle}
\endbibitem

\bibitem[\protect\citeauthoryear{Bagdasaryan
  et~al.}{2020}]{bagdasaryan2020backdoor}
\begin{bchapter}
\bauthor{\bsnm{Bagdasaryan}, \binits{E.}},
\bauthor{\bsnm{Veit}, \binits{A.}},
\bauthor{\bsnm{Hua}, \binits{Y.}},
\bauthor{\bsnm{Estrin}, \binits{D.}},
\bauthor{\bsnm{Shmatikov}, \binits{V.}}:
\bctitle{How to backdoor federated learning}.
In: \bbtitle{International Conference on Artificial Intelligence and
  Statistics},
pp. \bfpage{2938}--\blpage{2948}
(\byear{2020}).
\bcomment{PMLR}
\end{bchapter}
\endbibitem

\bibitem[\protect\citeauthoryear{Tolpegin et~al.}{2020}]{tolpegin2020data}
\begin{bchapter}
\bauthor{\bsnm{Tolpegin}, \binits{V.}},
\bauthor{\bsnm{Truex}, \binits{S.}},
\bauthor{\bsnm{Gursoy}, \binits{M.E.}},
\bauthor{\bsnm{Liu}, \binits{L.}}:
\bctitle{Data poisoning attacks against federated learning systems}.
In: \bbtitle{Computer Security--ESORICS 2020: 25th European Symposium on
  Research in Computer Security, ESORICS 2020, Guildford, UK, September 14--18,
  2020, Proceedings, Part I 25},
pp. \bfpage{480}--\blpage{501}
(\byear{2020}).
\bcomment{Springer}
\end{bchapter}
\endbibitem

\bibitem[\protect\citeauthoryear{Kone{\v{c}}n{\`y}
  et~al.}{2016}]{konevcny2016federated}
\begin{botherref}
\oauthor{\bsnm{Kone{\v{c}}n{\`y}}, \binits{J.}},
\oauthor{\bsnm{McMahan}, \binits{H.B.}},
\oauthor{\bsnm{Yu}, \binits{F.X.}},
\oauthor{\bsnm{Richt{\'a}rik}, \binits{P.}},
\oauthor{\bsnm{Suresh}, \binits{A.T.}},
\oauthor{\bsnm{Bacon}, \binits{D.}}:
Federated learning: Strategies for improving communication efficiency.
arXiv preprint arXiv:1610.05492
(2016)
\end{botherref}
\endbibitem

\bibitem[\protect\citeauthoryear{Ji et~al.}{2020}]{ji2020dynamic}
\begin{botherref}
\oauthor{\bsnm{Ji}, \binits{S.}},
\oauthor{\bsnm{Jiang}, \binits{W.}},
\oauthor{\bsnm{Walid}, \binits{A.}},
\oauthor{\bsnm{Li}, \binits{X.}}:
Dynamic sampling and selective masking for communication-efficient federated
  learning.
arXiv preprint arXiv:2003.09603
(2020)
\end{botherref}
\endbibitem

\bibitem[\protect\citeauthoryear{Tan et~al.}{2022}]{tan2022federated}
\begin{bchapter}
\bauthor{\bsnm{Tan}, \binits{Y.}},
\bauthor{\bsnm{Long}, \binits{G.}},
\bauthor{\bsnm{Ma}, \binits{J.}},
\bauthor{\bsnm{Liu}, \binits{L.}},
\bauthor{\bsnm{Zhou}, \binits{T.}},
\bauthor{\bsnm{Jiang}, \binits{J.}}:
\bctitle{Federated learning from pre-trained models: A contrastive learning
  approach}.
In: \bbtitle{Advances in Neural Information Processing Systems},
vol. \bseriesno{35},
pp. \bfpage{19332}--\blpage{19344}
(\byear{2022})
\end{bchapter}
\endbibitem

\bibitem[\protect\citeauthoryear{He et~al.}{2019}]{he2019central}
\begin{botherref}
\oauthor{\bsnm{He}, \binits{C.}},
\oauthor{\bsnm{Tan}, \binits{C.}},
\oauthor{\bsnm{Tang}, \binits{H.}},
\oauthor{\bsnm{Qiu}, \binits{S.}},
\oauthor{\bsnm{Liu}, \binits{J.}}:
Central server free federated learning over single-sided trust social networks.
arXiv preprint arXiv:1910.04956
(2019)
\end{botherref}
\endbibitem

\bibitem[\protect\citeauthoryear{Yeganeh et~al.}{2020}]{yeganeh2020inverse}
\begin{bchapter}
\bauthor{\bsnm{Yeganeh}, \binits{Y.}},
\bauthor{\bsnm{Farshad}, \binits{A.}},
\bauthor{\bsnm{Navab}, \binits{N.}},
\bauthor{\bsnm{Albarqouni}, \binits{S.}}:
\bctitle{Inverse distance aggregation for federated learning with non-iid
  data}.
In: \bbtitle{DCL Workshop at MICCAI},
pp. \bfpage{150}--\blpage{159}
(\byear{2020})
\end{bchapter}
\endbibitem

\bibitem[\protect\citeauthoryear{Chen et~al.}{2020}]{chen2020communication}
\begin{botherref}
\oauthor{\bsnm{Chen}, \binits{Y.}},
\oauthor{\bsnm{Sun}, \binits{X.}},
\oauthor{\bsnm{Jin}, \binits{Y.}}:
Communication-efficient federated deep learning with layerwise asynchronous
  model update and temporally weighted aggregation.
IEEE Transactions on Neural Networks and Learning Systems
(2020)
\end{botherref}
\endbibitem

\bibitem[\protect\citeauthoryear{Ji et~al.}{2019}]{ji2019learning}
\begin{bchapter}
\bauthor{\bsnm{Ji}, \binits{S.}},
\bauthor{\bsnm{Pan}, \binits{S.}},
\bauthor{\bsnm{Long}, \binits{G.}},
\bauthor{\bsnm{Li}, \binits{X.}},
\bauthor{\bsnm{Jiang}, \binits{J.}},
\bauthor{\bsnm{Huang}, \binits{Z.}}:
\bctitle{Learning private neural language modeling with attentive aggregation}.
In: \bbtitle{International Joint Conference on Neural Network}
(\byear{2019})
\end{bchapter}
\endbibitem

\bibitem[\protect\citeauthoryear{Li et~al.}{2020}]{li2020federated}
\begin{barticle}
\bauthor{\bsnm{Li}, \binits{T.}},
\bauthor{\bsnm{Sahu}, \binits{A.K.}},
\bauthor{\bsnm{Talwalkar}, \binits{A.}},
\bauthor{\bsnm{Smith}, \binits{V.}}:
\batitle{Federated learning: Challenges, methods, and future directions}.
\bjtitle{IEEE Signal Processing Magazine}
\bvolume{37}(\bissue{3}),
\bfpage{50}--\blpage{60}
(\byear{2020})
\end{barticle}
\endbibitem

\bibitem[\protect\citeauthoryear{Briggs et~al.}{2020}]{briggs2020federated}
\begin{bchapter}
\bauthor{\bsnm{Briggs}, \binits{C.}},
\bauthor{\bsnm{Fan}, \binits{Z.}},
\bauthor{\bsnm{Andras}, \binits{P.}}:
\bctitle{Federated learning with hierarchical clustering of local updates to
  improve training on non-iid data}.
In: \bbtitle{International Joint Conference on Neural Network}
(\byear{2020})
\end{bchapter}
\endbibitem

\bibitem[\protect\citeauthoryear{Yurochkin
  et~al.}{2019}]{yurochkin2019bayesian}
\begin{bchapter}
\bauthor{\bsnm{Yurochkin}, \binits{M.}},
\bauthor{\bsnm{Agarwal}, \binits{M.}},
\bauthor{\bsnm{Ghosh}, \binits{S.}},
\bauthor{\bsnm{Greenewald}, \binits{K.}},
\bauthor{\bsnm{Hoang}, \binits{N.}},
\bauthor{\bsnm{Khazaeni}, \binits{Y.}}:
\bctitle{Bayesian nonparametric federated learning of neural networks}.
In: \bbtitle{International Conference on Machine Learning},
pp. \bfpage{7252}--\blpage{7261}
(\byear{2019})
\end{bchapter}
\endbibitem

\bibitem[\protect\citeauthoryear{Li et~al.}{2020}]{li2020fair}
\begin{bchapter}
\bauthor{\bsnm{Li}, \binits{T.}},
\bauthor{\bsnm{Sanjabi}, \binits{M.}},
\bauthor{\bsnm{Beirami}, \binits{A.}},
\bauthor{\bsnm{Smith}, \binits{V.}}:
\bctitle{Fair resource allocation in federated learning}.
In: \bbtitle{International Conference on Learning Representations}
(\byear{2020})
\end{bchapter}
\endbibitem

\bibitem[\protect\citeauthoryear{Xie et~al.}{2022}]{xie2022robust}
\begin{botherref}
\oauthor{\bsnm{Xie}, \binits{Y.}},
\oauthor{\bsnm{Zhang}, \binits{W.}},
\oauthor{\bsnm{Pi}, \binits{R.}},
\oauthor{\bsnm{Wu}, \binits{F.}},
\oauthor{\bsnm{Chen}, \binits{Q.}},
\oauthor{\bsnm{Xie}, \binits{X.}},
\oauthor{\bsnm{Kim}, \binits{S.}}:
Robust federated learning against both data heterogeneity and poisoning attack
  via aggregation optimization.
arXiv preprint
(2022)
\end{botherref}
\endbibitem

\bibitem[\protect\citeauthoryear{Xiao et~al.}{2021}]{xiao2021novel}
\begin{bchapter}
\bauthor{\bsnm{Xiao}, \binits{J.}},
\bauthor{\bsnm{Du}, \binits{C.}},
\bauthor{\bsnm{Duan}, \binits{Z.}},
\bauthor{\bsnm{Guo}, \binits{W.}}:
\bctitle{A novel server-side aggregation strategy for federated learning in
  non-iid situations}.
In: \bbtitle{2021 20th International Symposium on Parallel and Distributed
  Computing (ISPDC)},
pp. \bfpage{17}--\blpage{24}
(\byear{2021}).
\bcomment{IEEE}
\end{bchapter}
\endbibitem

\bibitem[\protect\citeauthoryear{Liu et~al.}{2021}]{liu2021fedpa}
\begin{barticle}
\bauthor{\bsnm{Liu}, \binits{J.}},
\bauthor{\bsnm{Wang}, \binits{J.H.}},
\bauthor{\bsnm{Rong}, \binits{C.}},
\bauthor{\bsnm{Xu}, \binits{Y.}},
\bauthor{\bsnm{Yu}, \binits{T.}},
\bauthor{\bsnm{Wang}, \binits{J.}}:
\batitle{Fedpa: An adaptively partial model aggregation strategy in federated
  learning}.
\bjtitle{Computer Networks}
\bvolume{199},
\bfpage{108468}
(\byear{2021})
\end{barticle}
\endbibitem

\bibitem[\protect\citeauthoryear{Jiang et~al.}{2020}]{jiang2020decentralized}
\begin{botherref}
\oauthor{\bsnm{Jiang}, \binits{J.}},
\oauthor{\bsnm{Ji}, \binits{S.}},
\oauthor{\bsnm{Long}, \binits{G.}}:
Decentralized knowledge acquisition for mobile internet applications.
World Wide Web
(2020)
\end{botherref}
\endbibitem

\bibitem[\protect\citeauthoryear{Wu et~al.}{2020}]{wu2020fedmed}
\begin{barticle}
\bauthor{\bsnm{Wu}, \binits{X.}},
\bauthor{\bsnm{Liang}, \binits{Z.}},
\bauthor{\bsnm{Wang}, \binits{J.}}:
\batitle{{FedMed: A federated learning framework for language modeling}}.
\bjtitle{Sensors}
\bvolume{20}(\bissue{14}),
\bfpage{4048}
(\byear{2020})
\end{barticle}
\endbibitem

\bibitem[\protect\citeauthoryear{Huang et~al.}{2021}]{huang2021personalized}
\begin{bchapter}
\bauthor{\bsnm{Huang}, \binits{Y.}},
\bauthor{\bsnm{Chu}, \binits{L.}},
\bauthor{\bsnm{Zhou}, \binits{Z.}},
\bauthor{\bsnm{Wang}, \binits{L.}},
\bauthor{\bsnm{Liu}, \binits{J.}},
\bauthor{\bsnm{Pei}, \binits{J.}},
\bauthor{\bsnm{Zhang}, \binits{Y.}}:
\bctitle{Personalized cross-silo federated learning on non-iid data}.
In: \bbtitle{AAAI Conference on Artificial Intelligence}
(\byear{2021})
\end{bchapter}
\endbibitem

\bibitem[\protect\citeauthoryear{Wang et~al.}{2020}]{wang2020attention}
\begin{barticle}
\bauthor{\bsnm{Wang}, \binits{X.}},
\bauthor{\bsnm{Li}, \binits{R.}},
\bauthor{\bsnm{Wang}, \binits{C.}},
\bauthor{\bsnm{Li}, \binits{X.}},
\bauthor{\bsnm{Taleb}, \binits{T.}},
\bauthor{\bsnm{Leung}, \binits{V.C.}}:
\batitle{Attention-weighted federated deep reinforcement learning for
  device-to-device assisted heterogeneous collaborative edge caching}.
\bjtitle{IEEE Journal on Selected Areas in Communications}
\bvolume{39}(\bissue{1}),
\bfpage{154}--\blpage{169}
(\byear{2020})
\end{barticle}
\endbibitem

\bibitem[\protect\citeauthoryear{Guo et~al.}{2023}]{guo2023fedmcsa}
\begin{barticle}
\bauthor{\bsnm{Guo}, \binits{Q.}},
\bauthor{\bsnm{Qi}, \binits{Y.}},
\bauthor{\bsnm{Qi}, \binits{S.}},
\bauthor{\bsnm{Wu}, \binits{D.}},
\bauthor{\bsnm{Li}, \binits{Q.}}:
\batitle{Fedmcsa: Personalized federated learning via model components
  self-attention}.
\bjtitle{Neurocomputing}
\bvolume{560},
\bfpage{126831}
(\byear{2023})
\end{barticle}
\endbibitem

\bibitem[\protect\citeauthoryear{Zheng et~al.}{2022}]{zheng2022channelfed}
\begin{bchapter}
\bauthor{\bsnm{Zheng}, \binits{K.}},
\bauthor{\bsnm{Liu}, \binits{X.}},
\bauthor{\bsnm{Zhu}, \binits{G.}},
\bauthor{\bsnm{Wu}, \binits{X.}},
\bauthor{\bsnm{Niu}, \binits{J.}}:
\bctitle{{ChannelFed}: Enabling personalized federated learning via localized
  channel attention}.
In: \bbtitle{GLOBECOM 2022-2022 IEEE Global Communications Conference},
pp. \bfpage{2987}--\blpage{2992}
(\byear{2022}).
\bcomment{IEEE}
\end{bchapter}
\endbibitem

\bibitem[\protect\citeauthoryear{Yu et~al.}{2020}]{yu2020federated}
\begin{bchapter}
\bauthor{\bsnm{Yu}, \binits{F.X.}},
\bauthor{\bsnm{Rawat}, \binits{A.S.}},
\bauthor{\bsnm{Menon}, \binits{A.K.}},
\bauthor{\bsnm{Kumar}, \binits{S.}}:
\bctitle{Federated learning with only positive labels}.
In: \bbtitle{International Conference on Machine Learning}
(\byear{2020})
\end{bchapter}
\endbibitem

\bibitem[\protect\citeauthoryear{Li et~al.}{2020}]{li2020fedprox}
\begin{bchapter}
\bauthor{\bsnm{Li}, \binits{T.}},
\bauthor{\bsnm{Sahu}, \binits{A.K.}},
\bauthor{\bsnm{Zaheer}, \binits{M.}},
\bauthor{\bsnm{Sanjabi}, \binits{M.}},
\bauthor{\bsnm{Talwalkar}, \binits{A.}},
\bauthor{\bsnm{Smith}, \binits{V.}}:
\bctitle{Federated optimization in heterogeneous networks}.
In: \bbtitle{Conference on Machine Learning and Systems}
(\byear{2020})
\end{bchapter}
\endbibitem

\bibitem[\protect\citeauthoryear{Karimireddy
  et~al.}{2020}]{karimireddy2020mime}
\begin{botherref}
\oauthor{\bsnm{Karimireddy}, \binits{S.P.}},
\oauthor{\bsnm{Jaggi}, \binits{M.}},
\oauthor{\bsnm{Kale}, \binits{S.}},
\oauthor{\bsnm{Mohri}, \binits{M.}},
\oauthor{\bsnm{Reddi}, \binits{S.J.}},
\oauthor{\bsnm{Stich}, \binits{S.U.}},
\oauthor{\bsnm{Suresh}, \binits{A.T.}}:
Mime: Mimicking centralized stochastic algorithms in federated learning.
arXiv preprint arXiv:2008.03606
(2020)
\end{botherref}
\endbibitem

\bibitem[\protect\citeauthoryear{Durmus et~al.}{2021}]{durmus2021federated}
\begin{bchapter}
\bauthor{\bsnm{Durmus}, \binits{A.E.}},
\bauthor{\bsnm{Yue}, \binits{Z.}},
\bauthor{\bsnm{Ramon}, \binits{M.}},
\bauthor{\bsnm{Matthew}, \binits{M.}},
\bauthor{\bsnm{Paul}, \binits{W.}},
\bauthor{\bsnm{Venkatesh}, \binits{S.}}:
\bctitle{Federated learning based on dynamic regularization}.
In: \bbtitle{International Conference on Learning Representations}
(\byear{2021})
\end{bchapter}
\endbibitem

\bibitem[\protect\citeauthoryear{Kim et~al.}{2022}]{kim2022multi}
\begin{bchapter}
\bauthor{\bsnm{Kim}, \binits{J.}},
\bauthor{\bsnm{Kim}, \binits{G.}},
\bauthor{\bsnm{Han}, \binits{B.}}:
\bctitle{Multi-level branched regularization for federated learning}.
In: \bbtitle{International Conference on Machine Learning},
pp. \bfpage{11058}--\blpage{11073}
(\byear{2022}).
\bcomment{PMLR}
\end{bchapter}
\endbibitem

\bibitem[\protect\citeauthoryear{Cheng et~al.}{2022}]{cheng2022differentially}
\begin{bchapter}
\bauthor{\bsnm{Cheng}, \binits{A.}},
\bauthor{\bsnm{Wang}, \binits{P.}},
\bauthor{\bsnm{Zhang}, \binits{X.S.}},
\bauthor{\bsnm{Cheng}, \binits{J.}}:
\bctitle{Differentially private federated learning with local regularization
  and sparsification}.
In: \bbtitle{Proceedings of the IEEE/CVF Conference on Computer Vision and
  Pattern Recognition},
pp. \bfpage{10122}--\blpage{10131}
(\byear{2022})
\end{bchapter}
\endbibitem

\bibitem[\protect\citeauthoryear{Chen et~al.}{2023}]{chen2023workie}
\begin{bchapter}
\bauthor{\bsnm{Chen}, \binits{R.}},
\bauthor{\bsnm{Wan}, \binits{Q.}},
\bauthor{\bsnm{Prakash}, \binits{P.}},
\bauthor{\bsnm{Zhang}, \binits{L.}},
\bauthor{\bsnm{Yuan}, \binits{X.}},
\bauthor{\bsnm{Gong}, \binits{Y.}},
\bauthor{\bsnm{Fu}, \binits{X.}},
\bauthor{\bsnm{Pan}, \binits{M.}}:
\bctitle{Workie-talkie: Accelerating federated learning by overlapping
  computing and communications via contrastive regularization}.
In: \bbtitle{Proceedings of the IEEE/CVF International Conference on Computer
  Vision},
pp. \bfpage{16999}--\blpage{17009}
(\byear{2023})
\end{bchapter}
\endbibitem

\bibitem[\protect\citeauthoryear{Dinh et~al.}{2022}]{dinh2022new}
\begin{botherref}
\oauthor{\bsnm{Dinh}, \binits{C.T.}},
\oauthor{\bsnm{Vu}, \binits{T.T.}},
\oauthor{\bsnm{Tran}, \binits{N.H.}},
\oauthor{\bsnm{Dao}, \binits{M.N.}},
\oauthor{\bsnm{Zhang}, \binits{H.}}:
A new look and convergence rate of federated multitask learning with laplacian
  regularization.
IEEE Transactions on Neural Networks and Learning Systems
(2022)
\end{botherref}
\endbibitem

\bibitem[\protect\citeauthoryear{Tan et~al.}{2022}]{fedproto}
\begin{barticle}
\bauthor{\bsnm{Tan}, \binits{Y.}},
\bauthor{\bsnm{Long}, \binits{G.}},
\bauthor{\bsnm{LIU}, \binits{L.}},
\bauthor{\bsnm{Zhou}, \binits{T.}},
\bauthor{\bsnm{Lu}, \binits{Q.}},
\bauthor{\bsnm{Jiang}, \binits{J.}},
\bauthor{\bsnm{Zhang}, \binits{C.}}:
\batitle{Fedproto: Federated prototype learning across heterogeneous clients}.
\bjtitle{Proceedings of the AAAI Conference on Artificial Intelligence}
\bvolume{36}(\bissue{8}),
\bfpage{8432}--\blpage{8440}
(\byear{2022})
\doiurl{10.1609/aaai.v36i8.20819}
\end{barticle}
\endbibitem

\bibitem[\protect\citeauthoryear{Ghosh et~al.}{2020}]{ghosh2020efficient}
\begin{bchapter}
\bauthor{\bsnm{Ghosh}, \binits{A.}},
\bauthor{\bsnm{Chung}, \binits{J.}},
\bauthor{\bsnm{Yin}, \binits{D.}},
\bauthor{\bsnm{Ramchandran}, \binits{K.}}:
\bctitle{An efficient framework for clustered federated learning}.
In: \bbtitle{Advances in Neural Information Processing Systems}
(\byear{2020})
\end{bchapter}
\endbibitem

\bibitem[\protect\citeauthoryear{Long et~al.}{2023}]{xie2020multicenter}
\begin{barticle}
\bauthor{\bsnm{Long}, \binits{G.}},
\bauthor{\bsnm{Xie}, \binits{M.}},
\bauthor{\bsnm{Shen}, \binits{T.}},
\bauthor{\bsnm{Zhou}, \binits{T.}},
\bauthor{\bsnm{Wang}, \binits{X.}},
\bauthor{\bsnm{Jiang}, \binits{J.}}:
\batitle{Multi-center federated learning: clients clustering for better
  personalization}.
\bjtitle{World Wide Web}
\bvolume{26}(\bissue{1}),
\bfpage{481}--\blpage{500}
(\byear{2023})
\end{barticle}
\endbibitem

\bibitem[\protect\citeauthoryear{Muhammad et~al.}{2020}]{muhammad2020fedfast}
\begin{bchapter}
\bauthor{\bsnm{Muhammad}, \binits{K.}},
\bauthor{\bsnm{Wang}, \binits{Q.}},
\bauthor{\bsnm{O'Reilly-Morgan}, \binits{D.}},
\bauthor{\bsnm{Tragos}, \binits{E.}},
\bauthor{\bsnm{Smyth}, \binits{B.}},
\bauthor{\bsnm{Hurley}, \binits{N.}},
\bauthor{\bsnm{Geraci}, \binits{J.}},
\bauthor{\bsnm{Lawlor}, \binits{A.}}:
\bctitle{{FedFast: Going beyond average for faster training of federated
  recommender systems}}.
In: \bbtitle{SIGKDD},
pp. \bfpage{1234}--\blpage{1242}
(\byear{2020})
\end{bchapter}
\endbibitem

\bibitem[\protect\citeauthoryear{Dennis et~al.}{2021}]{dennis2021heterogeneity}
\begin{bchapter}
\bauthor{\bsnm{Dennis}, \binits{D.K.}},
\bauthor{\bsnm{Li}, \binits{T.}},
\bauthor{\bsnm{Smith}, \binits{V.}}:
\bctitle{Heterogeneity for the win: One-shot federated clustering}.
In: \bbtitle{International Conference on Machine Learning},
pp. \bfpage{2611}--\blpage{2620}
(\byear{2021}).
\bcomment{PMLR}
\end{bchapter}
\endbibitem

\bibitem[\protect\citeauthoryear{Chung et~al.}{2022}]{chung2022federated}
\begin{bchapter}
\bauthor{\bsnm{Chung}, \binits{J.}},
\bauthor{\bsnm{Lee}, \binits{K.}},
\bauthor{\bsnm{Ramchandran}, \binits{K.}}:
\bctitle{Federated unsupervised clustering with generative models}.
In: \bbtitle{AAAI 2022 International Workshop on Trustable, Verifiable and
  Auditable Federated Learning}
(\byear{2022})
\end{bchapter}
\endbibitem

\bibitem[\protect\citeauthoryear{Cai et~al.}{2023}]{cai2023fedce}
\begin{bchapter}
\bauthor{\bsnm{Cai}, \binits{L.}},
\bauthor{\bsnm{Chen}, \binits{N.}},
\bauthor{\bsnm{Cao}, \binits{Y.}},
\bauthor{\bsnm{He}, \binits{J.}},
\bauthor{\bsnm{Li}, \binits{Y.}}:
\bctitle{{FedCE}: Personalized federated learning method based on clustering
  ensembles}.
In: \bbtitle{Proceedings of the 31st ACM International Conference on
  Multimedia},
pp. \bfpage{1625}--\blpage{1633}
(\byear{2023})
\end{bchapter}
\endbibitem

\bibitem[\protect\citeauthoryear{Wang et~al.}{2020}]{wang2020federated}
\begin{bchapter}
\bauthor{\bsnm{Wang}, \binits{H.}},
\bauthor{\bsnm{Yurochkin}, \binits{M.}},
\bauthor{\bsnm{Sun}, \binits{Y.}},
\bauthor{\bsnm{Papailiopoulos}, \binits{D.}},
\bauthor{\bsnm{Khazaeni}, \binits{Y.}}:
\bctitle{Federated learning with matched averaging}.
In: \bbtitle{International Conference on Learning Representations}
(\byear{2020})
\end{bchapter}
\endbibitem

\bibitem[\protect\citeauthoryear{Chen and Chao}{2020}]{chen2020fedbe}
\begin{bchapter}
\bauthor{\bsnm{Chen}, \binits{H.-Y.}},
\bauthor{\bsnm{Chao}, \binits{W.-L.}}:
\bctitle{{FedBE}: Making bayesian model ensemble applicable to federated
  learning}.
In: \bbtitle{International Conference on Learning Representations}
(\byear{2020})
\end{bchapter}
\endbibitem

\bibitem[\protect\citeauthoryear{Zhang et~al.}{2022}]{zhang2022personalized}
\begin{bchapter}
\bauthor{\bsnm{Zhang}, \binits{X.}},
\bauthor{\bsnm{Li}, \binits{Y.}},
\bauthor{\bsnm{Li}, \binits{W.}},
\bauthor{\bsnm{Guo}, \binits{K.}},
\bauthor{\bsnm{Shao}, \binits{Y.}}:
\bctitle{Personalized federated learning via variational bayesian inference}.
In: \bbtitle{International Conference on Machine Learning},
pp. \bfpage{26293}--\blpage{26310}
(\byear{2022}).
\bcomment{PMLR}
\end{bchapter}
\endbibitem

\bibitem[\protect\citeauthoryear{Xiao and Cheng}{2023}]{xiao2023bayesian}
\begin{bchapter}
\bauthor{\bsnm{Xiao}, \binits{P.}},
\bauthor{\bsnm{Cheng}, \binits{S.}}:
\bctitle{Bayesian federated neural matching that completes full information}.
In: \bbtitle{Proceedings of the AAAI Conference on Artificial Intelligence},
vol. \bseriesno{37},
pp. \bfpage{10473}--\blpage{10480}
(\byear{2023})
\end{bchapter}
\endbibitem

\bibitem[\protect\citeauthoryear{Mohri et~al.}{2019}]{mohri2019agnostic}
\begin{bchapter}
\bauthor{\bsnm{Mohri}, \binits{M.}},
\bauthor{\bsnm{Sivek}, \binits{G.}},
\bauthor{\bsnm{Suresh}, \binits{A.T.}}:
\bctitle{Agnostic federated learning}.
In: \bbtitle{International Conference on Machine Learning}
(\byear{2019})
\end{bchapter}
\endbibitem

\bibitem[\protect\citeauthoryear{Ezzeldin et~al.}{2023}]{ezzeldin2023fairfed}
\begin{bchapter}
\bauthor{\bsnm{Ezzeldin}, \binits{Y.H.}},
\bauthor{\bsnm{Yan}, \binits{S.}},
\bauthor{\bsnm{He}, \binits{C.}},
\bauthor{\bsnm{Ferrara}, \binits{E.}},
\bauthor{\bsnm{Avestimehr}, \binits{A.S.}}:
\bctitle{Fairfed: Enabling group fairness in federated learning}.
In: \bbtitle{Proceedings of the AAAI Conference on Artificial Intelligence},
vol. \bseriesno{37},
pp. \bfpage{7494}--\blpage{7502}
(\byear{2023})
\end{bchapter}
\endbibitem

\bibitem[\protect\citeauthoryear{Lyu et~al.}{2020}]{lyu2020collaborative}
\begin{botherref}
\oauthor{\bsnm{Lyu}, \binits{L.}},
\oauthor{\bsnm{Xu}, \binits{X.}},
\oauthor{\bsnm{Wang}, \binits{Q.}},
\oauthor{\bsnm{Yu}, \binits{H.}}:
Collaborative fairness in federated learning.
Federated Learning: Privacy and Incentive,
189--204
(2020)
\end{botherref}
\endbibitem

\bibitem[\protect\citeauthoryear{Liu et~al.}{2022}]{liu2022fairness}
\begin{bchapter}
\bauthor{\bsnm{Liu}, \binits{S.}},
\bauthor{\bsnm{Ge}, \binits{Y.}},
\bauthor{\bsnm{Xu}, \binits{S.}},
\bauthor{\bsnm{Zhang}, \binits{Y.}},
\bauthor{\bsnm{Marian}, \binits{A.}}:
\bctitle{Fairness-aware federated matrix factorization}.
In: \bbtitle{Proceedings of the 16th ACM Conference on Recommender Systems},
pp. \bfpage{168}--\blpage{178}
(\byear{2022})
\end{bchapter}
\endbibitem

\bibitem[\protect\citeauthoryear{Liu et~al.}{2020}]{liu2020secure}
\begin{barticle}
\bauthor{\bsnm{Liu}, \binits{Y.}},
\bauthor{\bsnm{Kang}, \binits{Y.}},
\bauthor{\bsnm{Xing}, \binits{C.}},
\bauthor{\bsnm{Chen}, \binits{T.}},
\bauthor{\bsnm{Yang}, \binits{Q.}}:
\batitle{A secure federated transfer learning framework}.
\bjtitle{IEEE Intelligent Systems}
\bvolume{35},
\bfpage{70}--\blpage{82}
(\byear{2020})
\end{barticle}
\endbibitem

\bibitem[\protect\citeauthoryear{Peng et~al.}{2020}]{peng2020federated}
\begin{bchapter}
\bauthor{\bsnm{Peng}, \binits{X.}},
\bauthor{\bsnm{Huang}, \binits{Z.}},
\bauthor{\bsnm{Zhu}, \binits{Y.}},
\bauthor{\bsnm{Saenko}, \binits{K.}}:
\bctitle{Federated adversarial domain adaptation}.
In: \bbtitle{International Conference on Learning Representations}
(\byear{2020})
\end{bchapter}
\endbibitem

\bibitem[\protect\citeauthoryear{Yang et~al.}{2020}]{yang2020fedsteg}
\begin{botherref}
\oauthor{\bsnm{Yang}, \binits{H.}},
\oauthor{\bsnm{He}, \binits{H.}},
\oauthor{\bsnm{Zhang}, \binits{W.}},
\oauthor{\bsnm{Cao}, \binits{X.}}:
{FedSteg: A Federated Transfer Learning Framework for Secure Image
  Steganalysis}.
IEEE Transactions on Network Science and Engineering
(2020)
\end{botherref}
\endbibitem

\bibitem[\protect\citeauthoryear{Wang et~al.}{2022}]{wang2022efficient}
\begin{botherref}
\oauthor{\bsnm{Wang}, \binits{K.}},
\oauthor{\bsnm{Li}, \binits{J.}},
\oauthor{\bsnm{Wu}, \binits{W.}}, et al.:
An efficient intrusion detection method based on federated transfer learning
  and an extreme learning machine with privacy preservation.
Security and Communication Networks
\textbf{2022}
(2022)
\end{botherref}
\endbibitem

\bibitem[\protect\citeauthoryear{Chen et~al.}{2020}]{chen2020fedhealth}
\begin{barticle}
\bauthor{\bsnm{Chen}, \binits{Y.}},
\bauthor{\bsnm{Qin}, \binits{X.}},
\bauthor{\bsnm{Wang}, \binits{J.}},
\bauthor{\bsnm{Yu}, \binits{C.}},
\bauthor{\bsnm{Gao}, \binits{W.}}:
\batitle{Fedhealth: A federated transfer learning framework for wearable
  healthcare}.
\bjtitle{IEEE Intelligent Systems}
\bvolume{35}(\bissue{4}),
\bfpage{83}--\blpage{93}
(\byear{2020})
\end{barticle}
\endbibitem

\bibitem[\protect\citeauthoryear{Feng et~al.}{2022}]{feng2022semi}
\begin{barticle}
\bauthor{\bsnm{Feng}, \binits{S.}},
\bauthor{\bsnm{Li}, \binits{B.}},
\bauthor{\bsnm{Yu}, \binits{H.}},
\bauthor{\bsnm{Liu}, \binits{Y.}},
\bauthor{\bsnm{Yang}, \binits{Q.}}:
\batitle{Semi-supervised federated heterogeneous transfer learning}.
\bjtitle{Knowledge-Based Systems}
\bvolume{252},
\bfpage{109384}
(\byear{2022})
\end{barticle}
\endbibitem

\bibitem[\protect\citeauthoryear{Jin et~al.}{2023}]{jin2023fedcrack}
\begin{botherref}
\oauthor{\bsnm{Jin}, \binits{X.}},
\oauthor{\bsnm{Bu}, \binits{J.}},
\oauthor{\bsnm{Yu}, \binits{Z.}},
\oauthor{\bsnm{Zhang}, \binits{H.}},
\oauthor{\bsnm{Wang}, \binits{Y.}}:
{FedCrack}: Federated transfer learning with unsupervised representation for
  crack detection.
IEEE Transactions on Intelligent Transportation Systems
(2023)
\end{botherref}
\endbibitem

\bibitem[\protect\citeauthoryear{Smith et~al.}{2017}]{smith2017federated}
\begin{bchapter}
\bauthor{\bsnm{Smith}, \binits{V.}},
\bauthor{\bsnm{Chiang}, \binits{C.-K.}},
\bauthor{\bsnm{Sanjabi}, \binits{M.}},
\bauthor{\bsnm{Talwalkar}, \binits{A.S.}}:
\bctitle{Federated multi-task learning}.
In: \bbtitle{Advances in Neural Information Processing Systems},
pp. \bfpage{4427}--\blpage{4437}
(\byear{2017})
\end{bchapter}
\endbibitem

\bibitem[\protect\citeauthoryear{Caldas et~al.}{2018}]{caldas2018federated}
\begin{bchapter}
\bauthor{\bsnm{Caldas}, \binits{S.}},
\bauthor{\bsnm{Smith}, \binits{V.}},
\bauthor{\bsnm{Talwalkar}, \binits{A.}}:
\bctitle{Federated kernelized multi-task learning}.
In: \bbtitle{Conference on Machine Learning and Systems}
(\byear{2018})
\end{bchapter}
\endbibitem

\bibitem[\protect\citeauthoryear{Sattler et~al.}{2020}]{sattler2020clustered}
\begin{botherref}
\oauthor{\bsnm{Sattler}, \binits{F.}},
\oauthor{\bsnm{M{\"u}ller}, \binits{K.-R.}},
\oauthor{\bsnm{Samek}, \binits{W.}}:
Clustered federated learning: Model-agnostic distributed multitask optimization
  under privacy constraints.
IEEE Transactions on Neural Networks and Learning Systems
(2020)
\end{botherref}
\endbibitem

\bibitem[\protect\citeauthoryear{Cao et~al.}{2023}]{cao2023cross}
\begin{barticle}
\bauthor{\bsnm{Cao}, \binits{X.}},
\bauthor{\bsnm{Li}, \binits{Z.}},
\bauthor{\bsnm{Sun}, \binits{G.}},
\bauthor{\bsnm{Yu}, \binits{H.}},
\bauthor{\bsnm{Guizani}, \binits{M.}}:
\batitle{Cross-silo heterogeneous model federated multitask learning}.
\bjtitle{Knowledge-Based Systems}
\bvolume{265},
\bfpage{110347}
(\byear{2023})
\end{barticle}
\endbibitem

\bibitem[\protect\citeauthoryear{Marfoq et~al.}{2021}]{marfoq2021federated}
\begin{barticle}
\bauthor{\bsnm{Marfoq}, \binits{O.}},
\bauthor{\bsnm{Neglia}, \binits{G.}},
\bauthor{\bsnm{Bellet}, \binits{A.}},
\bauthor{\bsnm{Kameni}, \binits{L.}},
\bauthor{\bsnm{Vidal}, \binits{R.}}:
\batitle{Federated multi-task learning under a mixture of distributions}.
\bjtitle{Advances in Neural Information Processing Systems}
\bvolume{34},
\bfpage{15434}--\blpage{15447}
(\byear{2021})
\end{barticle}
\endbibitem

\bibitem[\protect\citeauthoryear{Chen and Zhang}{2022}]{chen2022fedmsplit}
\begin{bchapter}
\bauthor{\bsnm{Chen}, \binits{J.}},
\bauthor{\bsnm{Zhang}, \binits{A.}}:
\bctitle{{FedMSplit}: Correlation-adaptive federated multi-task learning across
  multimodal split networks}.
In: \bbtitle{Proceedings of the 28th ACM SIGKDD Conference on Knowledge
  Discovery and Data Mining},
pp. \bfpage{87}--\blpage{96}
(\byear{2022})
\end{bchapter}
\endbibitem

\bibitem[\protect\citeauthoryear{He et~al.}{2022}]{he2022spreadgnn}
\begin{bchapter}
\bauthor{\bsnm{He}, \binits{C.}},
\bauthor{\bsnm{Ceyani}, \binits{E.}},
\bauthor{\bsnm{Balasubramanian}, \binits{K.}},
\bauthor{\bsnm{Annavaram}, \binits{M.}},
\bauthor{\bsnm{Avestimehr}, \binits{S.}}:
\bctitle{{SpreadGNN}: Decentralized multi-task federated learning for graph
  neural networks on molecular data}.
In: \bbtitle{Proceedings of the AAAI Conference on Artificial Intelligence},
vol. \bseriesno{36},
pp. \bfpage{6865}--\blpage{6873}
(\byear{2022})
\end{bchapter}
\endbibitem

\bibitem[\protect\citeauthoryear{Yao et~al.}{2019}]{yao2019federated}
\begin{bchapter}
\bauthor{\bsnm{Yao}, \binits{X.}},
\bauthor{\bsnm{Huang}, \binits{T.}},
\bauthor{\bsnm{Zhang}, \binits{R.-X.}},
\bauthor{\bsnm{Li}, \binits{R.}},
\bauthor{\bsnm{Sun}, \binits{L.}}:
\bctitle{Federated learning with unbiased gradient aggregation and controllable
  meta updating}.
In: \bbtitle{Advances in Neural Information Processing Systems Workshop}
(\byear{2019})
\end{bchapter}
\endbibitem

\bibitem[\protect\citeauthoryear{Fallah et~al.}{2020}]{fallah2020personalized}
\begin{bchapter}
\bauthor{\bsnm{Fallah}, \binits{A.}},
\bauthor{\bsnm{Mokhtari}, \binits{A.}},
\bauthor{\bsnm{Ozdaglar}, \binits{A.}}:
\bctitle{Personalized federated learning with theoretical guarantees: A
  model-agnostic meta-learning approach}.
In: \bbtitle{Advances in Neural Information Processing Systems}
(\byear{2020})
\end{bchapter}
\endbibitem

\bibitem[\protect\citeauthoryear{Wang et~al.}{2023}]{wang2023memory}
\begin{barticle}
\bauthor{\bsnm{Wang}, \binits{B.}},
\bauthor{\bsnm{Yuan}, \binits{Z.}},
\bauthor{\bsnm{Ying}, \binits{Y.}},
\bauthor{\bsnm{Yang}, \binits{T.}}:
\batitle{Memory-based optimization methods for model-agnostic meta-learning and
  personalized federated learning}.
\bjtitle{Journal of Machine Learning Research}
\bvolume{24},
\bfpage{1}--\blpage{46}
(\byear{2023})
\end{barticle}
\endbibitem

\bibitem[\protect\citeauthoryear{Lin et~al.}{2020}]{lin2020meta}
\begin{bchapter}
\bauthor{\bsnm{Lin}, \binits{Y.}},
\bauthor{\bsnm{Ren}, \binits{P.}},
\bauthor{\bsnm{Chen}, \binits{Z.}},
\bauthor{\bsnm{Ren}, \binits{Z.}},
\bauthor{\bsnm{Yu}, \binits{D.}},
\bauthor{\bsnm{Ma}, \binits{J.}},
\bauthor{\bsnm{Rijke}, \binits{M.d.}},
\bauthor{\bsnm{Cheng}, \binits{X.}}:
\bctitle{Meta matrix factorization for federated rating predictions}.
In: \bbtitle{SIGIR},
pp. \bfpage{981}--\blpage{990}
(\byear{2020})
\end{bchapter}
\endbibitem

\bibitem[\protect\citeauthoryear{Li and Wang}{2019}]{li2019fedmd}
\begin{bchapter}
\bauthor{\bsnm{Li}, \binits{D.}},
\bauthor{\bsnm{Wang}, \binits{J.}}:
\bctitle{{FedMD: Heterogenous federated learning via model distillation}}.
In: \bbtitle{Advances in Neural Information Processing Systems Workshop}
(\byear{2019})
\end{bchapter}
\endbibitem

\bibitem[\protect\citeauthoryear{He et~al.}{2020}]{he2020group}
\begin{botherref}
\oauthor{\bsnm{He}, \binits{C.}},
\oauthor{\bsnm{Annavaram}, \binits{M.}},
\oauthor{\bsnm{Avestimehr}, \binits{S.}}:
Group knowledge transfer: Federated learning of large cnns at the edge.
Advances in Neural Information Processing Systems
(2020)
\end{botherref}
\endbibitem

\bibitem[\protect\citeauthoryear{Yang et~al.}{2023}]{yang2023fedfed}
\begin{bchapter}
\bauthor{\bsnm{Yang}, \binits{Z.}},
\bauthor{\bsnm{Zhang}, \binits{Y.}},
\bauthor{\bsnm{Zheng}, \binits{Y.}},
\bauthor{\bsnm{Tian}, \binits{X.}},
\bauthor{\bsnm{Peng}, \binits{H.}},
\bauthor{\bsnm{Liu}, \binits{T.}},
\bauthor{\bsnm{Han}, \binits{B.}}:
\bctitle{{FedFed}: Feature distillation against data heterogeneity in federated
  learning}.
In: \bbtitle{Thirty-seventh Conference on Neural Information Processing
  Systems}
(\byear{2023})
\end{bchapter}
\endbibitem

\bibitem[\protect\citeauthoryear{Lin et~al.}{2020}]{lin2020ensemble}
\begin{barticle}
\bauthor{\bsnm{Lin}, \binits{T.}},
\bauthor{\bsnm{Kong}, \binits{L.}},
\bauthor{\bsnm{Stich}, \binits{S.U.}},
\bauthor{\bsnm{Jaggi}, \binits{M.}}:
\batitle{Ensemble distillation for robust model fusion in federated learning}.
\bjtitle{Advances in Neural Information Processing Systems}
\bvolume{33},
\bfpage{2351}--\blpage{2363}
(\byear{2020})
\end{barticle}
\endbibitem

\bibitem[\protect\citeauthoryear{Yang et~al.}{2023}]{yang2023fedack}
\begin{bchapter}
\bauthor{\bsnm{Yang}, \binits{Y.}},
\bauthor{\bsnm{Yang}, \binits{R.}},
\bauthor{\bsnm{Peng}, \binits{H.}},
\bauthor{\bsnm{Li}, \binits{Y.}},
\bauthor{\bsnm{Li}, \binits{T.}},
\bauthor{\bsnm{Liao}, \binits{Y.}},
\bauthor{\bsnm{Zhou}, \binits{P.}}:
\bctitle{{FedACK}: Federated adversarial contrastive knowledge distillation for
  cross-lingual and cross-model social bot detection}.
In: \bbtitle{Proceedings of the ACM Web Conference 2023},
pp. \bfpage{1314}--\blpage{1323}
(\byear{2023})
\end{bchapter}
\endbibitem

\bibitem[\protect\citeauthoryear{Ma et~al.}{2022}]{CFeD}
\begin{bchapter}
\bauthor{\bsnm{Ma}, \binits{Y.}},
\bauthor{\bsnm{Xie}, \binits{Z.}},
\bauthor{\bsnm{Wang}, \binits{J.}},
\bauthor{\bsnm{Chen}, \binits{K.}},
\bauthor{\bsnm{Shou}, \binits{L.}}:
\bctitle{Continual federated learning based on knowledge distillation}.
In: \beditor{\bsnm{Raedt}, \binits{L.D.}} (ed.)
\bbtitle{Proceedings of the Thirty-First International Joint Conference on
  Artificial Intelligence, {IJCAI-22}},
pp. \bfpage{2182}--\blpage{2188}
(\byear{2022}).
\doiurl{10.24963/ijcai.2022/303} .
\bcomment{Main Track}.
\burl{https://doi.org/10.24963/ijcai.2022/303}
\end{bchapter}
\endbibitem

\bibitem[\protect\citeauthoryear{Wu et~al.}{2023}]{wu2023fedict}
\begin{botherref}
\oauthor{\bsnm{Wu}, \binits{Z.}},
\oauthor{\bsnm{Sun}, \binits{S.}},
\oauthor{\bsnm{Wang}, \binits{Y.}},
\oauthor{\bsnm{Liu}, \binits{M.}},
\oauthor{\bsnm{Pan}, \binits{Q.}},
\oauthor{\bsnm{Jiang}, \binits{X.}},
\oauthor{\bsnm{Gao}, \binits{B.}}:
{FedICT}: Federated multi-task distillation for multi-access edge computing.
IEEE Transactions on Parallel and Distributed Systems
(2023)
\end{botherref}
\endbibitem

\bibitem[\protect\citeauthoryear{Zhang et~al.}{2023}]{zhang2023distill}
\begin{botherref}
\oauthor{\bsnm{Zhang}, \binits{Y.}},
\oauthor{\bsnm{Zhang}, \binits{W.}},
\oauthor{\bsnm{Pu}, \binits{L.}},
\oauthor{\bsnm{Lin}, \binits{T.}},
\oauthor{\bsnm{Yan}, \binits{J.}}:
To distill or not to distill: Towards fast, accurate and communication
  efficient federated distillation learning.
IEEE Internet of Things Journal
(2023)
\end{botherref}
\endbibitem

\bibitem[\protect\citeauthoryear{Zhang et~al.}{2022}]{fedftg}
\begin{bchapter}
\bauthor{\bsnm{Zhang}, \binits{L.}},
\bauthor{\bsnm{Shen}, \binits{L.}},
\bauthor{\bsnm{Ding}, \binits{L.}},
\bauthor{\bsnm{Tao}, \binits{D.}},
\bauthor{\bsnm{Duan}, \binits{L.-Y.}}:
\bctitle{Fine-tuning global model via data-free knowledge distillation for
  non-iid federated learning}.
In: \bbtitle{Proceedings of the IEEE/CVF Conference on Computer Vision and
  Pattern Recognition},
pp. \bfpage{10174}--\blpage{10183}
(\byear{2022})
\end{bchapter}
\endbibitem

\bibitem[\protect\citeauthoryear{Jeong et~al.}{2021}]{jeong2020federated}
\begin{bchapter}
\bauthor{\bsnm{Jeong}, \binits{W.}},
\bauthor{\bsnm{Yoon}, \binits{J.}},
\bauthor{\bsnm{Yang}, \binits{E.}},
\bauthor{\bsnm{Hwang}, \binits{S.J.}}:
\bctitle{Federated semi-supervised learning with inter-client consistency \&
  disjoint learning}.
In: \bbtitle{International Conference on Learning Representations}
(\byear{2021})
\end{bchapter}
\endbibitem

\bibitem[\protect\citeauthoryear{Papernot
  et~al.}{2017}]{papernot2017semisupervised}
\begin{bchapter}
\bauthor{\bsnm{Papernot}, \binits{N.}},
\bauthor{\bsnm{Abadi}, \binits{M.}},
\bauthor{\bsnm{Erlingsson}, \binits{{\'U}.}},
\bauthor{\bsnm{Goodfellow}, \binits{I.}},
\bauthor{\bsnm{Talwar}, \binits{K.}}:
\bctitle{Semi-supervised knowledge transfer for deep learning from private
  training data}.
In: \bbtitle{International Conference on Learning Representations}
(\byear{2017})
\end{bchapter}
\endbibitem

\bibitem[\protect\citeauthoryear{Diao et~al.}{2022}]{semifl}
\begin{barticle}
\bauthor{\bsnm{Diao}, \binits{E.}},
\bauthor{\bsnm{Ding}, \binits{J.}},
\bauthor{\bsnm{Tarokh}, \binits{V.}}:
\batitle{Semifl: Semi-supervised federated learning for unlabeled clients with
  alternate training}.
\bjtitle{Advances in Neural Information Processing Systems}
\bvolume{35},
\bfpage{17871}--\blpage{17884}
(\byear{2022})
\end{barticle}
\endbibitem

\bibitem[\protect\citeauthoryear{Jiang et~al.}{2022}]{jiang2022dynamic}
\begin{bchapter}
\bauthor{\bsnm{Jiang}, \binits{M.}},
\bauthor{\bsnm{Yang}, \binits{H.}},
\bauthor{\bsnm{Li}, \binits{X.}},
\bauthor{\bsnm{Liu}, \binits{Q.}},
\bauthor{\bsnm{Heng}, \binits{P.-A.}},
\bauthor{\bsnm{Dou}, \binits{Q.}}:
\bctitle{Dynamic bank learning for semi-supervised federated image diagnosis
  with class imbalance}.
In: \bbtitle{International Conference on Medical Image Computing and
  Computer-Assisted Intervention},
pp. \bfpage{196}--\blpage{206}
(\byear{2022}).
\bcomment{Springer}
\end{bchapter}
\endbibitem

\bibitem[\protect\citeauthoryear{Wei and Huang}{2023}]{fapl}
\begin{botherref}
\oauthor{\bsnm{Wei}, \binits{X.-X.}},
\oauthor{\bsnm{Huang}, \binits{H.}}:
Balanced federated semi-supervised learning with fairness-aware
  pseudo-labeling.
IEEE Transactions on Neural Networks and Learning Systems
(2023)
\end{botherref}
\endbibitem

\bibitem[\protect\citeauthoryear{Liang et~al.}{2022}]{rscfed}
\begin{bchapter}
\bauthor{\bsnm{Liang}, \binits{X.}},
\bauthor{\bsnm{Lin}, \binits{Y.}},
\bauthor{\bsnm{Fu}, \binits{H.}},
\bauthor{\bsnm{Zhu}, \binits{L.}},
\bauthor{\bsnm{Li}, \binits{X.}}:
\bctitle{Rscfed: Random sampling consensus federated semi-supervised learning}.
In: \bbtitle{Proceedings of the IEEE/CVF Conference on Computer Vision and
  Pattern Recognition},
pp. \bfpage{10154}--\blpage{10163}
(\byear{2022})
\end{bchapter}
\endbibitem

\bibitem[\protect\citeauthoryear{Li et~al.}{2023}]{cbafed}
\begin{bchapter}
\bauthor{\bsnm{Li}, \binits{M.}},
\bauthor{\bsnm{Li}, \binits{Q.}},
\bauthor{\bsnm{Wang}, \binits{Y.}}:
\bctitle{Class balanced adaptive pseudo labeling for federated semi-supervised
  learning}.
In: \bbtitle{Proceedings of the IEEE/CVF Conference on Computer Vision and
  Pattern Recognition},
pp. \bfpage{16292}--\blpage{16301}
(\byear{2023})
\end{bchapter}
\endbibitem

\bibitem[\protect\citeauthoryear{Shang et~al.}{2023}]{suma}
\begin{botherref}
\oauthor{\bsnm{Shang}, \binits{X.}},
\oauthor{\bsnm{Huang}, \binits{G.}},
\oauthor{\bsnm{Lu}, \binits{Y.}},
\oauthor{\bsnm{Lou}, \binits{J.}},
\oauthor{\bsnm{Han}, \binits{B.}},
\oauthor{\bsnm{Cheung}, \binits{Y.-m.}},
\oauthor{\bsnm{Wang}, \binits{H.}}:
Federated semi-supervised learning with annotation heterogeneity.
arXiv preprint arXiv:2303.02445
(2023)
\end{botherref}
\endbibitem

\bibitem[\protect\citeauthoryear{Kang et~al.}{2022}]{fedcvt}
\begin{barticle}
\bauthor{\bsnm{Kang}, \binits{Y.}},
\bauthor{\bsnm{Liu}, \binits{Y.}},
\bauthor{\bsnm{Liang}, \binits{X.}}:
\batitle{Fedcvt: Semi-supervised vertical federated learning with cross-view
  training}.
\bjtitle{ACM Transactions on Intelligent Systems and Technology (TIST)}
\bvolume{13}(\bissue{4}),
\bfpage{1}--\blpage{16}
(\byear{2022})
\end{barticle}
\endbibitem

\bibitem[\protect\citeauthoryear{Fan and Liu}{2020}]{fan2020federated}
\begin{botherref}
\oauthor{\bsnm{Fan}, \binits{C.}},
\oauthor{\bsnm{Liu}, \binits{P.}}:
Federated generative adversarial learning.
arXiv preprint arXiv:2005.03793
(2020)
\end{botherref}
\endbibitem

\bibitem[\protect\citeauthoryear{Rasouli et~al.}{2020}]{rasouli2020fedgan}
\begin{botherref}
\oauthor{\bsnm{Rasouli}, \binits{M.}},
\oauthor{\bsnm{Sun}, \binits{T.}},
\oauthor{\bsnm{Rajagopal}, \binits{R.}}:
{FedGAN: Federated generative adversarial networks for distributed data}.
arXiv preprint arXiv:2006.07228
(2020)
\end{botherref}
\endbibitem

\bibitem[\protect\citeauthoryear{Augenstein
  et~al.}{2020}]{augenstein2020generative}
\begin{bchapter}
\bauthor{\bsnm{Augenstein}, \binits{S.}},
\bauthor{\bsnm{McMahan}, \binits{H.B.}},
\bauthor{\bsnm{Ramage}, \binits{D.}},
\bauthor{\bsnm{Ramaswamy}, \binits{S.}},
\bauthor{\bsnm{Kairouz}, \binits{P.}},
\bauthor{\bsnm{Chen}, \binits{M.}},
\bauthor{\bsnm{Mathews}, \binits{R.}},
\bauthor{\bsnm{Arcas}, \binits{B.A.}}:
\bctitle{Generative models for effective ml on private, decentralized
  datasets}.
In: \bbtitle{International Conference on Learning Representations}
(\byear{2020})
\end{bchapter}
\endbibitem

\bibitem[\protect\citeauthoryear{Qi et~al.}{2022}]{qi2022fairvfl}
\begin{barticle}
\bauthor{\bsnm{Qi}, \binits{T.}},
\bauthor{\bsnm{Wu}, \binits{F.}},
\bauthor{\bsnm{Wu}, \binits{C.}},
\bauthor{\bsnm{Lyu}, \binits{L.}},
\bauthor{\bsnm{Xu}, \binits{T.}},
\bauthor{\bsnm{Liao}, \binits{H.}},
\bauthor{\bsnm{Yang}, \binits{Z.}},
\bauthor{\bsnm{Huang}, \binits{Y.}},
\bauthor{\bsnm{Xie}, \binits{X.}}:
\batitle{Fairvfl: A fair vertical federated learning framework with contrastive
  adversarial learning}.
\bjtitle{Advances in Neural Information Processing Systems}
\bvolume{35},
\bfpage{7852}--\blpage{7865}
(\byear{2022})
\end{barticle}
\endbibitem

\bibitem[\protect\citeauthoryear{Li et~al.}{2023}]{li2023federated}
\begin{bchapter}
\bauthor{\bsnm{Li}, \binits{X.}},
\bauthor{\bsnm{Song}, \binits{Z.}},
\bauthor{\bsnm{Yang}, \binits{J.}}:
\bctitle{Federated adversarial learning: A framework with convergence
  analysis}.
In: \bbtitle{International Conference on Machine Learning},
pp. \bfpage{19932}--\blpage{19959}
(\byear{2023}).
\bcomment{PMLR}
\end{bchapter}
\endbibitem

\bibitem[\protect\citeauthoryear{Zhang et~al.}{2023}]{zhang2023delving}
\begin{bchapter}
\bauthor{\bsnm{Zhang}, \binits{J.}},
\bauthor{\bsnm{Li}, \binits{B.}},
\bauthor{\bsnm{Chen}, \binits{C.}},
\bauthor{\bsnm{Lyu}, \binits{L.}},
\bauthor{\bsnm{Wu}, \binits{S.}},
\bauthor{\bsnm{Ding}, \binits{S.}},
\bauthor{\bsnm{Wu}, \binits{C.}}:
\bctitle{Delving into the adversarial robustness of federated learning}.
In: \bbtitle{Proceedings of the AAAI Conference on Artificial Intelligence}
(\byear{2023})
\end{bchapter}
\endbibitem

\bibitem[\protect\citeauthoryear{Chen et~al.}{2022}]{chen2022calfat}
\begin{barticle}
\bauthor{\bsnm{Chen}, \binits{C.}},
\bauthor{\bsnm{Liu}, \binits{Y.}},
\bauthor{\bsnm{Ma}, \binits{X.}},
\bauthor{\bsnm{Lyu}, \binits{L.}}:
\batitle{Calfat: Calibrated federated adversarial training with label
  skewness}.
\bjtitle{Advances in Neural Information Processing Systems}
\bvolume{35},
\bfpage{3569}--\blpage{3581}
(\byear{2022})
\end{barticle}
\endbibitem

\bibitem[\protect\citeauthoryear{Hong et~al.}{2023}]{hong2023federated}
\begin{bchapter}
\bauthor{\bsnm{Hong}, \binits{J.}},
\bauthor{\bsnm{Wang}, \binits{H.}},
\bauthor{\bsnm{Wang}, \binits{Z.}},
\bauthor{\bsnm{Zhou}, \binits{J.}}:
\bctitle{Federated robustness propagation: sharing adversarial robustness in
  heterogeneous federated learning}.
In: \bbtitle{Proceedings of the AAAI Conference on Artificial Intelligence},
vol. \bseriesno{37},
pp. \bfpage{7893}--\blpage{7901}
(\byear{2023})
\end{bchapter}
\endbibitem

\bibitem[\protect\citeauthoryear{Bram et~al.}{2020}]{van2020towards}
\begin{bchapter}
\bauthor{\bsnm{Bram}, \binits{B.v.}},
\bauthor{\bsnm{Saeed}, \binits{A.}},
\bauthor{\bsnm{Ozcelebi}, \binits{T.}}:
\bctitle{Towards federated unsupervised representation learning}.
In: \bbtitle{ACM EdgeSys},
pp. \bfpage{31}--\blpage{36}
(\byear{2020})
\end{bchapter}
\endbibitem

\bibitem[\protect\citeauthoryear{Grammenos
  et~al.}{2020}]{grammenos2020federated}
\begin{bchapter}
\bauthor{\bsnm{Grammenos}, \binits{A.}},
\bauthor{\bsnm{Mendoza~Smith}, \binits{R.}},
\bauthor{\bsnm{Crowcroft}, \binits{J.}},
\bauthor{\bsnm{Mascolo}, \binits{C.}}:
\bctitle{Federated principal component analysis}.
In: \bbtitle{Advances in Neural Information Processing Systems}
(\byear{2020})
\end{bchapter}
\endbibitem

\bibitem[\protect\citeauthoryear{Zhang et~al.}{2023}]{zhang2020federated}
\begin{barticle}
\bauthor{\bsnm{Zhang}, \binits{F.}},
\bauthor{\bsnm{Kuang}, \binits{K.}},
\bauthor{\bsnm{Chen}, \binits{L.}},
\bauthor{\bsnm{You}, \binits{Z.}},
\bauthor{\bsnm{Shen}, \binits{T.}},
\bauthor{\bsnm{Xiao}, \binits{J.}},
\bauthor{\bsnm{Zhang}, \binits{Y.}},
\bauthor{\bsnm{Wu}, \binits{C.}},
\bauthor{\bsnm{Wu}, \binits{F.}},
\bauthor{\bsnm{Zhuang}, \binits{Y.}}, \betal:
\batitle{Federated unsupervised representation learning}.
\bjtitle{Frontiers of Information Technology \& Electronic Engineering}
\bvolume{24}(\bissue{8}),
\bfpage{1181}--\blpage{1193}
(\byear{2023})
\end{barticle}
\endbibitem

\bibitem[\protect\citeauthoryear{Zhuang et~al.}{2022}]{zhuang2021divergence}
\begin{bchapter}
\bauthor{\bsnm{Zhuang}, \binits{W.}},
\bauthor{\bsnm{Wen}, \binits{Y.}},
\bauthor{\bsnm{Zhang}, \binits{S.}}:
\bctitle{Divergence-aware federated self-supervised learning}.
In: \bbtitle{International Conference on Learning Representations}
(\byear{2022})
\end{bchapter}
\endbibitem

\bibitem[\protect\citeauthoryear{Lubana et~al.}{2022}]{lubana2022orchestra}
\begin{bchapter}
\bauthor{\bsnm{Lubana}, \binits{E.}},
\bauthor{\bsnm{Tang}, \binits{C.I.}},
\bauthor{\bsnm{Kawsar}, \binits{F.}},
\bauthor{\bsnm{Dick}, \binits{R.}},
\bauthor{\bsnm{Mathur}, \binits{A.}}:
\bctitle{Orchestra: Unsupervised federated learning via globally consistent
  clustering}.
In: \bbtitle{International Conference on Machine Learning},
pp. \bfpage{14461}--\blpage{14484}
(\byear{2022}).
\bcomment{PMLR}
\end{bchapter}
\endbibitem

\bibitem[\protect\citeauthoryear{Rehman et~al.}{2023}]{rehman2023dawa}
\begin{bchapter}
\bauthor{\bsnm{Rehman}, \binits{Y.A.U.}},
\bauthor{\bsnm{Gao}, \binits{Y.}},
\bauthor{\bsnm{Gusmao}, \binits{P.P.B.}},
\bauthor{\bsnm{Alibeigi}, \binits{M.}},
\bauthor{\bsnm{Shen}, \binits{J.}},
\bauthor{\bsnm{Lane}, \binits{N.D.}}:
\bctitle{{L-DAWA}: Layer-wise divergence aware weight aggregation in federated
  self-supervised visual representation learning}.
In: \bbtitle{Proceedings of the IEEE/CVF International Conference on Computer
  Vision},
pp. \bfpage{16464}--\blpage{16473}
(\byear{2023})
\end{bchapter}
\endbibitem

\bibitem[\protect\citeauthoryear{Han et~al.}{2022}]{han2022fedx}
\begin{bchapter}
\bauthor{\bsnm{Han}, \binits{S.}},
\bauthor{\bsnm{Park}, \binits{S.}},
\bauthor{\bsnm{Wu}, \binits{F.}},
\bauthor{\bsnm{Kim}, \binits{S.}},
\bauthor{\bsnm{Wu}, \binits{C.}},
\bauthor{\bsnm{Xie}, \binits{X.}},
\bauthor{\bsnm{Cha}, \binits{M.}}:
\bctitle{Fedx: Unsupervised federated learning with cross knowledge
  distillation}.
In: \bbtitle{European Conference on Computer Vision},
pp. \bfpage{691}--\blpage{707}
(\byear{2022}).
\bcomment{Springer}
\end{bchapter}
\endbibitem

\bibitem[\protect\citeauthoryear{Zhuo et~al.}{2019}]{zhuo2019federated}
\begin{botherref}
\oauthor{\bsnm{Zhuo}, \binits{H.H.}},
\oauthor{\bsnm{Feng}, \binits{W.}},
\oauthor{\bsnm{Xu}, \binits{Q.}},
\oauthor{\bsnm{Yang}, \binits{Q.}},
\oauthor{\bsnm{Lin}, \binits{Y.}}:
Federated deep reinforcement learning.
arXiv preprint arXiv:1901.08277
(2019)
\end{botherref}
\endbibitem

\bibitem[\protect\citeauthoryear{Wang et~al.}{2020}]{wang2020optimizing}
\begin{bchapter}
\bauthor{\bsnm{Wang}, \binits{H.}},
\bauthor{\bsnm{Kaplan}, \binits{Z.}},
\bauthor{\bsnm{Niu}, \binits{D.}},
\bauthor{\bsnm{Li}, \binits{B.}}:
\bctitle{{Optimizing Federated Learning on Non-IID Data with Reinforcement
  Learning}}.
In: \bbtitle{IEEE International Conference on Computer Communications},
pp. \bfpage{1698}--\blpage{1707}
(\byear{2020}).
\bcomment{IEEE}
\end{bchapter}
\endbibitem

\bibitem[\protect\citeauthoryear{Cha et~al.}{2020}]{cha2020proxy}
\begin{botherref}
\oauthor{\bsnm{Cha}, \binits{H.}},
\oauthor{\bsnm{Park}, \binits{J.}},
\oauthor{\bsnm{Kim}, \binits{H.}},
\oauthor{\bsnm{Bennis}, \binits{M.}},
\oauthor{\bsnm{Kim}, \binits{S.-L.}}:
Proxy experience replay: Federated distillation for distributed reinforcement
  learning.
IEEE Intelligent Systems
(2020)
\end{botherref}
\endbibitem

\bibitem[\protect\citeauthoryear{Zhan and Zhang}{2020}]{zhan2020incentive}
\begin{bchapter}
\bauthor{\bsnm{Zhan}, \binits{Y.}},
\bauthor{\bsnm{Zhang}, \binits{J.}}:
\bctitle{An incentive mechanism design for efficient edge learning by deep
  reinforcement learning approach}.
In: \bbtitle{IEEE International Conference on Computer Communications},
pp. \bfpage{2489}--\blpage{2498}
(\byear{2020}).
\bcomment{IEEE}
\end{bchapter}
\endbibitem

\bibitem[\protect\citeauthoryear{Khodadadian et~al.}{2022}]{FedSAM}
\begin{bchapter}
\bauthor{\bsnm{Khodadadian}, \binits{S.}},
\bauthor{\bsnm{Sharma}, \binits{P.}},
\bauthor{\bsnm{Joshi}, \binits{G.}},
\bauthor{\bsnm{Maguluri}, \binits{S.T.}}:
\bctitle{Federated reinforcement learning: Linear speedup under markovian
  sampling}.
In: \bbtitle{International Conference on Machine Learning},
pp. \bfpage{10997}--\blpage{11057}
(\byear{2022}).
\bcomment{PMLR}
\end{bchapter}
\endbibitem

\bibitem[\protect\citeauthoryear{Jin et~al.}{2022}]{jin2022federated}
\begin{bchapter}
\bauthor{\bsnm{Jin}, \binits{H.}},
\bauthor{\bsnm{Peng}, \binits{Y.}},
\bauthor{\bsnm{Yang}, \binits{W.}},
\bauthor{\bsnm{Wang}, \binits{S.}},
\bauthor{\bsnm{Zhang}, \binits{Z.}}:
\bctitle{Federated reinforcement learning with environment heterogeneity}.
In: \bbtitle{International Conference on Artificial Intelligence and
  Statistics},
pp. \bfpage{18}--\blpage{37}
(\byear{2022}).
\bcomment{PMLR}
\end{bchapter}
\endbibitem

\bibitem[\protect\citeauthoryear{Mai et~al.}{2023}]{SCCD}
\begin{barticle}
\bauthor{\bsnm{Mai}, \binits{W.}},
\bauthor{\bsnm{Yao}, \binits{J.}},
\bauthor{\bsnm{Chen}, \binits{G.}},
\bauthor{\bsnm{Zhang}, \binits{Y.}},
\bauthor{\bsnm{Cheung}, \binits{Y.-M.}},
\bauthor{\bsnm{Han}, \binits{B.}}:
\batitle{Server-client collaborative distillation for federated reinforcement
  learning}.
\bjtitle{ACM Transactions on Knowledge Discovery from Data}
\bvolume{18}(\bissue{1}),
\bfpage{1}--\blpage{22}
(\byear{2023})
\end{barticle}
\endbibitem

\bibitem[\protect\citeauthoryear{Fan et~al.}{2023}]{fan2023fedhql}
\begin{bchapter}
\bauthor{\bsnm{Fan}, \binits{F.X.}},
\bauthor{\bsnm{Ma}, \binits{Y.}},
\bauthor{\bsnm{Dai}, \binits{Z.}},
\bauthor{\bsnm{Tan}, \binits{C.}},
\bauthor{\bsnm{Low}, \binits{B.K.H.}}:
\bctitle{Fedhql: Federated heterogeneous q-learning}.
In: \bbtitle{Proceedings of the 2023 International Conference on Autonomous
  Agents and Multiagent Systems},
pp. \bfpage{2810}--\blpage{2812}
(\byear{2023})
\end{bchapter}
\endbibitem

\bibitem[\protect\citeauthoryear{Yang et~al.}{2019}]{yang2019federated}
\begin{barticle}
\bauthor{\bsnm{Yang}, \binits{Q.}},
\bauthor{\bsnm{Liu}, \binits{Y.}},
\bauthor{\bsnm{Chen}, \binits{T.}},
\bauthor{\bsnm{Tong}, \binits{Y.}}:
\batitle{Federated machine learning: Concept and applications}.
\bjtitle{ACM Transactions on Intelligent Systems and Technology}
\bvolume{10}(\bissue{2}),
\bfpage{12}
(\byear{2019})
\end{barticle}
\endbibitem

\bibitem[\protect\citeauthoryear{Li et~al.}{2019}]{li2019federated}
\begin{botherref}
\oauthor{\bsnm{Li}, \binits{Q.}},
\oauthor{\bsnm{Wen}, \binits{Z.}},
\oauthor{\bsnm{He}, \binits{B.}}:
Federated learning systems: Vision, hype and reality for data privacy and
  protection.
arXiv preprint arXiv:1907.09693
(2019)
\end{botherref}
\endbibitem

\bibitem[\protect\citeauthoryear{Kairouz et~al.}{2019}]{kairouz2019advances}
\begin{botherref}
\oauthor{\bsnm{Kairouz}, \binits{P.}},
\oauthor{\bsnm{McMahan}, \binits{H.B.}},
\oauthor{\bsnm{Avent}, \binits{B.}},
\oauthor{\bsnm{Bellet}, \binits{A.}},
\oauthor{\bsnm{Bennis}, \binits{M.}},
\oauthor{\bsnm{Bhagoji}, \binits{A.N.}},
\oauthor{\bsnm{Bonawitz}, \binits{K.}},
\oauthor{\bsnm{Charles}, \binits{Z.}},
\oauthor{\bsnm{Cormode}, \binits{G.}},
\oauthor{\bsnm{Cummings}, \binits{R.}}, et al.:
Advances and open problems in federated learning.
arXiv preprint arXiv:1912.04977
(2019)
\end{botherref}
\endbibitem

\bibitem[\protect\citeauthoryear{Xu and Wang}{2019}]{xu2019federated}
\begin{botherref}
\oauthor{\bsnm{Xu}, \binits{J.}},
\oauthor{\bsnm{Wang}, \binits{F.}}:
Federated learning for healthcare informatics.
arXiv preprint arXiv:1911.06270
(2019)
\end{botherref}
\endbibitem

\bibitem[\protect\citeauthoryear{Lyu et~al.}{2020}]{lyu2020threats}
\begin{botherref}
\oauthor{\bsnm{Lyu}, \binits{L.}},
\oauthor{\bsnm{Yu}, \binits{H.}},
\oauthor{\bsnm{Yang}, \binits{Q.}}:
Threats to federated learning: A survey.
arXiv preprint arXiv:2003.02133
(2020)
\end{botherref}
\endbibitem

\bibitem[\protect\citeauthoryear{Lim et~al.}{2020}]{lim2020federated}
\begin{botherref}
\oauthor{\bsnm{Lim}, \binits{W.Y.B.}},
\oauthor{\bsnm{Luong}, \binits{N.C.}},
\oauthor{\bsnm{Hoang}, \binits{D.T.}},
\oauthor{\bsnm{Jiao}, \binits{Y.}},
\oauthor{\bsnm{Liang}, \binits{Y.-C.}},
\oauthor{\bsnm{Yang}, \binits{Q.}},
\oauthor{\bsnm{Niyato}, \binits{D.}},
\oauthor{\bsnm{Miao}, \binits{C.}}:
Federated learning in mobile edge networks: A comprehensive survey.
IEEE Communications Surveys \& Tutorials
(2020)
\end{botherref}
\endbibitem

\bibitem[\protect\citeauthoryear{Niknam et~al.}{2020}]{niknam2020federated}
\begin{barticle}
\bauthor{\bsnm{Niknam}, \binits{S.}},
\bauthor{\bsnm{Dhillon}, \binits{H.S.}},
\bauthor{\bsnm{Reed}, \binits{J.H.}}:
\batitle{Federated learning for wireless communications: Motivation,
  opportunities, and challenges}.
\bjtitle{IEEE Communications Magazine}
\bvolume{58}(\bissue{6}),
\bfpage{46}--\blpage{51}
(\byear{2020})
\end{barticle}
\endbibitem

\bibitem[\protect\citeauthoryear{Jin et~al.}{2020}]{jin2020survey}
\begin{botherref}
\oauthor{\bsnm{Jin}, \binits{Y.}},
\oauthor{\bsnm{Wei}, \binits{X.}},
\oauthor{\bsnm{Liu}, \binits{Y.}},
\oauthor{\bsnm{Yang}, \binits{Q.}}:
A survey towards federated semi-supervised learning.
arXiv preprint arXiv:2002.11545
(2020)
\end{botherref}
\endbibitem

\bibitem[\protect\citeauthoryear{Lo et~al.}{2020}]{lo2020systematic}
\begin{botherref}
\oauthor{\bsnm{Lo}, \binits{S.K.}},
\oauthor{\bsnm{Lu}, \binits{Q.}},
\oauthor{\bsnm{Wang}, \binits{C.}},
\oauthor{\bsnm{Paik}, \binits{H.}},
\oauthor{\bsnm{Zhu}, \binits{L.}}:
A systematic literature review on federated machine learning: From a software
  engineering perspective.
arXiv preprint arXiv:2007.11354
(2020)
\end{botherref}
\endbibitem

\bibitem[\protect\citeauthoryear{Li et~al.}{2020}]{li2020on}
\begin{bchapter}
\bauthor{\bsnm{Li}, \binits{X.}},
\bauthor{\bsnm{Huang}, \binits{K.}},
\bauthor{\bsnm{Yang}, \binits{W.}},
\bauthor{\bsnm{Wang}, \binits{S.}},
\bauthor{\bsnm{Zhang}, \binits{Z.}}:
\bctitle{On the convergence of fedavg on non-iid data}.
In: \bbtitle{International Conference on Learning Representations}
(\byear{2020})
\end{bchapter}
\endbibitem

\bibitem[\protect\citeauthoryear{Karimireddy
  et~al.}{2020}]{karimireddy2020scaffold}
\begin{bchapter}
\bauthor{\bsnm{Karimireddy}, \binits{S.P.}},
\bauthor{\bsnm{Kale}, \binits{S.}},
\bauthor{\bsnm{Mohri}, \binits{M.}},
\bauthor{\bsnm{Reddi}, \binits{S.J.}},
\bauthor{\bsnm{Stich}, \binits{S.U.}},
\bauthor{\bsnm{Suresh}, \binits{A.T.}}:
\bctitle{Scaffold: Stochastic controlled averaging for federated learning}.
In: \bbtitle{International Conference on Machine Learning},
pp. \bfpage{5132}--\blpage{5143}
(\byear{2020})
\end{bchapter}
\endbibitem

\bibitem[\protect\citeauthoryear{Reddi et~al.}{2021}]{reddi2020adaptive}
\begin{bchapter}
\bauthor{\bsnm{Reddi}, \binits{S.}},
\bauthor{\bsnm{Charles}, \binits{Z.}},
\bauthor{\bsnm{Zaheer}, \binits{M.}},
\bauthor{\bsnm{Garrett}, \binits{Z.}},
\bauthor{\bsnm{Rush}, \binits{K.}},
\bauthor{\bsnm{Kone{\v{c}}n{\`y}}, \binits{J.}},
\bauthor{\bsnm{Kumar}, \binits{S.}},
\bauthor{\bsnm{McMahan}, \binits{H.B.}}:
\bctitle{Adaptive federated optimization}.
In: \bbtitle{International Conference on Learning Representations}
(\byear{2021})
\end{bchapter}
\endbibitem

\bibitem[\protect\citeauthoryear{Singh and Jaggi}{2020}]{singh2020model}
\begin{botherref}
\oauthor{\bsnm{Singh}, \binits{S.P.}},
\oauthor{\bsnm{Jaggi}, \binits{M.}}:
Model fusion via optimal transport.
Advances in Neural Information Processing Systems
\textbf{33}
(2020)
\end{botherref}
\endbibitem

\bibitem[\protect\citeauthoryear{Chai et~al.}{2020}]{chai2020secure}
\begin{barticle}
\bauthor{\bsnm{Chai}, \binits{D.}},
\bauthor{\bsnm{Wang}, \binits{L.}},
\bauthor{\bsnm{Chen}, \binits{K.}},
\bauthor{\bsnm{Yang}, \binits{Q.}}:
\batitle{Secure federated matrix factorization}.
\bjtitle{IEEE Intelligent Systems}
\bvolume{36}(\bissue{5}),
\bfpage{11}--\blpage{20}
(\byear{2020})
\end{barticle}
\endbibitem

\bibitem[\protect\citeauthoryear{Tan et~al.}{2023}]{tan2023heterogeneity}
\begin{bchapter}
\bauthor{\bsnm{Tan}, \binits{Y.}},
\bauthor{\bsnm{Chen}, \binits{C.}},
\bauthor{\bsnm{Zhuang}, \binits{W.}},
\bauthor{\bsnm{Dong}, \binits{X.}},
\bauthor{\bsnm{Lyu}, \binits{L.}},
\bauthor{\bsnm{Long}, \binits{G.}}:
\bctitle{Is heterogeneity notorious? taming heterogeneity to handle test-time
  shift in federated learning}.
In: \bbtitle{Thirty-seventh Conference on Neural Information Processing
  Systems}
(\byear{2023})
\end{bchapter}
\endbibitem

\bibitem[\protect\citeauthoryear{Alawad et~al.}{2020}]{alawad2020privacy}
\begin{botherref}
\oauthor{\bsnm{Alawad}, \binits{M.}},
\oauthor{\bsnm{Yoon}, \binits{H.-J.}},
\oauthor{\bsnm{Gao}, \binits{S.}},
\oauthor{\bsnm{Mumphrey}, \binits{B.}},
\oauthor{\bsnm{Wu}, \binits{X.-C.}},
\oauthor{\bsnm{Durbin}, \binits{E.B.}},
\oauthor{\bsnm{Jeong}, \binits{J.C.}},
\oauthor{\bsnm{Hands}, \binits{I.}},
\oauthor{\bsnm{Rust}, \binits{D.}},
\oauthor{\bsnm{Coyle}, \binits{L.}}, et al.:
Privacy-preserving deep learning nlp models for cancer registries.
IEEE Transactions on Emerging Topics in Computing
(2020)
\end{botherref}
\endbibitem

\bibitem[\protect\citeauthoryear{Tang et~al.}{2022}]{tang2022virtual}
\begin{bchapter}
\bauthor{\bsnm{Tang}, \binits{Z.}},
\bauthor{\bsnm{Zhang}, \binits{Y.}},
\bauthor{\bsnm{Shi}, \binits{S.}},
\bauthor{\bsnm{He}, \binits{X.}},
\bauthor{\bsnm{Han}, \binits{B.}},
\bauthor{\bsnm{Chu}, \binits{X.}}:
\bctitle{Virtual homogeneity learning: Defending against data heterogeneity in
  federated learning}.
In: \bbtitle{International Conference on Machine Learning},
pp. \bfpage{21111}--\blpage{21132}
(\byear{2022}).
\bcomment{PMLR}
\end{bchapter}
\endbibitem

\bibitem[\protect\citeauthoryear{Lu et~al.}{2023}]{lu2023federated}
\begin{botherref}
\oauthor{\bsnm{Lu}, \binits{Y.}},
\oauthor{\bsnm{Chen}, \binits{L.}},
\oauthor{\bsnm{Zhang}, \binits{Y.}},
\oauthor{\bsnm{Zhang}, \binits{Y.}},
\oauthor{\bsnm{Han}, \binits{B.}},
\oauthor{\bsnm{Cheung}, \binits{Y.-m.}},
\oauthor{\bsnm{Wang}, \binits{H.}}:
Federated learning with extremely noisy clients via negative distillation.
arXiv preprint arXiv:2312.12703
(2023)
\end{botherref}
\endbibitem

\bibitem[\protect\citeauthoryear{Finn et~al.}{2017}]{finn2017model}
\begin{bchapter}
\bauthor{\bsnm{Finn}, \binits{C.}},
\bauthor{\bsnm{Abbeel}, \binits{P.}},
\bauthor{\bsnm{Levine}, \binits{S.}}:
\bctitle{Model-agnostic meta-learning for fast adaptation of deep networks}.
In: \bbtitle{International Conference on Machine Learning},
pp. \bfpage{1126}--\blpage{1135}
(\byear{2017})
\end{bchapter}
\endbibitem

\bibitem[\protect\citeauthoryear{Ji et~al.}{2019}]{ji2019knowledge}
\begin{botherref}
\oauthor{\bsnm{Ji}, \binits{S.}},
\oauthor{\bsnm{Long}, \binits{G.}},
\oauthor{\bsnm{Pan}, \binits{S.}},
\oauthor{\bsnm{Zhu}, \binits{T.}},
\oauthor{\bsnm{Jiang}, \binits{J.}},
\oauthor{\bsnm{Wang}, \binits{S.}},
\oauthor{\bsnm{Li}, \binits{X.}}:
Knowledge transferring via model aggregation for online social care.
arXiv preprint arXiv:1905.07665
(2019)
\end{botherref}
\endbibitem

\bibitem[\protect\citeauthoryear{Jiang et~al.}{2019}]{jiang2019improving}
\begin{bchapter}
\bauthor{\bsnm{Jiang}, \binits{Y.}},
\bauthor{\bsnm{Kone{\v{c}}n{\`y}}, \binits{J.}},
\bauthor{\bsnm{Rush}, \binits{K.}},
\bauthor{\bsnm{Kannan}, \binits{S.}}:
\bctitle{Improving federated learning personalization via model agnostic meta
  learning}.
In: \bbtitle{Advances in Neural Information Processing Systems Workshop}
(\byear{2019})
\end{bchapter}
\endbibitem

\bibitem[\protect\citeauthoryear{Lin et~al.}{2021}]{lin2021metagater}
\begin{bchapter}
\bauthor{\bsnm{Lin}, \binits{S.}},
\bauthor{\bsnm{Yang}, \binits{L.}},
\bauthor{\bsnm{He}, \binits{Z.}},
\bauthor{\bsnm{Fan}, \binits{D.}},
\bauthor{\bsnm{Zhang}, \binits{J.}}:
\bctitle{Metagater: Fast learning of conditional channel gated networks via
  federated meta-learning}.
In: \bbtitle{2021 IEEE 18th International Conference on Mobile Ad Hoc and Smart
  Systems (MASS)},
pp. \bfpage{164}--\blpage{172}
(\byear{2021}).
\bcomment{IEEE}
\end{bchapter}
\endbibitem

\bibitem[\protect\citeauthoryear{Sohn et~al.}{2020}]{sohn2020fixmatch}
\begin{barticle}
\bauthor{\bsnm{Sohn}, \binits{K.}},
\bauthor{\bsnm{Berthelot}, \binits{D.}},
\bauthor{\bsnm{Carlini}, \binits{N.}},
\bauthor{\bsnm{Zhang}, \binits{Z.}},
\bauthor{\bsnm{Zhang}, \binits{H.}},
\bauthor{\bsnm{Raffel}, \binits{C.A.}},
\bauthor{\bsnm{Cubuk}, \binits{E.D.}},
\bauthor{\bsnm{Kurakin}, \binits{A.}},
\bauthor{\bsnm{Li}, \binits{C.-L.}}:
\batitle{Fixmatch: Simplifying semi-supervised learning with consistency and
  confidence}.
\bjtitle{Advances in neural information processing systems}
\bvolume{33},
\bfpage{596}--\blpage{608}
(\byear{2020})
\end{barticle}
\endbibitem

\bibitem[\protect\citeauthoryear{Yang et~al.}{2022}]{yang2022survey}
\begin{botherref}
\oauthor{\bsnm{Yang}, \binits{X.}},
\oauthor{\bsnm{Song}, \binits{Z.}},
\oauthor{\bsnm{King}, \binits{I.}},
\oauthor{\bsnm{Xu}, \binits{Z.}}:
A survey on deep semi-supervised learning.
IEEE Transactions on Knowledge and Data Engineering
(2022)
\end{botherref}
\endbibitem

\bibitem[\protect\citeauthoryear{He et~al.}{2020}]{mocov1}
\begin{bchapter}
\bauthor{\bsnm{He}, \binits{K.}},
\bauthor{\bsnm{Fan}, \binits{H.}},
\bauthor{\bsnm{Wu}, \binits{Y.}},
\bauthor{\bsnm{Xie}, \binits{S.}},
\bauthor{\bsnm{Girshick}, \binits{R.}}:
\bctitle{Momentum contrast for unsupervised visual representation learning}.
In: \bbtitle{Proceedings of the IEEE/CVF Conference on Computer Vision and
  Pattern Recognition},
pp. \bfpage{9729}--\blpage{9738}
(\byear{2020})
\end{bchapter}
\endbibitem

\bibitem[\protect\citeauthoryear{Chen et~al.}{2020}]{mocov2}
\begin{botherref}
\oauthor{\bsnm{Chen}, \binits{X.}},
\oauthor{\bsnm{Fan}, \binits{H.}},
\oauthor{\bsnm{Girshick}, \binits{R.}},
\oauthor{\bsnm{He}, \binits{K.}}:
Improved baselines with momentum contrastive learning.
arXiv preprint arXiv:2003.04297
(2020)
\end{botherref}
\endbibitem

\bibitem[\protect\citeauthoryear{Grill et~al.}{2020}]{BYOL}
\begin{barticle}
\bauthor{\bsnm{Grill}, \binits{J.-B.}},
\bauthor{\bsnm{Strub}, \binits{F.}},
\bauthor{\bsnm{Altch{\'e}}, \binits{F.}},
\bauthor{\bsnm{Tallec}, \binits{C.}},
\bauthor{\bsnm{Richemond}, \binits{P.}},
\bauthor{\bsnm{Buchatskaya}, \binits{E.}},
\bauthor{\bsnm{Doersch}, \binits{C.}},
\bauthor{\bsnm{Avila~Pires}, \binits{B.}},
\bauthor{\bsnm{Guo}, \binits{Z.}},
\bauthor{\bsnm{Gheshlaghi~Azar}, \binits{M.}}, \betal:
\batitle{Bootstrap your own latent-a new approach to self-supervised learning}.
\bjtitle{Advances in neural information processing systems}
\bvolume{33},
\bfpage{21271}--\blpage{21284}
(\byear{2020})
\end{barticle}
\endbibitem

\bibitem[\protect\citeauthoryear{Chen et~al.}{2020}]{SimCLR}
\begin{bchapter}
\bauthor{\bsnm{Chen}, \binits{T.}},
\bauthor{\bsnm{Kornblith}, \binits{S.}},
\bauthor{\bsnm{Norouzi}, \binits{M.}},
\bauthor{\bsnm{Hinton}, \binits{G.}}:
\bctitle{A simple framework for contrastive learning of visual
  representations}.
In: \bbtitle{International Conference on Machine Learning},
pp. \bfpage{1597}--\blpage{1607}
(\byear{2020}).
\bcomment{PMLR}
\end{bchapter}
\endbibitem

\bibitem[\protect\citeauthoryear{Chen and He}{2021}]{simsiam}
\begin{bchapter}
\bauthor{\bsnm{Chen}, \binits{X.}},
\bauthor{\bsnm{He}, \binits{K.}}:
\bctitle{Exploring simple siamese representation learning}.
In: \bbtitle{Proceedings of the IEEE/CVF Conference on Computer Vision and
  Pattern Recognition},
pp. \bfpage{15750}--\blpage{15758}
(\byear{2021})
\end{bchapter}
\endbibitem

\bibitem[\protect\citeauthoryear{Watkins and Dayan}{1992}]{watkins1992q}
\begin{barticle}
\bauthor{\bsnm{Watkins}, \binits{C.J.}},
\bauthor{\bsnm{Dayan}, \binits{P.}}:
\batitle{Q-learning}.
\bjtitle{Machine learning}
\bvolume{8},
\bfpage{279}--\blpage{292}
(\byear{1992})
\end{barticle}
\endbibitem

\bibitem[\protect\citeauthoryear{Sutton}{1988}]{sutton1988learning}
\begin{barticle}
\bauthor{\bsnm{Sutton}, \binits{R.S.}}:
\batitle{Learning to predict by the methods of temporal differences}.
\bjtitle{Machine learning}
\bvolume{3},
\bfpage{9}--\blpage{44}
(\byear{1988})
\end{barticle}
\endbibitem

\bibitem[\protect\citeauthoryear{Lattimore and
  Szepesv{\'a}ri}{2020}]{lattimore2020bandit}
\begin{bbook}
\bauthor{\bsnm{Lattimore}, \binits{T.}},
\bauthor{\bsnm{Szepesv{\'a}ri}, \binits{C.}}:
\bbtitle{Bandit Algorithms},
(\byear{2020})
\end{bbook}
\endbibitem

\bibitem[\protect\citeauthoryear{Khalatbarisoltani
  et~al.}{2023}]{khalatbarisoltani2023integrating}
\begin{botherref}
\oauthor{\bsnm{Khalatbarisoltani}, \binits{A.}},
\oauthor{\bsnm{Boulon}, \binits{L.}},
\oauthor{\bsnm{Hu}, \binits{X.}}:
Integrating model predictive control with federated reinforcement learning for
  decentralized energy management of fuel cell vehicles.
IEEE Transactions on Intelligent Transportation Systems
(2023)
\end{botherref}
\endbibitem

\bibitem[\protect\citeauthoryear{Qiu et~al.}{2023}]{qiu2023federated}
\begin{barticle}
\bauthor{\bsnm{Qiu}, \binits{D.}},
\bauthor{\bsnm{Xue}, \binits{J.}},
\bauthor{\bsnm{Zhang}, \binits{T.}},
\bauthor{\bsnm{Wang}, \binits{J.}},
\bauthor{\bsnm{Sun}, \binits{M.}}:
\batitle{Federated reinforcement learning for smart building joint peer-to-peer
  energy and carbon allowance trading}.
\bjtitle{Applied Energy}
\bvolume{333},
\bfpage{120526}
(\byear{2023})
\end{barticle}
\endbibitem

\bibitem[\protect\citeauthoryear{Zhang et~al.}{2022}]{zhang2022federated}
\begin{bchapter}
\bauthor{\bsnm{Zhang}, \binits{Z.}},
\bauthor{\bsnm{Jiang}, \binits{Y.}},
\bauthor{\bsnm{Shi}, \binits{Y.}},
\bauthor{\bsnm{Shi}, \binits{Y.}},
\bauthor{\bsnm{Chen}, \binits{W.}}:
\bctitle{Federated reinforcement learning for real-time electric vehicle
  charging and discharging control}.
In: \bbtitle{2022 IEEE Globecom Workshops (GC Wkshps)},
pp. \bfpage{1717}--\blpage{1722}
(\byear{2022}).
\bcomment{IEEE}
\end{bchapter}
\endbibitem

\bibitem[\protect\citeauthoryear{Arivazhagan et~al.}{2019}]{FedPer}
\begin{botherref}
\oauthor{\bsnm{Arivazhagan}, \binits{M.G.}},
\oauthor{\bsnm{Aggarwal}, \binits{V.}},
\oauthor{\bsnm{Singh}, \binits{A.K.}},
\oauthor{\bsnm{Choudhary}, \binits{S.}}:
Federated learning with personalization layers.
ar{X}iv: 1912.00818
(2019)
{\href{https://arxiv.org/abs/1912.00818}{{arXiv:1912.00818}}}
{[cs.LG]}
\end{botherref}
\endbibitem

\bibitem[\protect\citeauthoryear{Liang et~al.}{2020}]{liang2020think}
\begin{botherref}
\oauthor{\bsnm{Liang}, \binits{P.P.}},
\oauthor{\bsnm{Liu}, \binits{T.}},
\oauthor{\bsnm{Ziyin}, \binits{L.}},
\oauthor{\bsnm{Salakhutdinov}, \binits{R.}},
\oauthor{\bsnm{Morency}, \binits{L.-P.}}:
Think locally, act globally: Federated learning with local and global
  representations.
Advances in Neural Information Processing Systems
(2020)
\end{botherref}
\endbibitem

\bibitem[\protect\citeauthoryear{Deng et~al.}{2020}]{APFL}
\begin{botherref}
\oauthor{\bsnm{Deng}, \binits{Y.}},
\oauthor{\bsnm{Kamani}, \binits{M.M.}},
\oauthor{\bsnm{Mahdavi}, \binits{M.}}:
Adaptive personalized federated learning.
ar{X}iv: 2003.13461
(2020)
\end{botherref}
\endbibitem

\bibitem[\protect\citeauthoryear{Tan et~al.}{2021}]{tan2021towards}
\begin{botherref}
\oauthor{\bsnm{Tan}, \binits{A.Z.}},
\oauthor{\bsnm{Yu}, \binits{H.}},
\oauthor{\bsnm{Cui}, \binits{L.}},
\oauthor{\bsnm{Yang}, \binits{Q.}}:
Towards personalized federated learning.
arXiv preprint arXiv:2103.00710
(2021)
\end{botherref}
\endbibitem

\bibitem[\protect\citeauthoryear{Mansour et~al.}{2020}]{mansour2020three}
\begin{botherref}
\oauthor{\bsnm{Mansour}, \binits{Y.}},
\oauthor{\bsnm{Mohri}, \binits{M.}},
\oauthor{\bsnm{Ro}, \binits{J.}},
\oauthor{\bsnm{Suresh}, \binits{A.T.}}:
Three approaches for personalization with applications to federated learning.
arXiv preprint arXiv:2002.10619
(2020)
\end{botherref}
\endbibitem

\bibitem[\protect\citeauthoryear{Li et~al.}{2021}]{MOON}
\begin{botherref}
\oauthor{\bsnm{Li}, \binits{Q.}},
\oauthor{\bsnm{He}, \binits{B.}},
\oauthor{\bsnm{Song}, \binits{D.}}:
Model-contrastive federated learning.
ar{X}iv: 2103.16257
(2021)
{\href{https://arxiv.org/abs/2103.16257}{{arXiv:2103.16257}}}
{[cs.LG]}
\end{botherref}
\endbibitem

\bibitem[\protect\citeauthoryear{Liu et~al.}{2021}]{liu2021graph}
\begin{botherref}
\oauthor{\bsnm{Liu}, \binits{Y.}},
\oauthor{\bsnm{Pan}, \binits{S.}},
\oauthor{\bsnm{Jin}, \binits{M.}},
\oauthor{\bsnm{Zhou}, \binits{C.}},
\oauthor{\bsnm{Xia}, \binits{F.}},
\oauthor{\bsnm{Yu}, \binits{P.S.}}:
Graph self-supervised learning: A survey.
arXiv preprint arXiv:2103.00111
(2021)
\end{botherref}
\endbibitem

\bibitem[\protect\citeauthoryear{Yang et~al.}{2021}]{yang2021interpretable}
\begin{botherref}
\oauthor{\bsnm{Yang}, \binits{Y.}},
\oauthor{\bsnm{Guan}, \binits{Z.}},
\oauthor{\bsnm{Li}, \binits{J.}},
\oauthor{\bsnm{Zhao}, \binits{W.}},
\oauthor{\bsnm{Cui}, \binits{J.}},
\oauthor{\bsnm{Wang}, \binits{Q.}}:
Interpretable and efficient heterogeneous graph convolutional network.
IEEE Transactions on Knowledge and Data Engineering
(2021)
\end{botherref}
\endbibitem

\bibitem[\protect\citeauthoryear{Jeong et~al.}{2018}]{jeong2018communication}
\begin{bchapter}
\bauthor{\bsnm{Jeong}, \binits{E.}},
\bauthor{\bsnm{Oh}, \binits{S.}},
\bauthor{\bsnm{Kim}, \binits{H.}},
\bauthor{\bsnm{Park}, \binits{J.}},
\bauthor{\bsnm{Bennis}, \binits{M.}},
\bauthor{\bsnm{Kim}, \binits{S.-L.}}:
\bctitle{Communication-efficient on-device machine learning: Federated
  distillation and augmentation under non-{IID} private data}.
In: \bbtitle{Advances in Neural Information Processing Systems}
(\byear{2018})
\end{bchapter}
\endbibitem

\bibitem[\protect\citeauthoryear{Long et~al.}{2021}]{long2021federated}
\begin{bchapter}
\bauthor{\bsnm{Long}, \binits{G.}},
\bauthor{\bsnm{Shen}, \binits{T.}},
\bauthor{\bsnm{Tan}, \binits{Y.}},
\bauthor{\bsnm{Gerrard}, \binits{L.}},
\bauthor{\bsnm{Clarke}, \binits{A.}},
\bauthor{\bsnm{Jiang}, \binits{J.}}:
\bctitle{Federated learning for privacy-preserving open innovation future on
  digital health}.
In: \bbtitle{Humanity Driven AI: Productivity, Well-being, Sustainability and
  Partnership},
pp. \bfpage{113}--\blpage{133}
(\byear{2021})
\end{bchapter}
\endbibitem

\bibitem[\protect\citeauthoryear{Zhu et~al.}{2020}]{zhu2020federated}
\begin{botherref}
\oauthor{\bsnm{Zhu}, \binits{H.}},
\oauthor{\bsnm{Zhang}, \binits{H.}},
\oauthor{\bsnm{Jin}, \binits{Y.}}:
From federated learning to federated neural architecture search: a survey.
Complex \& Intelligent Systems,
1--19
(2020)
\end{botherref}
\endbibitem

\bibitem[\protect\citeauthoryear{He et~al.}{2020}]{he2020fednas}
\begin{bchapter}
\bauthor{\bsnm{He}, \binits{C.}},
\bauthor{\bsnm{Annavaram}, \binits{M.}},
\bauthor{\bsnm{Avestimehr}, \binits{S.}}:
\bctitle{Fed{NAS}: Federated deep learning via neural architecture search}.
In: \bbtitle{Proceedings of the IEEE Conference on Computer Vision and Pattern
  Recognition}
(\byear{2020})
\end{bchapter}
\endbibitem

\bibitem[\protect\citeauthoryear{Singh et~al.}{2020}]{singh2020differentially}
\begin{bchapter}
\bauthor{\bsnm{Singh}, \binits{I.}},
\bauthor{\bsnm{Zhou}, \binits{H.}},
\bauthor{\bsnm{Yang}, \binits{K.}},
\bauthor{\bsnm{Ding}, \binits{M.}},
\bauthor{\bsnm{Lin}, \binits{B.}},
\bauthor{\bsnm{Xie}, \binits{P.}}:
\bctitle{Differentially-private federated neural architecture search}.
In: \bbtitle{FL-International Conference on Machine Learning Workshop}
(\byear{2020})
\end{bchapter}
\endbibitem

\bibitem[\protect\citeauthoryear{Hoang et~al.}{2019}]{hoang2019collective}
\begin{bchapter}
\bauthor{\bsnm{Hoang}, \binits{M.}},
\bauthor{\bsnm{Hoang}, \binits{N.}},
\bauthor{\bsnm{Low}, \binits{B.K.H.}},
\bauthor{\bsnm{Kingsford}, \binits{C.}}:
\bctitle{Collective model fusion for multiple black-box experts}.
In: \bbtitle{International Conference on Machine Learning},
pp. \bfpage{2742}--\blpage{2750}
(\byear{2019}).
\bcomment{PMLR}
\end{bchapter}
\endbibitem

\bibitem[\protect\citeauthoryear{Liu et~al.}{2021}]{FLAME}
\begin{barticle}
\bauthor{\bsnm{Liu}, \binits{R.}},
\bauthor{\bsnm{Cao}, \binits{Y.}},
\bauthor{\bsnm{Chen}, \binits{H.}},
\bauthor{\bsnm{Guo}, \binits{R.}},
\bauthor{\bsnm{Yoshikawa}, \binits{M.}}:
\batitle{Flame: Differentially private federated learning in the shuffle
  model}.
\bjtitle{Proceedings of the AAAI Conference on Artificial Intelligence}
\bvolume{35}(\bissue{10}),
\bfpage{8688}--\blpage{8696}
(\byear{2021})
\doiurl{10.1609/aaai.v35i10.17053}
\end{barticle}
\endbibitem

\bibitem[\protect\citeauthoryear{Thapa et~al.}{2022}]{splitfed}
\begin{barticle}
\bauthor{\bsnm{Thapa}, \binits{C.}},
\bauthor{\bsnm{Mahawaga~Arachchige}, \binits{P.C.}},
\bauthor{\bsnm{Camtepe}, \binits{S.}},
\bauthor{\bsnm{Sun}, \binits{L.}}:
\batitle{Splitfed: When federated learning meets split learning}.
\bjtitle{Proceedings of the AAAI Conference on Artificial Intelligence}
\bvolume{36}(\bissue{8}),
\bfpage{8485}--\blpage{8493}
(\byear{2022})
\doiurl{10.1609/aaai.v36i8.20825}
\end{barticle}
\endbibitem

\bibitem[\protect\citeauthoryear{Zawad et~al.}{2021}]{heterogeneity_robustness}
\begin{barticle}
\bauthor{\bsnm{Zawad}, \binits{S.}},
\bauthor{\bsnm{Ali}, \binits{A.}},
\bauthor{\bsnm{Chen}, \binits{P.-Y.}},
\bauthor{\bsnm{Anwar}, \binits{A.}},
\bauthor{\bsnm{Zhou}, \binits{Y.}},
\bauthor{\bsnm{Baracaldo}, \binits{N.}},
\bauthor{\bsnm{Tian}, \binits{Y.}},
\bauthor{\bsnm{Yan}, \binits{F.}}:
\batitle{Curse or redemption? how data heterogeneity affects the robustness of
  federated learning}.
\bjtitle{Proceedings of the AAAI Conference on Artificial Intelligence}
\bvolume{35}(\bissue{12}),
\bfpage{10807}--\blpage{10814}
(\byear{2021})
\doiurl{10.1609/aaai.v35i12.17291}
\end{barticle}
\endbibitem

\bibitem[\protect\citeauthoryear{Ozdayi et~al.}{2021}]{robust_learning_rate}
\begin{barticle}
\bauthor{\bsnm{Ozdayi}, \binits{M.S.}},
\bauthor{\bsnm{Kantarcioglu}, \binits{M.}},
\bauthor{\bsnm{Gel}, \binits{Y.R.}}:
\batitle{Defending against backdoors in federated learning with robust learning
  rate}.
\bjtitle{Proceedings of the AAAI Conference on Artificial Intelligence}
\bvolume{35}(\bissue{10}),
\bfpage{9268}--\blpage{9276}
(\byear{2021})
\doiurl{10.1609/aaai.v35i10.17118}
\end{barticle}
\endbibitem

\bibitem[\protect\citeauthoryear{Zhao et~al.}{2022}]{fedinv}
\begin{barticle}
\bauthor{\bsnm{Zhao}, \binits{B.}},
\bauthor{\bsnm{Sun}, \binits{P.}},
\bauthor{\bsnm{Wang}, \binits{T.}},
\bauthor{\bsnm{Jiang}, \binits{K.}}:
\batitle{Fedinv: Byzantine-robust federated learning by inversing local model
  updates}.
\bjtitle{Proceedings of the AAAI Conference on Artificial Intelligence}
\bvolume{36}(\bissue{8}),
\bfpage{9171}--\blpage{9179}
(\byear{2022})
\doiurl{10.1609/aaai.v36i8.20903}
\end{barticle}
\endbibitem

\bibitem[\protect\citeauthoryear{Zhang et~al.}{2022}]{neurotoxin}
\begin{bchapter}
\bauthor{\bsnm{Zhang}, \binits{Z.}},
\bauthor{\bsnm{Panda}, \binits{A.}},
\bauthor{\bsnm{Song}, \binits{L.}},
\bauthor{\bsnm{Yang}, \binits{Y.}},
\bauthor{\bsnm{Mahoney}, \binits{M.}},
\bauthor{\bsnm{Mittal}, \binits{P.}},
\bauthor{\bsnm{Kannan}, \binits{R.}},
\bauthor{\bsnm{Gonzalez}, \binits{J.}}:
\bctitle{Neurotoxin: Durable backdoors in federated learning}.
In: \bbtitle{Proceedings of the 39th International Conference on Machine
  Learning},
vol. \bseriesno{162},
pp. \bfpage{26429}--\blpage{26446}
(\byear{2022})
\end{bchapter}
\endbibitem

\bibitem[\protect\citeauthoryear{Cao et~al.}{2021}]{provably_secure_FL}
\begin{barticle}
\bauthor{\bsnm{Cao}, \binits{X.}},
\bauthor{\bsnm{Jia}, \binits{J.}},
\bauthor{\bsnm{Gong}, \binits{N.Z.}}:
\batitle{Provably secure federated learning against malicious clients}.
\bjtitle{Proceedings of the AAAI Conference on Artificial Intelligence}
\bvolume{35}(\bissue{8}),
\bfpage{6885}--\blpage{6893}
(\byear{2021})
\doiurl{10.1609/aaai.v35i8.16849}
\end{barticle}
\endbibitem

\bibitem[\protect\citeauthoryear{Wen et~al.}{2022}]{gradient_magnification}
\begin{bchapter}
\bauthor{\bsnm{Wen}, \binits{Y.}},
\bauthor{\bsnm{Geiping}, \binits{J.A.}},
\bauthor{\bsnm{Fowl}, \binits{L.}},
\bauthor{\bsnm{Goldblum}, \binits{M.}},
\bauthor{\bsnm{Goldstein}, \binits{T.}}:
\bctitle{Fishing for user data in large-batch federated learning via gradient
  magnification}.
In: \beditor{\bsnm{Chaudhuri}, \binits{K.}},
\beditor{\bsnm{Jegelka}, \binits{S.}},
\beditor{\bsnm{Song}, \binits{L.}},
\beditor{\bsnm{Szepesvari}, \binits{C.}},
\beditor{\bsnm{Niu}, \binits{G.}},
\beditor{\bsnm{Sabato}, \binits{S.}} (eds.)
\bbtitle{Proceedings of the 39th International Conference on Machine Learning}.
\bsertitle{Proceedings of Machine Learning Research},
vol. \bseriesno{162},
pp. \bfpage{23668}--\blpage{23684}
(\byear{2022}).
\burl{https://proceedings.mlr.press/v162/wen22a.html}
\end{bchapter}
\endbibitem

\bibitem[\protect\citeauthoryear{Gupta et~al.}{2022}]{gupta2022recovering}
\begin{barticle}
\bauthor{\bsnm{Gupta}, \binits{S.}},
\bauthor{\bsnm{Huang}, \binits{Y.}},
\bauthor{\bsnm{Zhong}, \binits{Z.}},
\bauthor{\bsnm{Gao}, \binits{T.}},
\bauthor{\bsnm{Li}, \binits{K.}},
\bauthor{\bsnm{Chen}, \binits{D.}}:
\batitle{Recovering private text in federated learning of language models}.
\bjtitle{Advances in Neural Information Processing Systems}
\bvolume{35},
\bfpage{8130}--\blpage{8143}
(\byear{2022})
\end{barticle}
\endbibitem

\bibitem[\protect\citeauthoryear{Bietti
  et~al.}{2022}]{privacy_accuracy_tradeoffs}
\begin{bchapter}
\bauthor{\bsnm{Bietti}, \binits{A.}},
\bauthor{\bsnm{Wei}, \binits{C.-Y.}},
\bauthor{\bsnm{Dudik}, \binits{M.}},
\bauthor{\bsnm{Langford}, \binits{J.}},
\bauthor{\bsnm{Wu}, \binits{S.}}:
\bctitle{Personalization improves privacy-accuracy tradeoffs in federated
  learning}.
In: \beditor{\bsnm{Chaudhuri}, \binits{K.}},
\beditor{\bsnm{Jegelka}, \binits{S.}},
\beditor{\bsnm{Song}, \binits{L.}},
\beditor{\bsnm{Szepesvari}, \binits{C.}},
\beditor{\bsnm{Niu}, \binits{G.}},
\beditor{\bsnm{Sabato}, \binits{S.}} (eds.)
\bbtitle{Proceedings of the 39th International Conference on Machine Learning}.
\bsertitle{Proceedings of Machine Learning Research},
vol. \bseriesno{162},
pp. \bfpage{1945}--\blpage{1962}
(\byear{2022}).
\burl{https://proceedings.mlr.press/v162/bietti22a.html}
\end{bchapter}
\endbibitem

\bibitem[\protect\citeauthoryear{Zhang et~al.}{2022}]{clipping_for_fl}
\begin{bchapter}
\bauthor{\bsnm{Zhang}, \binits{X.}},
\bauthor{\bsnm{Chen}, \binits{X.}},
\bauthor{\bsnm{Hong}, \binits{M.}},
\bauthor{\bsnm{Wu}, \binits{S.}},
\bauthor{\bsnm{Yi}, \binits{J.}}:
\bctitle{Understanding clipping for federated learning: Convergence and
  client-level differential privacy}.
In: \beditor{\bsnm{Chaudhuri}, \binits{K.}},
\beditor{\bsnm{Jegelka}, \binits{S.}},
\beditor{\bsnm{Song}, \binits{L.}},
\beditor{\bsnm{Szepesvari}, \binits{C.}},
\beditor{\bsnm{Niu}, \binits{G.}},
\beditor{\bsnm{Sabato}, \binits{S.}} (eds.)
\bbtitle{Proceedings of the 39th International Conference on Machine Learning}.
\bsertitle{Proceedings of Machine Learning Research},
vol. \bseriesno{162},
pp. \bfpage{26048}--\blpage{26067}
(\byear{2022}).
\burl{https://proceedings.mlr.press/v162/zhang22b.html}
\end{bchapter}
\endbibitem

\bibitem[\protect\citeauthoryear{Hu et~al.}{2021}]{FedSPA}
\begin{bchapter}
\bauthor{\bsnm{Hu}, \binits{R.}},
\bauthor{\bsnm{Gong}, \binits{Y.}},
\bauthor{\bsnm{Guo}, \binits{Y.}}:
\bctitle{Federated learning with sparsification-amplified privacy and adaptive
  optimization}.
In: \beditor{\bsnm{Zhou}, \binits{Z.-H.}} (ed.)
\bbtitle{Proceedings of the Thirtieth International Joint Conference on
  Artificial Intelligence, {IJCAI-21}},
pp. \bfpage{1463}--\blpage{1469}
(\byear{2021}).
\doiurl{10.24963/ijcai.2021/202} .
\bcomment{Main Track}.
\burl{https://doi.org/10.24963/ijcai.2021/202}
\end{bchapter}
\endbibitem

\bibitem[\protect\citeauthoryear{Sun et~al.}{2021}]{LDP_FL}
\begin{bchapter}
\bauthor{\bsnm{Sun}, \binits{L.}},
\bauthor{\bsnm{Qian}, \binits{J.}},
\bauthor{\bsnm{Chen}, \binits{X.}}:
\bctitle{{LDP-FL}: Practical private aggregation in federated learning with
  local differential privacy}.
In: \beditor{\bsnm{Zhou}, \binits{Z.-H.}} (ed.)
\bbtitle{Proceedings of the Thirtieth International Joint Conference on
  Artificial Intelligence, {IJCAI-21}},
pp. \bfpage{1571}--\blpage{1578}
(\byear{2021}).
\doiurl{10.24963/ijcai.2021/217} .
\bcomment{Main Track}.
\burl{https://doi.org/10.24963/ijcai.2021/217}
\end{bchapter}
\endbibitem

\bibitem[\protect\citeauthoryear{Peng et~al.}{2021}]{FKGE}
\begin{bchapter}
\bauthor{\bsnm{Peng}, \binits{H.}},
\bauthor{\bsnm{Li}, \binits{H.}},
\bauthor{\bsnm{Song}, \binits{Y.}},
\bauthor{\bsnm{Zheng}, \binits{V.}},
\bauthor{\bsnm{Li}, \binits{J.}}:
\bctitle{Differentially private federated knowledge graphs embedding}.
In: \bbtitle{Proceedings of the 30th ACM International Conference on
  Information \& Knowledge Management},
pp. \bfpage{1416}--\blpage{1425}
(\byear{2021})
\end{bchapter}
\endbibitem

\bibitem[\protect\citeauthoryear{Fan et~al.}{2023}]{fan2023flsg}
\begin{botherref}
\oauthor{\bsnm{Fan}, \binits{K.}},
\oauthor{\bsnm{Hong}, \binits{J.}},
\oauthor{\bsnm{Li}, \binits{W.}},
\oauthor{\bsnm{Zhao}, \binits{X.}},
\oauthor{\bsnm{Li}, \binits{H.}},
\oauthor{\bsnm{Yang}, \binits{Y.}}:
Flsg: A novel defense strategy against inference attacks in vertical federated
  learning.
IEEE Internet of Things Journal
(2023)
\end{botherref}
\endbibitem

\bibitem[\protect\citeauthoryear{Rong et~al.}{2022}]{recommender_poisoning}
\begin{bchapter}
\bauthor{\bsnm{Rong}, \binits{D.}},
\bauthor{\bsnm{He}, \binits{Q.}},
\bauthor{\bsnm{Chen}, \binits{J.}}:
\bctitle{Poisoning deep learning based recommender model in federated learning
  scenarios}.
In: \beditor{\bsnm{Raedt}, \binits{L.D.}} (ed.)
\bbtitle{Proceedings of the Thirty-First International Joint Conference on
  Artificial Intelligence, {IJCAI-22}},
pp. \bfpage{2204}--\blpage{2210}
(\byear{2022}).
\doiurl{10.24963/ijcai.2022/306} .
\bcomment{Main Track}.
\burl{https://doi.org/10.24963/ijcai.2022/306}
\end{bchapter}
\endbibitem

\bibitem[\protect\citeauthoryear{Huang
  et~al.}{2021}]{gradient_inversion_attacks}
\begin{bchapter}
\bauthor{\bsnm{Huang}, \binits{Y.}},
\bauthor{\bsnm{Gupta}, \binits{S.}},
\bauthor{\bsnm{Song}, \binits{Z.}},
\bauthor{\bsnm{Li}, \binits{K.}},
\bauthor{\bsnm{Arora}, \binits{S.}}:
\bctitle{Evaluating gradient inversion attacks and defenses in federated
  learning}.
In: \beditor{\bsnm{Ranzato}, \binits{M.}},
\beditor{\bsnm{Beygelzimer}, \binits{A.}},
\beditor{\bsnm{Dauphin}, \binits{Y.}},
\beditor{\bsnm{Liang}, \binits{P.S.}},
\beditor{\bsnm{Vaughan}, \binits{J.W.}} (eds.)
\bbtitle{Advances in Neural Information Processing Systems},
vol. \bseriesno{34},
pp. \bfpage{7232}--\blpage{7241}
(\byear{2021}).
\burl{https://proceedings.neurips.cc/paper/2021/file/3b3fff6463464959dcd1b68d0320f781-Paper.pdf}
\end{bchapter}
\endbibitem

\bibitem[\protect\citeauthoryear{Jin et~al.}{2021}]{CAFE}
\begin{bchapter}
\bauthor{\bsnm{Jin}, \binits{X.}},
\bauthor{\bsnm{Chen}, \binits{P.-Y.}},
\bauthor{\bsnm{Hsu}, \binits{C.-Y.}},
\bauthor{\bsnm{Yu}, \binits{C.-M.}},
\bauthor{\bsnm{Chen}, \binits{T.}}:
\bctitle{{CAFE}: Catastrophic data leakage in vertical federated learning}.
In: \beditor{\bsnm{Ranzato}, \binits{M.}},
\beditor{\bsnm{Beygelzimer}, \binits{A.}},
\beditor{\bsnm{Dauphin}, \binits{Y.}},
\beditor{\bsnm{Liang}, \binits{P.S.}},
\beditor{\bsnm{Vaughan}, \binits{J.W.}} (eds.)
\bbtitle{Advances in Neural Information Processing Systems},
vol. \bseriesno{34},
pp. \bfpage{994}--\blpage{1006}
(\byear{2021}).
\burl{https://proceedings.neurips.cc/paper/2021/file/08040837089cdf46631a10aca5258e16-Paper.pdf}
\end{bchapter}
\endbibitem

\bibitem[\protect\citeauthoryear{Sun et~al.}{2021}]{FLWBC}
\begin{bchapter}
\bauthor{\bsnm{Sun}, \binits{J.}},
\bauthor{\bsnm{Li}, \binits{A.}},
\bauthor{\bsnm{DiValentin}, \binits{L.}},
\bauthor{\bsnm{Hassanzadeh}, \binits{A.}},
\bauthor{\bsnm{Chen}, \binits{Y.}},
\bauthor{\bsnm{Li}, \binits{H.}}:
\bctitle{{FL-WBC}: Enhancing robustness against model poisoning attacks in
  federated learning from a client perspective}.
In: \beditor{\bsnm{Ranzato}, \binits{M.}},
\beditor{\bsnm{Beygelzimer}, \binits{A.}},
\beditor{\bsnm{Dauphin}, \binits{Y.}},
\beditor{\bsnm{Liang}, \binits{P.S.}},
\beditor{\bsnm{Vaughan}, \binits{J.W.}} (eds.)
\bbtitle{Advances in Neural Information Processing Systems},
vol. \bseriesno{34},
pp. \bfpage{12613}--\blpage{12624}
(\byear{2021}).
\burl{https://proceedings.neurips.cc/paper/2021/file/692baebec3bb4b53d7ebc3b9fabac31b-Paper.pdf}
\end{bchapter}
\endbibitem

\bibitem[\protect\citeauthoryear{Park et~al.}{2023}]{park2023feddefender}
\begin{bchapter}
\bauthor{\bsnm{Park}, \binits{S.}},
\bauthor{\bsnm{Han}, \binits{S.}},
\bauthor{\bsnm{Wu}, \binits{F.}},
\bauthor{\bsnm{Kim}, \binits{S.}},
\bauthor{\bsnm{Zhu}, \binits{B.}},
\bauthor{\bsnm{Xie}, \binits{X.}},
\bauthor{\bsnm{Cha}, \binits{M.}}:
\bctitle{Feddefender: Client-side attack-tolerant federated learning}.
In: \bbtitle{Proceedings of the 29th ACM SIGKDD Conference on Knowledge
  Discovery and Data Mining},
pp. \bfpage{1850}--\blpage{1861}
(\byear{2023})
\end{bchapter}
\endbibitem

\bibitem[\protect\citeauthoryear{Park et~al.}{2021}]{sageflow}
\begin{bchapter}
\bauthor{\bsnm{Park}, \binits{J.}},
\bauthor{\bsnm{Han}, \binits{D.-J.}},
\bauthor{\bsnm{Choi}, \binits{M.}},
\bauthor{\bsnm{Moon}, \binits{J.}}:
\bctitle{Sageflow: Robust federated learning against both stragglers and
  adversaries}.
In: \beditor{\bsnm{Ranzato}, \binits{M.}},
\beditor{\bsnm{Beygelzimer}, \binits{A.}},
\beditor{\bsnm{Dauphin}, \binits{Y.}},
\beditor{\bsnm{Liang}, \binits{P.S.}},
\beditor{\bsnm{Vaughan}, \binits{J.W.}} (eds.)
\bbtitle{Advances in Neural Information Processing Systems},
vol. \bseriesno{34},
pp. \bfpage{840}--\blpage{851}
(\byear{2021}).
\burl{https://proceedings.neurips.cc/paper/2021/file/076a8133735eb5d7552dc195b125a454-Paper.pdf}
\end{bchapter}
\endbibitem

\bibitem[\protect\citeauthoryear{Agarwal et~al.}{2021}]{skellam_mechanism}
\begin{bchapter}
\bauthor{\bsnm{Agarwal}, \binits{N.}},
\bauthor{\bsnm{Kairouz}, \binits{P.}},
\bauthor{\bsnm{Liu}, \binits{Z.}}:
\bctitle{The skellam mechanism for differentially private federated learning}.
In: \beditor{\bsnm{Ranzato}, \binits{M.}},
\beditor{\bsnm{Beygelzimer}, \binits{A.}},
\beditor{\bsnm{Dauphin}, \binits{Y.}},
\beditor{\bsnm{Liang}, \binits{P.S.}},
\beditor{\bsnm{Vaughan}, \binits{J.W.}} (eds.)
\bbtitle{Advances in Neural Information Processing Systems},
vol. \bseriesno{34},
pp. \bfpage{5052}--\blpage{5064}
(\byear{2021}).
\burl{https://proceedings.neurips.cc/paper/2021/file/285baacbdf8fda1de94b19282acd23e2-Paper.pdf}
\end{bchapter}
\endbibitem

\bibitem[\protect\citeauthoryear{Chang et~al.}{2023}]{chang2023privacy}
\begin{barticle}
\bauthor{\bsnm{Chang}, \binits{Y.}},
\bauthor{\bsnm{Zhang}, \binits{K.}},
\bauthor{\bsnm{Gong}, \binits{J.}},
\bauthor{\bsnm{Qian}, \binits{H.}}:
\batitle{Privacy-preserving federated learning via functional encryption,
  revisited}.
\bjtitle{IEEE Transactions on Information Forensics and Security}
\bvolume{18},
\bfpage{1855}--\blpage{1869}
(\byear{2023})
\end{barticle}
\endbibitem

\bibitem[\protect\citeauthoryear{Hijazi et~al.}{2023}]{hijazi2023secure}
\begin{botherref}
\oauthor{\bsnm{Hijazi}, \binits{N.M.}},
\oauthor{\bsnm{Aloqaily}, \binits{M.}},
\oauthor{\bsnm{Guizani}, \binits{M.}},
\oauthor{\bsnm{Ouni}, \binits{B.}},
\oauthor{\bsnm{Karray}, \binits{F.}}:
Secure federated learning with fully homomorphic encryption for iot
  communications.
IEEE Internet of Things Journal
(2023)
\end{botherref}
\endbibitem

\bibitem[\protect\citeauthoryear{Zhao et~al.}{2022}]{zhao2022practical}
\begin{barticle}
\bauthor{\bsnm{Zhao}, \binits{P.}},
\bauthor{\bsnm{Cao}, \binits{Z.}},
\bauthor{\bsnm{Jiang}, \binits{J.}},
\bauthor{\bsnm{Gao}, \binits{F.}}:
\batitle{Practical private aggregation in federated learning against inference
  attack}.
\bjtitle{IEEE Internet of Things Journal}
\bvolume{10}(\bissue{1}),
\bfpage{318}--\blpage{329}
(\byear{2022})
\end{barticle}
\endbibitem

\bibitem[\protect\citeauthoryear{Gao et~al.}{2021}]{M_LSCG}
\begin{barticle}
\bauthor{\bsnm{Gao}, \binits{H.}},
\bauthor{\bsnm{Xu}, \binits{A.}},
\bauthor{\bsnm{Huang}, \binits{H.}}:
\batitle{On the convergence of communication-efficient local sgd for federated
  learning}.
\bjtitle{Proceedings of the AAAI Conference on Artificial Intelligence}
\bvolume{35}(\bissue{9}),
\bfpage{7510}--\blpage{7518}
(\byear{2021})
\doiurl{10.1609/aaai.v35i9.16920}
\end{barticle}
\endbibitem

\bibitem[\protect\citeauthoryear{Wang et~al.}{2022}]{comm_effect_adaptive}
\begin{bchapter}
\bauthor{\bsnm{Wang}, \binits{Y.}},
\bauthor{\bsnm{Lin}, \binits{L.}},
\bauthor{\bsnm{Chen}, \binits{J.}}:
\bctitle{Communication-efficient adaptive federated learning}.
In: \beditor{\bsnm{Chaudhuri}, \binits{K.}},
\beditor{\bsnm{Jegelka}, \binits{S.}},
\beditor{\bsnm{Song}, \binits{L.}},
\beditor{\bsnm{Szepesvari}, \binits{C.}},
\beditor{\bsnm{Niu}, \binits{G.}},
\beditor{\bsnm{Sabato}, \binits{S.}} (eds.)
\bbtitle{Proceedings of the 39th International Conference on Machine Learning}.
\bsertitle{Proceedings of Machine Learning Research},
vol. \bseriesno{162},
pp. \bfpage{22802}--\blpage{22838}
(\byear{2022}).
\burl{https://proceedings.mlr.press/v162/wang22o.html}
\end{bchapter}
\endbibitem

\bibitem[\protect\citeauthoryear{Tang et~al.}{2022}]{tang2022gossipfl}
\begin{barticle}
\bauthor{\bsnm{Tang}, \binits{Z.}},
\bauthor{\bsnm{Shi}, \binits{S.}},
\bauthor{\bsnm{Li}, \binits{B.}},
\bauthor{\bsnm{Chu}, \binits{X.}}:
\batitle{Gossipfl: A decentralized federated learning framework with sparsified
  and adaptive communication}.
\bjtitle{IEEE Transactions on Parallel and Distributed Systems}
\bvolume{34}(\bissue{3}),
\bfpage{909}--\blpage{922}
(\byear{2022})
\end{barticle}
\endbibitem

\bibitem[\protect\citeauthoryear{Yi et~al.}{2022}]{QSFL}
\begin{bchapter}
\bauthor{\bsnm{Yi}, \binits{L.}},
\bauthor{\bsnm{Gang}, \binits{W.}},
\bauthor{\bsnm{Xiaoguang}, \binits{L.}}:
\bctitle{{QSFL}: A two-level uplink communication optimization framework for
  federated learning}.
In: \beditor{\bsnm{Chaudhuri}, \binits{K.}},
\beditor{\bsnm{Jegelka}, \binits{S.}},
\beditor{\bsnm{Song}, \binits{L.}},
\beditor{\bsnm{Szepesvari}, \binits{C.}},
\beditor{\bsnm{Niu}, \binits{G.}},
\beditor{\bsnm{Sabato}, \binits{S.}} (eds.)
\bbtitle{Proceedings of the 39th International Conference on Machine Learning}.
\bsertitle{Proceedings of Machine Learning Research},
vol. \bseriesno{162},
pp. \bfpage{25501}--\blpage{25513}
(\byear{2022}).
\burl{https://proceedings.mlr.press/v162/yi22a.html}
\end{bchapter}
\endbibitem

\bibitem[\protect\citeauthoryear{Zhu et~al.}{2022}]{zhu2022resilient}
\begin{barticle}
\bauthor{\bsnm{Zhu}, \binits{Z.}},
\bauthor{\bsnm{Hong}, \binits{J.}},
\bauthor{\bsnm{Drew}, \binits{S.}},
\bauthor{\bsnm{Zhou}, \binits{J.}}:
\batitle{Resilient and communication efficient learning for heterogeneous
  federated systems}.
\bjtitle{Proceedings of machine learning research}
\bvolume{162},
\bfpage{27504}
(\byear{2022})
\end{barticle}
\endbibitem

\bibitem[\protect\citeauthoryear{Yapp et~al.}{2021}]{BFEL}
\begin{bchapter}
\bauthor{\bsnm{Yapp}, \binits{A.Z.H.}},
\bauthor{\bsnm{Koh}, \binits{H.S.N.}},
\bauthor{\bsnm{Lai}, \binits{Y.T.}},
\bauthor{\bsnm{Kang}, \binits{J.}},
\bauthor{\bsnm{Li}, \binits{X.}},
\bauthor{\bsnm{Ng}, \binits{J.S.}},
\bauthor{\bsnm{Jiang}, \binits{H.}},
\bauthor{\bsnm{Lim}, \binits{W.Y.B.}},
\bauthor{\bsnm{Xiong}, \binits{Z.}},
\bauthor{\bsnm{Niyato}, \binits{D.}}:
\bctitle{Communication-efficient and scalable decentralized federated edge
  learning}.
In: \beditor{\bsnm{Zhou}, \binits{Z.-H.}} (ed.)
\bbtitle{Proceedings of the Thirtieth International Joint Conference on
  Artificial Intelligence, {IJCAI-21}},
pp. \bfpage{5032}--\blpage{5035}
(\byear{2021}).
\doiurl{10.24963/ijcai.2021/720} .
\bcomment{Demo Track}.
\burl{https://doi.org/10.24963/ijcai.2021/720}
\end{bchapter}
\endbibitem

\bibitem[\protect\citeauthoryear{Zhu et~al.}{2021}]{DGA}
\begin{bchapter}
\bauthor{\bsnm{Zhu}, \binits{L.}},
\bauthor{\bsnm{Lin}, \binits{H.}},
\bauthor{\bsnm{Lu}, \binits{Y.}},
\bauthor{\bsnm{Lin}, \binits{Y.}},
\bauthor{\bsnm{Han}, \binits{S.}}:
\bctitle{Delayed gradient averaging: Tolerate the communication latency for
  federated learning}.
In: \beditor{\bsnm{Ranzato}, \binits{M.}},
\beditor{\bsnm{Beygelzimer}, \binits{A.}},
\beditor{\bsnm{Dauphin}, \binits{Y.}},
\beditor{\bsnm{Liang}, \binits{P.S.}},
\beditor{\bsnm{Vaughan}, \binits{J.W.}} (eds.)
\bbtitle{Advances in Neural Information Processing Systems},
vol. \bseriesno{34},
pp. \bfpage{29995}--\blpage{30007}
(\byear{2021}).
\burl{https://proceedings.neurips.cc/paper/2021/file/fc03d48253286a798f5116ec00e99b2b-Paper.pdf}
\end{bchapter}
\endbibitem

\bibitem[\protect\citeauthoryear{Isik et~al.}{2022}]{isik2022sparse}
\begin{bchapter}
\bauthor{\bsnm{Isik}, \binits{B.}},
\bauthor{\bsnm{Pase}, \binits{F.}},
\bauthor{\bsnm{Gunduz}, \binits{D.}},
\bauthor{\bsnm{Weissman}, \binits{T.}},
\bauthor{\bsnm{Michele}, \binits{Z.}}:
\bctitle{Sparse random networks for communication-efficient federated
  learning}.
In: \bbtitle{The Eleventh International Conference on Learning Representations}
(\byear{2022})
\end{bchapter}
\endbibitem

\bibitem[\protect\citeauthoryear{Wang et~al.}{2022}]{wang2022progfed}
\begin{bchapter}
\bauthor{\bsnm{Wang}, \binits{H.-P.}},
\bauthor{\bsnm{Stich}, \binits{S.}},
\bauthor{\bsnm{He}, \binits{Y.}},
\bauthor{\bsnm{Fritz}, \binits{M.}}:
\bctitle{Progfed: effective, communication, and computation efficient federated
  learning by progressive training}.
In: \bbtitle{International Conference on Machine Learning},
pp. \bfpage{23034}--\blpage{23054}
(\byear{2022}).
\bcomment{PMLR}
\end{bchapter}
\endbibitem

\bibitem[\protect\citeauthoryear{Li and Wang}{2022}]{li2022communication}
\begin{barticle}
\bauthor{\bsnm{Li}, \binits{C.}},
\bauthor{\bsnm{Wang}, \binits{H.}}:
\batitle{Communication efficient federated learning for generalized linear
  bandits}.
\bjtitle{Advances in Neural Information Processing Systems}
\bvolume{35},
\bfpage{38411}--\blpage{38423}
(\byear{2022})
\end{barticle}
\endbibitem

\bibitem[\protect\citeauthoryear{Sun and Wei}{2022}]{sun2022communication}
\begin{barticle}
\bauthor{\bsnm{Sun}, \binits{Z.}},
\bauthor{\bsnm{Wei}, \binits{E.}}:
\batitle{A communication-efficient algorithm with linear convergence for
  federated minimax learning}.
\bjtitle{Advances in Neural Information Processing Systems}
\bvolume{35},
\bfpage{6060}--\blpage{6073}
(\byear{2022})
\end{barticle}
\endbibitem

\bibitem[\protect\citeauthoryear{Cui et~al.}{2022}]{cui2022optimizing}
\begin{botherref}
\oauthor{\bsnm{Cui}, \binits{Y.}},
\oauthor{\bsnm{Cao}, \binits{K.}},
\oauthor{\bsnm{Zhou}, \binits{J.}},
\oauthor{\bsnm{Wei}, \binits{T.}}:
Optimizing training efficiency and cost of hierarchical federated learning in
  heterogeneous mobile-edge cloud computing.
IEEE Transactions on Computer-Aided Design of Integrated Circuits and Systems
(2022)
\end{botherref}
\endbibitem

\bibitem[\protect\citeauthoryear{Tang et~al.}{2023}]{tang2023fusionai}
\begin{botherref}
\oauthor{\bsnm{Tang}, \binits{Z.}},
\oauthor{\bsnm{Wang}, \binits{Y.}},
\oauthor{\bsnm{He}, \binits{X.}},
\oauthor{\bsnm{Zhang}, \binits{L.}},
\oauthor{\bsnm{Pan}, \binits{X.}},
\oauthor{\bsnm{Wang}, \binits{Q.}},
\oauthor{\bsnm{Zeng}, \binits{R.}},
\oauthor{\bsnm{Zhao}, \binits{K.}},
\oauthor{\bsnm{Shi}, \binits{S.}},
\oauthor{\bsnm{He}, \binits{B.}}, et al.:
Fusionai: Decentralized training and deploying llms with massive consumer-level
  gpus.
arXiv preprint arXiv:2309.01172
(2023)
\end{botherref}
\endbibitem

\bibitem[\protect\citeauthoryear{Tan et~al.}{2023}]{tan2023federated}
\begin{bchapter}
\bauthor{\bsnm{Tan}, \binits{Y.}},
\bauthor{\bsnm{Liu}, \binits{Y.}},
\bauthor{\bsnm{Long}, \binits{G.}},
\bauthor{\bsnm{Jiang}, \binits{J.}},
\bauthor{\bsnm{Lu}, \binits{Q.}},
\bauthor{\bsnm{Zhang}, \binits{C.}}:
\bctitle{Federated learning on non-iid graphs via structural knowledge
  sharing}.
In: \bbtitle{Proceedings of the AAAI Conference on Artificial Intelligence},
vol. \bseriesno{37},
pp. \bfpage{9953}--\blpage{9961}
(\byear{2023})
\end{bchapter}
\endbibitem

\bibitem[\protect\citeauthoryear{Pan and Zhu}{2022}]{pan2022fedwalk}
\begin{bchapter}
\bauthor{\bsnm{Pan}, \binits{Q.}},
\bauthor{\bsnm{Zhu}, \binits{Y.}}:
\bctitle{Fedwalk: Communication efficient federated unsupervised node embedding
  with differential privacy}.
In: \bbtitle{Proceedings of the 28th ACM SIGKDD Conference on Knowledge
  Discovery and Data Mining},
pp. \bfpage{1317}--\blpage{1326}
(\byear{2022})
\end{bchapter}
\endbibitem

\bibitem[\protect\citeauthoryear{Liang et~al.}{2021}]{liang2021fedrec++}
\begin{bchapter}
\bauthor{\bsnm{Liang}, \binits{F.}},
\bauthor{\bsnm{Pan}, \binits{W.}},
\bauthor{\bsnm{Ming}, \binits{Z.}}:
\bctitle{Fedrec++: Lossless federated recommendation with explicit feedback}.
In: \bbtitle{Proceedings of the AAAI Conference on Artificial Intelligence},
vol. \bseriesno{35},
pp. \bfpage{4224}--\blpage{4231}
(\byear{2021})
\end{bchapter}
\endbibitem

\bibitem[\protect\citeauthoryear{Zhu et~al.}{2022}]{zhu2022cali3f}
\begin{bchapter}
\bauthor{\bsnm{Zhu}, \binits{Z.}},
\bauthor{\bsnm{Si}, \binits{S.}},
\bauthor{\bsnm{Wang}, \binits{J.}},
\bauthor{\bsnm{Xiao}, \binits{J.}}:
\bctitle{Cali3f: Calibrated fast fair federated recommendation system}.
In: \bbtitle{2022 International Joint Conference on Neural Networks (IJCNN)},
pp. \bfpage{1}--\blpage{8}
(\byear{2022}).
\bcomment{IEEE}
\end{bchapter}
\endbibitem

\bibitem[\protect\citeauthoryear{Liu et~al.}{2022}]{liu2022federated}
\begin{barticle}
\bauthor{\bsnm{Liu}, \binits{Z.}},
\bauthor{\bsnm{Yang}, \binits{L.}},
\bauthor{\bsnm{Fan}, \binits{Z.}},
\bauthor{\bsnm{Peng}, \binits{H.}},
\bauthor{\bsnm{Yu}, \binits{P.S.}}:
\batitle{Federated social recommendation with graph neural network}.
\bjtitle{ACM Transactions on Intelligent Systems and Technology (TIST)}
\bvolume{13}(\bissue{4}),
\bfpage{1}--\blpage{24}
(\byear{2022})
\end{barticle}
\endbibitem

\bibitem[\protect\citeauthoryear{Yuan et~al.}{2023}]{yuan2023federated}
\begin{bchapter}
\bauthor{\bsnm{Yuan}, \binits{W.}},
\bauthor{\bsnm{Yin}, \binits{H.}},
\bauthor{\bsnm{Wu}, \binits{F.}},
\bauthor{\bsnm{Zhang}, \binits{S.}},
\bauthor{\bsnm{He}, \binits{T.}},
\bauthor{\bsnm{Wang}, \binits{H.}}:
\bctitle{Federated unlearning for on-device recommendation}.
In: \bbtitle{Proceedings of the Sixteenth ACM International Conference on Web
  Search and Data Mining},
pp. \bfpage{393}--\blpage{401}
(\byear{2023})
\end{bchapter}
\endbibitem

\bibitem[\protect\citeauthoryear{Xu et~al.}{2021}]{xu2021privacy}
\begin{barticle}
\bauthor{\bsnm{Xu}, \binits{X.}},
\bauthor{\bsnm{Peng}, \binits{H.}},
\bauthor{\bsnm{Bhuiyan}, \binits{M.Z.A.}},
\bauthor{\bsnm{Hao}, \binits{Z.}},
\bauthor{\bsnm{Liu}, \binits{L.}},
\bauthor{\bsnm{Sun}, \binits{L.}},
\bauthor{\bsnm{He}, \binits{L.}}:
\batitle{Privacy-preserving federated depression detection from multisource
  mobile health data}.
\bjtitle{IEEE transactions on industrial informatics}
\bvolume{18}(\bissue{7}),
\bfpage{4788}--\blpage{4797}
(\byear{2021})
\end{barticle}
\endbibitem

\bibitem[\protect\citeauthoryear{Che et~al.}{2022}]{che2022federated}
\begin{barticle}
\bauthor{\bsnm{Che}, \binits{S.}},
\bauthor{\bsnm{Kong}, \binits{Z.}},
\bauthor{\bsnm{Peng}, \binits{H.}},
\bauthor{\bsnm{Sun}, \binits{L.}},
\bauthor{\bsnm{Leow}, \binits{A.}},
\bauthor{\bsnm{Chen}, \binits{Y.}},
\bauthor{\bsnm{He}, \binits{L.}}:
\batitle{Federated multi-view learning for private medical data integration and
  analysis}.
\bjtitle{ACM Transactions on Intelligent Systems and Technology (TIST)}
\bvolume{13}(\bissue{4}),
\bfpage{1}--\blpage{23}
(\byear{2022})
\end{barticle}
\endbibitem

\bibitem[\protect\citeauthoryear{Liu et~al.}{2022}]{liu2022contribution}
\begin{bchapter}
\bauthor{\bsnm{Liu}, \binits{Z.}},
\bauthor{\bsnm{Chen}, \binits{Y.}},
\bauthor{\bsnm{Zhao}, \binits{Y.}},
\bauthor{\bsnm{Yu}, \binits{H.}},
\bauthor{\bsnm{Liu}, \binits{Y.}},
\bauthor{\bsnm{Bao}, \binits{R.}},
\bauthor{\bsnm{Jiang}, \binits{J.}},
\bauthor{\bsnm{Nie}, \binits{Z.}},
\bauthor{\bsnm{Xu}, \binits{Q.}},
\bauthor{\bsnm{Yang}, \binits{Q.}}:
\bctitle{Contribution-aware federated learning for smart healthcare}.
In: \bbtitle{Proceedings of the AAAI Conference on Artificial Intelligence},
vol. \bseriesno{36},
pp. \bfpage{12396}--\blpage{12404}
(\byear{2022})
\end{bchapter}
\endbibitem

\bibitem[\protect\citeauthoryear{Chen et~al.}{2023}]{chen2023medical}
\begin{barticle}
\bauthor{\bsnm{Chen}, \binits{Z.}},
\bauthor{\bsnm{Li}, \binits{W.}},
\bauthor{\bsnm{Xing}, \binits{X.}},
\bauthor{\bsnm{Yuan}, \binits{Y.}}:
\batitle{Medical federated learning with joint graph purification for noisy
  label learning}.
\bjtitle{Medical Image Analysis}
\bvolume{90},
\bfpage{102976}
(\byear{2023})
\end{barticle}
\endbibitem

\bibitem[\protect\citeauthoryear{Zhu et~al.}{2023}]{zhu2023feddm}
\begin{botherref}
\oauthor{\bsnm{Zhu}, \binits{M.}},
\oauthor{\bsnm{Chen}, \binits{Z.}},
\oauthor{\bsnm{Yuan}, \binits{Y.}}:
{FedDM}: Federated weakly supervised segmentation via annotation calibration
  and gradient de-conflicting.
IEEE Transactions on Medical Imaging
(2023)
\end{botherref}
\endbibitem

\bibitem[\protect\citeauthoryear{Long et~al.}{2020}]{long2020federated}
\begin{bchapter}
\bauthor{\bsnm{Long}, \binits{G.}},
\bauthor{\bsnm{Tan}, \binits{Y.}},
\bauthor{\bsnm{Jiang}, \binits{J.}},
\bauthor{\bsnm{Zhang}, \binits{C.}}:
\bctitle{Federated learning for open banking}.
In: \bbtitle{Federated Learning: Privacy and Incentive},
pp. \bfpage{240}--\blpage{254}
(\byear{2020})
\end{bchapter}
\endbibitem

\bibitem[\protect\citeauthoryear{Wang et~al.}{2019}]{wang2019federated}
\begin{botherref}
\oauthor{\bsnm{Wang}, \binits{K.}},
\oauthor{\bsnm{Mathews}, \binits{R.}},
\oauthor{\bsnm{Kiddon}, \binits{C.}},
\oauthor{\bsnm{Eichner}, \binits{H.}},
\oauthor{\bsnm{Beaufays}, \binits{F.}},
\oauthor{\bsnm{Ramage}, \binits{D.}}:
Federated evaluation of on-device personalization.
arXiv preprint arXiv:1910.10252
(2019)
\end{botherref}
\endbibitem

\bibitem[\protect\citeauthoryear{He et~al.}{2020}]{chaoyanghe2020fedml}
\begin{botherref}
\oauthor{\bsnm{He}, \binits{C.}},
\oauthor{\bsnm{Li}, \binits{S.}},
\oauthor{\bsnm{So}, \binits{J.}},
\oauthor{\bsnm{Zhang}, \binits{M.}},
\oauthor{\bsnm{Wang}, \binits{H.}},
\oauthor{\bsnm{Wang}, \binits{X.}},
\oauthor{\bsnm{Vepakomma}, \binits{P.}},
\oauthor{\bsnm{Singh}, \binits{A.}},
\oauthor{\bsnm{Qiu}, \binits{H.}},
\oauthor{\bsnm{Shen}, \binits{L.}},
\oauthor{\bsnm{Zhao}, \binits{P.}},
\oauthor{\bsnm{Kang}, \binits{Y.}},
\oauthor{\bsnm{Liu}, \binits{Y.}},
\oauthor{\bsnm{Raskar}, \binits{R.}},
\oauthor{\bsnm{Yang}, \binits{Q.}},
\oauthor{\bsnm{Annavaram}, \binits{M.}},
\oauthor{\bsnm{Avestimehr}, \binits{S.}}:
{FedML: A Research Library and Benchmark for Federated Machine Learning}.
arXiv preprint arXiv:2007.13518
(2020)
\end{botherref}
\endbibitem

\bibitem[\protect\citeauthoryear{Tang et~al.}{2023}]{tang2023fedml}
\begin{botherref}
\oauthor{\bsnm{Tang}, \binits{Z.}},
\oauthor{\bsnm{Chu}, \binits{X.}},
\oauthor{\bsnm{Ran}, \binits{R.Y.}},
\oauthor{\bsnm{Lee}, \binits{S.}},
\oauthor{\bsnm{Shi}, \binits{S.}},
\oauthor{\bsnm{Zhang}, \binits{Y.}},
\oauthor{\bsnm{Wang}, \binits{Y.}},
\oauthor{\bsnm{Liang}, \binits{A.Q.}},
\oauthor{\bsnm{Avestimehr}, \binits{S.}},
\oauthor{\bsnm{He}, \binits{C.}}:
Fedml parrot: A scalable federated learning system via heterogeneity-aware
  scheduling on sequential and hierarchical training.
arXiv preprint arXiv:2303.01778
(2023)
\end{botherref}
\endbibitem

\end{thebibliography}

\end{document}